\documentclass[journal]{IEEEtran}
\usepackage{amsmath,amssymb,amsfonts}
\usepackage{graphicx}
\usepackage{placeins}
\usepackage{booktabs}
\usepackage{array}
\usepackage{multirow}
\usepackage{cite}
\usepackage{xcolor}
\usepackage{tikz}
\usetikzlibrary{arrows.meta,positioning,calc}
\usepackage[hidelinks]{hyperref}
\newtheorem{theorem}{Theorem}
\newtheorem{proposition}{Proposition}
\newtheorem{corollary}{Corollary}
\newtheorem{assumption}{Assumption}

\input{generated/results_macros}

\newif\ifshowappendix
\ifdefined\submissionversion\showappendixfalse\else\showappendixtrue\fi
\ifshowappendix\newcommand{\suppref}{the appendix}\else\newcommand{\suppref}{the supplementary material}\fi

\newcommand{\bv}[1]{\boldsymbol{#1}}

\newcommand{\sref}[1]{\ref{#1}}\newcommand{\mref}[1]{\ref{#1}}
\begin{document}
\title{Actuator-Lag-Aligned Composite Learning for Adaptive Quadrotor Control Under a Rotor-Command Activity Budget}

\author{
	Xuan-Quy~Dao\hspace{1mm}\href{https://orcid.org/0000-0001-5415-7538}
	{\includegraphics[scale=0.06]{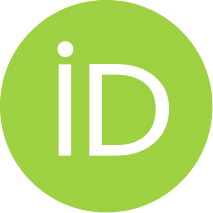}} and
	Ngoc-Bich~Le\hspace{1mm}\href{https://orcid.org/0000-0001-7431-0157}
	{\includegraphics[scale=0.06]{orcid.pdf}}

	\thanks{X.-Q. Dao is with Eastern International University,
		Ho Chi Minh City, Vietnam
		(e-mail: \texttt{quy.dao@eiu.edu.vn}). (\emph{Corresponding author: Xuan-Quy Dao.})}%
	\thanks{N.-B. Le is with Industrial University of Ho Chi Minh City,
		Ho Chi Minh City, Vietnam
		(e-mail: \texttt{lengocbich@iuh.edu.vn}).}
}

\markboth{}{Dao et al.: Actuator-Lag-Aligned Composite Learning for Quadrotor Control}
\maketitle

\begin{abstract}
Composite-learning model reference adaptive control (MRAC) adds a prediction error to the tracking-error update. In a quadrotor cascade, the target can use commanded or realized inputs. The two are compared with every controller selected to keep its rotor-command activity in a noisy step within 1.05 times that of a linear-quadratic regulator (LQR). In a radial-basis-function (RBF) neural MRAC, the realized-input target needs only a nominal motor model and the measured attitude, not acceleration or rotor-speed sensing. Ablations and a one-axis analysis trace the benefit to delaying the commanded thrust and torques through a first-order actuator model. A linear filter with the motor time constant matches the nonlinear motor model, equal filtering of both sides of the target does not, and removing the RBF neurons changes the errors of the lag-aligned profiles by less than 1\,mm. With lag-aligned targets in both loops and a multi-seed guard, the mean error over six evaluation scenarios (Eval6) is \val{rIE}{bench}{eval6}\,mm with the linear filter and \val{det}{GUARDboth}{eval6}\,mm with the motor model; the best guarded non-neural comparator reaches \val{rB}{indi2}{eval6}\,mm. The advantage is conditional. If maneuver activity must also stay at LQR level, geometric control with a disturbance observer has the lowest design cost, and the activity benefit disappears at twice the modeled motor constant. In PX4 software-in-the-loop simulation, the lag-aligned controller tracks better than LQR on every tested seed, but no tested gain keeps hover activity within the threshold on all seeds.
\end{abstract}


\section{Introduction}\label{sec:intro}
\IEEEPARstart{Q}{uadrotor} unmanned aerial vehicles are used for inspection, mapping and aerial
photography, where accurate trajectory tracking must be maintained despite wind, payload and
inertia changes, aerodynamic drag and unmodeled actuator dynamics. Model-based controllers such as
linear-quadratic (LQ) control \cite{bouabdallah2004} and geometric control \cite{lee2010geometric} provide nominal
tracking baselines, and uncertainties of this kind motivate adaptive control. MRAC was evaluated in flight on quadrotors \cite{dydek2013} and is a standard
framework for aerospace adaptive control \cite{lavretsky2013}; neural networks (NNs) extend MRAC to
unstructured, state-dependent uncertainty \cite{lewis1996,joshi2019,patil2022}.

Madebo \emph{et al.} \cite{madebo2024} proposed a hierarchical NN-based MRAC for a quaternion quadrotor model: a feed-forward NN (FFNN) trained to reproduce an MRAC position controller and a recurrent NN (RNN) self-tuning attitude regulator, both refined online by back-propagation (BP) when the tracking error exceeds a threshold, which outperformed an LQR and conventional MRAC in simulation. Three issues remain open. (i) The position network is trained on MRAC data and the attitude regulator on a learned plant model, with no common Lyapunov proof for the threshold-triggered updates. (ii) Faster adaptation in MRAC can produce oscillatory transients \cite{gibson2013}, and conventional gradient MRAC needs persistent excitation for parameter convergence \cite{narendra1987}. (iii) Sensor noise, motor lag and comparisons under a common command-activity limit are not documented.

Composite and concurrent learning combine the tracking error with a prediction error of the uncertainty model \cite{lavretsky2009,chowdhary2010}. With a sufficiently informative stored history \cite{chowdhary2010,chowdhary2013}, interval excitation \cite{pan2016} or finite excitation with retained data \cite{cho2018}, they reach parameter convergence without persistent excitation. In a cascade, however, a prediction error computed from the \emph{commanded} input also contains the lag of the inner loop and of the motors, which is not a function of the regressor. This paper asks whether a prediction target built from realized inputs permits faster adaptation when all controllers must respect the same command-activity budget. The target does not depend on the form of the regressor; it is tested here in a radial-basis-function (RBF) neural MRAC and, as an ablation, with a linear regressor. The target is built from an open-loop nominal motor model and the measured attitude, without acceleration or rotor-speed measurements; Section~\ref{sec:related} compares it with methods that use such measurements. The contributions are:
\begin{enumerate}
\item a derivative-free learning target that replaces the commanded inputs by realized inputs computed from the measured attitude and a nominal motor model, with ablations and a one-axis linear analysis showing that its benefit under an activity budget comes from a first-order actuator model in the input branch of the target, which the nonlinear motor model does not improve, symmetric filtering of the target does not replace, and a linear regressor without RBF neurons nearly preserves;
\item common-gain and activity-constrained comparisons of CL-Inner (motor predictor in the attitude-loop target only) and CL-Both (predictors in both targets) against adaptive, neural, incremental nonlinear dynamic inversion (INDI), $\mathcal{L}_1$ and geometric/disturbance-observer baselines, including paired random-parameter tests. The result depends on the activity threshold: on the seed-1 tuning grids of the families in Fig.~\ref{fig:pareto}, CL-Both has the lowest design cost for noise-driven activity thresholds up to \val{par}{all}{bothhi} times that of LQR; if only maneuver activity is limited, INDI attains lower design cost at much higher noise-driven activity, and when maneuver activity is held at the LQR level, geometric control with a disturbance observer has the lowest design cost (Section~\ref{sec:budget}).
\end{enumerate}
The actuator model does not enter the control law, the allocation or an incremental inversion. It enters only the input branch of the composite-learning target, the signal that updates the weights, where it delays the commanded thrust and torques toward the acceleration they produce. In a one-axis linear model with a matched actuator, this delay removes from the prediction error a filtered derivative of the command (Section~\ref{sec:extension}, Appendix~\sref{app:r2-lin}). In the simulated cascade, the observed effect is lower noise-driven command activity, so larger adaptation gains meet the same threshold. The resulting benefit is high-gain compensation rather than parameter convergence (Section~\ref{sec:excitation}), and it remains when the RBF neurons are removed from the regressor.

An absolute quaternion reference model and an implicit projected weight update are given boundedness and damping results (Theorem~\ref{thm:1}, Corollary~\ref{cor:lambda}, Propositions~\ref{prop:implicit}--\ref{prop:reference}), each with its scope; together they do not form a stability proof for the full cascade. A software-in-the-loop check in the PX4 autopilot, with its estimator in the loop (Section~\ref{sec:sitl}), tests how the selected gain transfers to a flight stack.

The rest of the paper is organized as follows. Section~\ref{sec:related} reviews adaptive and learning-based quadrotor control, neural MRAC, measured-acceleration methods and composite learning. Section~\ref{sec:model} presents the Newton--quaternion quadrotor model and the control-oriented model with its uncertainty. Section~\ref{sec:design} describes the controller: the nominal laws and reference models, the NN regressors, the composite-learning law with realized inputs, its stability analysis and implementation, and the compared configurations with their common selection rule. Section~\ref{sec:results} reports the simulation results: the activity-constrained selection, deterministic tracking and the predictor mechanism, weight excitation, paired random-parameter tests, the activity budget and its sensitivities, and the PX4 software-in-the-loop check, followed by the limitations. Section~\ref{sec:concl} concludes the paper. Scenario and comparator details, full result tables, the motor-mismatch analysis, the measurement-path variant, the SITL details and the composite laws without RBF neurons are given in \suppref{}.

\section{Related Work}\label{sec:related}
\emph{Adaptive and learning-based quadrotor control.} Adaptive quadrotor control has been evaluated in flight \cite{dydek2013}, and $\mathcal{L}_1$ augmentation of geometric control decouples estimation from control through a low-pass filter \cite{wu2025l1quad}. MRAC has been combined with an NN compensator of parameter uncertainty \cite{glushchenko2023}; finite-time adaptive NN control has addressed actuator saturation and disturbances \cite{liu2021ast}, neural path-integral predictive control has been applied to quadrotor tracking \cite{li2025drones}, and deep reinforcement learning has beaten human champions in drone racing \cite{kaufmann2023champion}, a field surveyed in \cite{hanover2024racing}. The present design adapts a small set of output weights online and is compared under a common command-activity constraint.

\emph{Neural MRAC.} For robot manipulators, Lewis et al. derived NN weight laws that keep tracking errors and weights bounded \cite{lewis1996}. DMRAC adapts the output layer of a deep NN online \cite{joshi2019}, laws for all layers were given in \cite{patil2022}, and a state-predictor-based DMRAC tracks quadrotor trajectories \cite{cheng2025}. Neural-Fly learns a wind-invariant representation offline and adapts its last layer online with a composite law driven by the prediction error of an aerodynamic residual force inferred from onboard measurements \cite{oconnell2022}. The RNN self-tuning regulator (RNN-STR) of \cite{madebo2024} trains its controller through a learned plant identifier; the B1 implementation used here is defined in Section~\ref{sec:design}.

\emph{Measured-acceleration methods.} INDI inverts the dynamics incrementally from measured or filtered accelerations and the actuator state, so unmodeled forces and moments are rejected without an explicit model \cite{smeur2016,tal2021}; it has been combined with meta-learning in the allocation layer \cite{zhang2023metaindi} and serves as the inner loop in a comparison of nonlinear model predictive control (MPC) and differential-flatness control for agile flight \cite{sun2022comparative}. INDI implementations pass the acceleration and actuator signals through the same filter so that both carry the same delay \cite{smeur2016}. Wang et al. run an onboard actuator model inside INDI to improve robustness to model mismatch and to remove the oscillation that appears when the controller samples faster than the actuators \cite{wang2021onboard}, and learned INDI replaces the residual computation with a neural network so that rotor-speed sensors are not needed \cite{cobo2026lindi}. The target of this paper addresses a related time alignment, between the commanded input and the acceleration it produces, but uses it as the \emph{learning target} of the composite law rather than for inversion. INDI and $\mathcal{L}_1$ augmentation are included as comparators with the same realized-input signals (Section~\ref{sec:fair}).

\emph{Composite learning.} Composite adaptation drives the parameter update by both tracking and prediction errors \cite{slotine1989composite}. Combined/composite MRAC \cite{lavretsky2009}, concurrent learning \cite{chowdhary2010,chowdhary2013}, composite learning with interval excitation \cite{pan2016,pan2017} and composite MRAC under finite excitation \cite{cho2018}, surveyed in \cite{guo2023composite}, use stored or filtered data to obtain parameter convergence without persistent excitation, and closed-loop reference models (CRMs) suppress transient oscillations \cite{gibson2013}. In the filtered forms the same stable filter is applied to the regressor and to the input. For quadrotors, Manconi \emph{et al.} \cite{manconi2022} update neural compensation and a disturbance observer from state-estimation errors, so composite adaptation and filtered prediction errors predate this work. Accounting for known actuator dynamics in the model that is compared with measured accelerations is standard in incremental control \cite{smeur2016,tal2021,wang2021onboard}. The difference here is where the aligned signal is used: in the prediction target that updates the weights of a composite law, as a lag in the input branch only, separate from symmetric filtering of the target (Section~\ref{sec:extension}).

\begin{figure*}[!t]
\centering
\begin{tikzpicture}[>={Latex[length=1.5mm]}, font=\scriptsize, line width=0.45pt,
  node distance=2.8mm,
  blk/.style={draw, rounded corners=1.5pt, minimum height=9mm, align=center, fill=white, inner sep=1.6pt},
  ada/.style={blk, fill=blue!7, draw=blue!55!black},
  rin/.style={blk, fill=orange!12, draw=orange!70!black},
  plt/.style={blk, fill=gray!12},
  sum/.style={draw, circle, inner sep=0pt, minimum size=3.2mm, fill=white},
  sig/.style={font=\scriptsize, inner sep=1pt}]
\node[blk] (ref) {Reference\\ $\bv{p}_d,\dot{\bv{p}}_d,\ddot{\bv{p}}_d,\psi_d$};
\node[blk, right=of ref] (lqo) {Nominal outer law \eqref{eq:ublo}\\ $\bv{u}_{bl}=\ddot{\bv{p}}_d-\bv{K}_o(\bv{z}-\bv{z}_d)$};
\node[sum, right=2.5mm of lqo] (so) {};
\node[blk, right=4mm of so] (att) {Thrust/attitude\\ command\\ (tilt $\le35^\circ$)};
\node[blk, right=4mm of att] (lqi) {Nominal inner law\\ $\bv{\alpha}_{bl}=-\bv{K}_i[\bv{\epsilon};\bv{\omega}]$};
\node[sum, right=2.5mm of lqi] (si) {};
\node[blk, right=4mm of si] (mix) {$\bv{\tau}=\bv{J}_0\bv{\alpha}+\bv{\omega}\times \bv{J}_0\bv{\omega}$\\ $\bv{\Omega}_c=\sqrt{\mathrm{sat}(\bv{M}^{-1}[T;\bv{\tau}])}$};
\node[plt, right=8mm of mix] (plant) {Quadrotor\\ \eqref{eq:pos}--\eqref{eq:motor}\\ lag $\tau_m$, sat.};
\node[plt, right=of plant] (sens) {Sensors\\ $+$ noise};
\draw[->] (ref) -- (lqo);
\draw[->] (lqo) -- (so);
\draw[->] (so) -- node[sig,above]{$\bv{u}$} (att);
\draw[->] (att) -- node[sig,above]{$\bv{q}_d$} (lqi);
\draw[->] (lqi) -- (si);
\draw[->] (si) -- node[sig,above]{$\bv{\alpha}$} (mix);
\draw[->] (mix) -- node[sig,above,pos=0.3]{$\bv{\Omega}_c$} (plant);
\draw[->] (plant) -- (sens);
\draw[->] (att.north) -- ++(0,0.3) -| node[sig,pos=0.25,above]{$T$} (mix.north);
\node[ada, below=8mm of so] (nno) {Outer RBF-NN, RM \eqref{eq:crmo}\\ $\bv{u}_{ad}=-\hat{\bv{W}}_o^\top\bv{\phi}_o$};
\node[ada, below=8mm of si] (nni) {Inner RBF-NN, RM \eqref{eq:crmi}\\ $\bv{\alpha}_{ad}=-\hat{\bv{W}}_i^\top\bv{\phi}_i$};
\node[ada, below=6mm of nno] (clo) {Composite law \eqref{eq:law}, \eqref{eq:implicit}\\ $\bv{y}_f=H[\dot{\bv{v}}-\bv{u}_{\mathrm{eff}}]$};
\node[ada, below=6mm of nni] (cli) {Composite law \eqref{eq:law}, \eqref{eq:implicit}\\ $\bv{y}_f=H[\dot{\bv{\omega}}-\bv{\alpha}_{\mathrm{eff}}]$};
\draw[->] (nno) -- node[sig,right]{$\bv{u}_{ad}$} (so);
\draw[->] (nni) -- node[sig,right]{$\bv{\alpha}_{ad}$} (si);
\draw[->] (clo) -- node[sig,right]{$\hat{\bv{W}}_o$} (nno);
\draw[->] (cli) -- node[sig,right]{$\hat{\bv{W}}_i$} (nni);
\draw[->] (ref.south) |- ($(nno.west)+(0,0.17)$) node[sig,pos=0.8,above]{$\bv{z}_d$};
\draw[->] ($(att.south)+(0.4,0)$) |- ($(nni.west)+(0,0.17)$) node[sig,pos=0.6,above]{$\bv{q}_d$};
\coordinate (tap) at ($(mix.east)!0.5!(plant.west)$);
\node[rin, anchor=north] (mot) at (tap |- nni.north) {Actuator model: motor\\ $\tau_{m0}\dot{\hat{\bv{\Omega}}}=\bv{\Omega}_c-\hat{\bv{\Omega}}$, $[\hat T;\hat{\bv{\tau}}] =\bv{M}\hat{\bv{\Omega}}^{\circ2}$\\ or LPF $\tau_{\rm LP}\dot{\hat{\bv{\chi}}}=[T;\bv{\tau}]-\hat{\bv{\chi}}$};
\node[rin, anchor=north] (rin) at (tap |- cli.north) {Realized inputs \eqref{eq:ueff}\\ $\bv{u}_{\mathrm{eff}}=\bv{R}(\bv{q})\bv{e}_3\hat T/m_0-g\bv{e}_3$\\ $\bv{\alpha}_{\mathrm{eff}}=\bv{J}_0^{-1}(\hat{\bv{\tau}}-\bv{\omega}\times \bv{J}_0\bv{\omega})$};
\fill (tap) circle (0.7pt);
\draw[->] (tap) -- (mot.north);
\draw[->] (mot) -- node[sig,right]{$\hat T,\hat{\bv{\tau}}$} (rin);
\draw[->] (rin.west) -- node[sig,above]{$\bv{\alpha}_{\mathrm{eff}}$} (rin.west -| cli.east);
\draw[->] ($(rin.south)+(-0.3,0)$) -- ++(0,-0.4) coordinate (ue) -| (clo.south);
\node[sig, above left] at (ue) {$\bv{u}_{\mathrm{eff}}$: $\hat T$ (CL-Both) or $T$ (CL-Inner)};
\coordinate (bus) at ($(ue)+(0,-0.45)$);
\coordinate (busR) at (sens.south |- bus);
\coordinate (busL) at ($(lqo.south west)+(0.25,0)$);
\draw[line width=0.8pt] (sens.south) -- (busR) -- (busL |- bus);
\node[sig, anchor=north east] at (busR) {$\bv{y}=(\bv{p},\bv{v},\bv{q},\bv{\omega})+$noise};
\draw[->] (busL |- bus) -- (busL);
\coordinate (bi) at ($(lqi.south west)+(0.25,0)$);
\draw[->] (bi |- bus) -- (bi);
\foreach \tg in {nno, nni} {
  \coordinate (s\tg) at ($(\tg.west)+(-0.22,-0.17)$);
}
\coordinate (sclo) at ($(clo.west)+(-0.22,-0.17)$);
\coordinate (scli) at ($(cli.west)+(-0.22,-0.17)$);
\draw (snno) -- (sclo) -- (sclo |- bus);
\draw (snni) -- (scli) -- (scli |- bus);
\draw[->] (snno) -- ($(nno.west)+(0,-0.17)$);
\draw[->] (sclo) -- ($(clo.west)+(0,-0.17)$);
\draw[->] (snni) -- ($(nni.west)+(0,-0.17)$);
\draw[->] (scli) -- ($(cli.west)+(0,-0.17)$);
\coordinate (sr) at ($(rin.east)+(0.3,0)$);
\draw[->] (sr |- bus) -- node[sig,right,pos=0.6]{$\bv{q},\bv{\omega}$} (sr) -- (rin.east);
\foreach \pt in {busL, bi, sclo, scli, sr} \fill (\pt |- bus) circle (0.8pt);
\end{tikzpicture}
\caption{Controller architecture. White blocks: nominal cascade shared by all neural controllers; grey: plant and sensors. Blue: adaptive elements of both loops (reference models, RBF networks and composite laws). Orange: realized-input predictor of the proposed framework, i.e.\ actuator model and realized inputs \eqref{eq:ueff} from the measured attitude; CL-LPF replaces the motor-model block by a first-order filter on $[T;\bv{\tau}]$. CL-Both uses the predicted thrust $\hat T$ in the outer target and CL-Inner the commanded thrust $T$; both use the predicted torque $\hat{\bv{\tau}}$ in the inner target. CL-att uses commanded thrust/torque, MRAC and NN-MRAC have no composite law, and LQR has no adaptive element. Geometric control with a disturbance observer (GEO-DOB), INDI and $\mathcal{L}_1$ replace the adaptive elements by a disturbance estimate, an incremental inversion or an $\mathcal{L}_1$ predictor driven by the same realized inputs, keeping the thrust/attitude construction and allocation. CL-ucmd, CL-Outer and the B1 surrogate are described in Section~\ref{sec:fair} and reported in \suppref{}.}
\label{fig:arch}
\end{figure*}
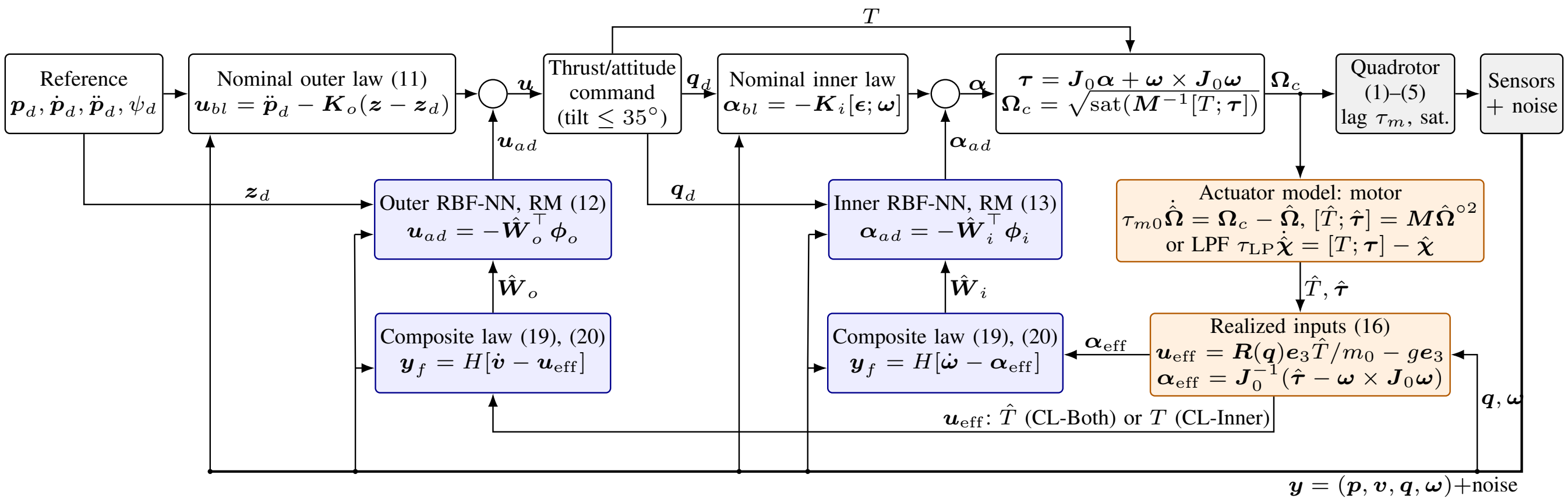

\section{System Modeling}\label{sec:model}

\begin{table}[!t]
\centering\setlength{\tabcolsep}{1.8pt}\caption{Quadrotor parameters and assumptions}\label{tab:params}
\renewcommand{\arraystretch}{1.1}\footnotesize
\begin{tabular}{@{}llc@{}}
\toprule
Parameter & Value & Source\\ \midrule
$m$ & 0.468\,kg & \cite{madebo2024}\\
$l$ & 0.225\,m & \cite{madebo2024}\\
$\Omega_{\max}$ & 1047\,rad/s (10\,000 rpm) & \cite{madebo2024}\\
$k$ & $2.98\times10^{-6}$\,N\,s$^2$ & \cite{luukkonen2011}\\
$b$ & $1.14\times10^{-7}$\,N\,m\,s$^2$ & \cite{luukkonen2011}\\
$\bv{J}$ & diag(4.856, 4.856, 8.801)$\times10^{-3}$\,kg\,m$^2$ & \cite{luukkonen2011}\\
$I_M$, $c_d$ & $3.357\times10^{-5}$\,kg\,m$^2$, 0.25\,kg/s & \cite{luukkonen2011}\\
$c_a$, $\tau_m$ & $10^{-3}$\,N\,m\,s, 20\,ms & assumed\\
Sensor noise (std) & $\bv{p}$: 1\,cm, $\bv{v}$: 2\,cm/s, & assumed\\
 & att.: 5\,mrad, gyro: 0.01\,rad/s & \\
Sampling, solver & 1\,kHz, fourth-order Runge--Kutta (RK4) & assumed\\ \bottomrule
\end{tabular}
\end{table}
\subsection{Newton--Quaternion Quadrotor Model}
Let $\mathcal{I}$ be an inertial frame with $\bv{e}_3$ pointing upwards and $\mathcal{B}$ the body
frame. The state of the simulation model is $\bv{x}=[\bv{p}^\top,\bv{v}^\top,\bv{q}^\top,\bv{\omega}^\top,\bv{\Omega}^\top]^\top
\in\mathbb{R}^{17}$, where $\bv{p},\bv{v}\in\mathbb{R}^3$ are the position and velocity in $\mathcal{I}$,
$\bv{q}=[q_0,\bv{q}_v^\top]^\top$ is the unit quaternion (scalar first) of the body attitude, $\bv{\omega}$ is the body
rate and $\bv{\Omega}\in\mathbb{R}^4$ the rotor speeds. As in the baseline \cite{madebo2024}, the
quaternion parameterization avoids the Euler-angle singularity:
\begin{align}
\dot {\bv{p}} &= \bv{v}, \label{eq:pos}\\
m\dot {\bv{v}} &= \bv{R}(\bv{q})\bv{e}_3 T - m g \bv{e}_3 - \bv{C}_d \bv{v} + \bv{d}_F(t), \label{eq:vel}\\
\dot {\bv{q}} &= \tfrac12\, \bv{q}\otimes[0,\ \bv{\omega}^\top]^\top, \label{eq:quat}\\
\bv{J}\dot{\bv{\omega}} &= \bv{\tau} - \bv{\omega}\times \bv{J}\bv{\omega} - \bv{C}_a\bv{\omega} - I_M\,(\bv{\omega}\times \bv{e}_3)\,\Omega_r + \bv{d}_\tau(t), \label{eq:rot}\\
\tau_m\dot\Omega_i &= \mathrm{sat}_{[0,\,\Omega_{\max}]}(\Omega_{c,i}) - \Omega_i,\quad i=1,\dots,4, \label{eq:motor}
\end{align}
with $\bv{R}(\bv{q})\in SO(3)$ the rotation matrix, $\Omega_r=\Omega_1-\Omega_2+\Omega_3-\Omega_4$ (gyroscopic term as in \cite{luukkonen2011}; the rotor spin-up torque $I_M\dot\Omega_r$ is neglected), drag matrices
$\bv{C}_d=\mathrm{diag}(c_{d,x},c_{d,y},c_{d,z})$, $\bv{C}_a=\mathrm{diag}(c_{a,x},c_{a,y},c_{a,z})$ and external force/torque disturbances
$\bv{d}_F$, $\bv{d}_\tau$. Thrust and torques follow the ``$+$'' allocation
\begin{equation}
\begin{bmatrix}T\\ \bv{\tau}\end{bmatrix}=
\underbrace{\begin{bmatrix} k & k & k & k\\ 0 & -lk & 0 & lk\\ -lk & 0 & lk & 0\\ b & -b & b & -b\end{bmatrix}}_{\bv{M}}
\begin{bmatrix}\Omega_1^2\\ \vdots\\ \Omega_4^2\end{bmatrix}. \label{eq:alloc}
\end{equation}
No controller knows the plant motor time constant $\tau_m$ in \eqref{eq:motor}, and every controller clips its rotor
commands at the plant limit $\Omega_{\max}$. The realized-input configurations, GEO-DOB, INDI and $\mathcal{L}_1$ use a nominal motor model with
$\tau_{m0}=20$\,ms (Section~\ref{sec:cl}), and the effect of $\tau_m\neq\tau_{m0}$ is studied in
Section~\ref{sec:budget}. Motor lag and rotor-speed clipping are included to make the evaluation more realistic
than in \cite{madebo2024}.
The parameters are listed in Table~\ref{tab:params}. The arm length and maximum rotor speed follow \cite{madebo2024}. The mass is 0.468\,kg, as in Table I of \cite{madebo2024}. The thrust and drag factors, the inertias and the translational drag are those of \cite{luukkonen2011}, which
reproduce the actuator limits reported in \cite{madebo2024} ($T_{\max}=13.07$\,N,
$\tau_{\psi,\max}=0.25$\,N\,m at $10\,000$\,rpm).

\subsection{Control-Oriented Model and Uncertainty}
Every controller uses the nominal mass $m_0$ and inertia $\bv{J}_0$; the realized-input configurations, GEO-DOB, INDI and $\mathcal{L}_1$ additionally use the nominal motor time constant $\tau_{m0}=20$\,ms. With the thrust vector $\bv{F}=m_0(\bv{u}+g \bv{e}_3)$ realized through the attitude, the
translational dynamics become
\begin{equation}
\begin{aligned}
\dot {\bv{v}} &= \bv{u} + \bv{d}_o,\qquad \bv{d}_o = \bv{d}_{o}^{\rm phys} + \bv{\delta}_{att},\\
\bv{d}_{o}^{\rm phys} &= \Big(\tfrac{m_0}{m}-1\Big)\big(\bv{u}_{\mathrm{eff}}^{\rm true}+g \bv{e}_3\big)-\tfrac{1}{m}\bv{C}_d \bv{v}+\tfrac{1}{m}\bv{d}_F,
\end{aligned}
\label{eq:transunc}
\end{equation}
where $\bv{u}\in\mathbb{R}^3$ is the commanded acceleration, $\bv{u}_{\mathrm{eff}}^{\rm true}$ the nominal-mass acceleration actually produced by the current attitude and true rotor speeds, and
$\bv{\delta}_{att}=\bv{u}_{\mathrm{eff}}^{\rm true}-\bv{u}$ the deficit caused by the attitude-tracking error and the motor lag.
With the torque command $\bv{\tau}_c=\bv{J}_0\bv{\alpha}+\bv{\omega}\times \bv{J}_0\bv{\omega}$, the rotational dynamics become
\begin{equation}
\begin{aligned}
\dot{\bv{\omega}} &= \bv{\alpha} + \bv{d}_i,\qquad \bv{d}_i = \bv{d}_{i}^{\rm phys} + \bv{\delta}_m,\\
\bv{d}_{i}^{\rm phys} &= (\bv{J}^{-1}\bv{J}_0-\bv{I})\bv{\alpha}_{\mathrm{eff}}^{\rm true} + \bv{J}^{-1}\big[\bv{\omega}\times \bv{J}_0\bv{\omega}-\bv{\omega}\times \bv{J}\bv{\omega}\\
&\qquad - \bv{C}_a\bv{\omega} - I_M(\bv{\omega}\times \bv{e}_3)\Omega_r + \bv{d}_\tau\big].
\end{aligned}
\label{eq:rotunc}
\end{equation}
The quantities $\bv{u}_{\mathrm{eff}}^{\rm true}=\bv{R}(\bv{q})\bv{e}_3T/m_0-g\bv{e}_3$ and $\bv{\alpha}_{\mathrm{eff}}^{\rm true}=\bv{J}_0^{-1}(\bv{\tau}-\bv{\omega}\times \bv{J}_0\bv{\omega})$ use the actual plant thrust and torque, whereas \eqref{eq:ueff} below uses model predictions, and $\bv{\delta}_m=\bv{\alpha}_{\mathrm{eff}}^{\rm true}-\bv{\alpha}$ is caused by motor lag and saturation. Both loops are written in the matched form

\begin{equation}
\dot {\bv{z}} = \bv{A} \bv{z} + \bv{B}\,\big(\bv{u} + \bv{d}^{\rm phys}(\bv{z},\bv{u}_{bl},t) + \bv{\delta}(t)\big), \label{eq:matched}
\end{equation}
with $\bv{A}=\begin{bmatrix}0&\bv{I}_3\\0&0\end{bmatrix}$, $\bv{B}=\begin{bmatrix}0\\ \bv{I}_3\end{bmatrix}$, $\bv{z}=[\bv{p}^\top,\bv{v}^\top]^\top$
(outer loop, $\bv{\delta}=\bv{\delta}_{att}$) or $\bv{z}=[\bv{\epsilon}^\top,\bv{\omega}^\top]^\top$ (inner loop, $\bv{\delta}=\bv{\delta}_m$),
where $\bv{\epsilon}$ is the attitude error defined below. The form is exact for the translational loop. For attitude it is a local design approximation, because quaternion error kinematics and the subtraction of rates expressed in different body frames do not give an exact double integrator. It is used for design with small attitude and rate errors; no stability result is given for the attitude loop, and Proposition~\ref{prop:reference} concerns only its reference model. Only the physical part is approximated by a
linear-in-the-weights neural network on a compact set,
\begin{equation}
\bv{d}^{\rm phys} = \bv{W}^{\ast\top}\bv{\phi}(\cdot)+\bv{\varepsilon},\qquad \|\bv{\varepsilon}\|\le\bar\varepsilon, \label{eq:nn}
\end{equation}
where $\bv{W}^\ast$ are unknown ideal weights and $\bv{\phi}$ is the regressor of Section~\ref{sec:nnreg}; the lag
term $\bv{\delta}$ is treated as a bounded disturbance, $\|\bv{\delta}\|\le\bar\delta$.

\begin{assumption}\label{ass:1}
(i) On the operating set, $\bv{d}^{\rm phys}$ is matched and bounded and admits the fixed-weight approximation \eqref{eq:nn}; time-dependent forces and moments are included in its bounded approximation residual, allowing gust steps. (ii) The reference and its first two derivatives are bounded on the finite simulation horizon; acceleration discontinuities are treated piecewise.
(iii) The inner loop is faster than the outer loop (time-scale separation), so that $\bv{\delta}_{att}$ is
bounded, and the rotor-speed limits are not reached persistently. (iv) The input gains are known
($m=m_0$, $\bv{J}=\bv{J}_0$) for the purpose of the analysis.
\end{assumption}

\noindent Assumption~\ref{ass:1}(iv) is restrictive: the input-gain error ($m_0/m$, $\bv{J}^{-1}\bv{J}_0$) is
represented only through the nominal command $\bv{u}_{bl}$ in the regressor, and its rigorous treatment is
\emph{not covered} by Theorem~\ref{thm:1}; the corresponding cases ($\bv{J}=2\bv{J}_0$, mass $+20$\,\%) are
examined in simulation. Applied thrust in the tracking dynamics scales with $m_0/m$, whereas $\bv{y}_f$ subtracts realized acceleration computed with the nominal mass; Corollary~\ref{cor:lambda} treats this mismatch for a constant mass ratio in the idealized translational loop, where it scales the composite damping and adds a bounded filter-commutation residual. Inertia errors, time-varying gains and the cascade are not covered, and a joint proof with parametric input-gain uncertainty remains future work. The allocation \eqref{eq:alloc} is written for the ``$+$'' configuration;
an ``$\times$'' configuration only changes the matrix $\bv{M}$.

\section{Controller Design}\label{sec:design}
\subsection{Architecture}
Fig.~\ref{fig:arch} shows the cascade. It is identical for all neural controllers, which differ only in
their adaptive elements and prediction targets; B1 and GEO-DOB replace the nominal and adaptive laws but
keep the thrust/attitude construction and allocation. The outer loop computes the acceleration command $\bv{u}$, from which the
thrust $T=\bv{F}^\top \bv{R}(\bv{q})\bv{e}_3$ and the desired attitude $\bv{q}_d$ are derived ($\bv{b}_{3d}=\bv{F}/\|\bv{F}\|$ and the
heading $\psi_d$, with a tilt limit of $35^\circ$). The inner loop computes the angular-acceleration
command $\bv{\alpha}$, the torque $\bv{\tau}=\bv{J}_0\bv{\alpha}+\bv{\omega}\times \bv{J}_0\bv{\omega}$ and, via $\bv{M}^{-1}$ in
\eqref{eq:alloc}, the rotor-speed commands.

\subsection{Nominal Laws and Reference Models}
Both loops use the same nominal law, an LQR for the double integrator per axis,
$\bv{K}=R^{-1}\bv{B}^\top \bv{X}$ with $\bv{Q}_o=\mathrm{diag}(16,0.01)$, $\bv{Q}_i=\mathrm{diag}(10^4,0.01)$, $R=1$. This
gives $\bv{K}_o=[4,\,2.83]$ and $\bv{K}_i=[100,\,14.14]$, i.e.\ $\omega_n=2$\,rad/s, $\zeta=0.71$ (outer;
rise time \val{an}{rmo}{tr}\,s, settling time \val{an}{rmo}{ts}\,s, overshoot \val{an}{rmo}{OS}\,\%)
and $\omega_n=10$\,rad/s, $\zeta=0.71$ (inner). The outer specification thus approximately matches the settling
time (3\,s) and overshoot (5\,\%) of the position reference model of \cite{madebo2024}.
For the linear cascade proxy (small angles, translational drag, first-order torque lag; rotational drag omitted) the inner loop has a phase margin of
\val{an}{in}{PM}$^\circ$ at \val{an}{in}{wc}\,rad/s (delay margin \val{an}{in}{DM}\,ms, sensitivity peak $M_s=\|(1+L)^{-1}\|_\infty=$~\val{an}{in}{Ms}, with $L$ the loop gain)
and the outer loop a gain margin of \val{an}{out}{GM}\,dB and a phase margin of \val{an}{out}{PM}$^\circ$
($M_s=$~\val{an}{out}{Ms}). The nominal linear proxy remains stable at the sampled parameter values for inertia scalings from
\val{an}{rob}{jmin} (lower sampled value) up to \val{an}{rob}{jmax} and for tested motor time constants up to \val{an}{rob}{tmax}\,ms (linearized analysis of the nominal LQR cascade), so
the parametric test $\bv{J}=2\bv{J}_0$ does not destabilize the nominal loop; it degrades tracking, which the
adaptive elements must recover. The linearized model predicts the nonlinear 1\,m step response
with a maximum deviation of \val{an}{step}{dev}\,mm.

\emph{Outer loop.} With $\bv{z}=[\bv{p};\bv{v}]$, $\bv{z}_d=[\bv{p}_d;\dot {\bv{p}}_d]$,
\begin{equation}
\bv{u}_{bl} = \ddot {\bv{p}}_d - k_{1o}(\bv{p}-\bv{p}_d) - k_{2o}(\bv{v}-\dot {\bv{p}}_d), \label{eq:ublo}
\end{equation}
and the reference model, written in the closed-loop form (CRM) of \cite{gibson2013}
\begin{equation}
\dot {\bv{z}}_m = \bv{A} \bv{z}_m + \bv{B}\big(\ddot {\bv{p}}_d - \bv{K}_o(\bv{z}_m-\bv{z}_d)\big) + \ell_o (\bv{z} - \bv{z}_m), \label{eq:crmo}
\end{equation}
with $\bv{z}_m(0)=\bv{z}(0)$. For $\ell_o=0$ \eqref{eq:crmo} is the classical open-loop reference model of MRAC; $\ell$ is
treated as a tuning option (Section~\ref{sec:fair}).

\emph{Inner loop.} The attitude error of a quaternion $\bv{q}_b$ with respect to $\bv{q}_a$ is
$\bv{\epsilon}(\bv{q}_a,\bv{q}_b)=2\,\operatorname{sgn}_{+}(\tilde q_0)\tilde {\bv{q}}_v$, $\tilde {\bv{q}}=\bv{q}_a^{-1}\otimes \bv{q}_b$, where $\operatorname{sgn}_{+}(x)=1$ for $x\ge0$ and $-1$ otherwise. This error equals the rotation vector to first order. The implementation chooses the positive-scalar quaternion representative, taking the positive branch at zero; the error is discontinuous at relative rotations of $180^\circ$. The nominal law is
$\bv{\alpha}_{bl}=-k_{1i}\bv{\epsilon}(\bv{q}_d,\bv{q})-k_{2i}\bv{\omega}$. The reference model is an \emph{absolute}
rotational model $(\bv{q}_m,\bv{\omega}_m)$ driven by the command $\bv{q}_d$:
\begin{equation}
\begin{aligned}
\dot {\bv{q}}_m &= \tfrac12 \bv{q}_m\otimes\big[0,\ (\bv{\omega}_m+\ell_i\,\bv{\epsilon}(\bv{q}_m,\bv{q}))^\top\big]^\top,\\
\dot {\bv{\omega}}_m &= -k_{1i}\bv{\epsilon}(\bv{q}_d,\bv{q}_m)-k_{2i}\bv{\omega}_m+\ell_i(\bv{\omega}-\bv{\omega}_m).
\end{aligned}\label{eq:crmi}
\end{equation}
Both the nominal plant controller and the absolute reference model receive $\bv{q}_d$. A command jump does not reset either attitude state and creates no direct jump in $\bv{\epsilon}(\bv{q}_m,\bv{q})$, although subsequent reference and plant dynamics can differ. The rate error $\bv{\omega}-\bv{\omega}_m$ is a local approximation because these rates refer to different body frames; an exact nonlinear proof would require frame transport and quaternion kinematics.

\subsection{Neural-Network Regressors}\label{sec:nnreg}
Each loop uses a linear-in-the-weights RBF network with bias and linear terms,
\begin{align}
\bv{\phi}_o &= \big[1,\ \bar {\bv{v}}^\top,\ \bar {\bv{u}}_{bl}^\top,\ \varphi_1(\bar {\bv{v}}),\dots,\varphi_8(\bar {\bv{v}})\big]^\top\in\mathbb{R}^{15},\\
\bv{\phi}_i &= \big[1,\ \bar{\bv{\omega}}^\top,\ \bar{\bv{\alpha}}_{bl}^\top,\ \bar\omega_2\bar\omega_3,\ \bar\omega_3\bar\omega_1,\ \bar\omega_1\bar\omega_2,\nonumber\\
&\qquad\ \varphi_1(\bar{\bv{\omega}}),\dots,\varphi_8(\bar{\bv{\omega}})\big]^\top\in\mathbb{R}^{18},
\end{align}
with normalized signals ($\bar {\bv{v}}=\bv{v}/3$, $\bar {\bv{u}}_{bl}=\bv{u}_{bl}/5$, $\bar{\bv{\omega}}=\bv{\omega}/3$,
$\bar{\bv{\alpha}}_{bl}=\bv{\alpha}_{bl}/20$) and Gaussian neurons $\varphi_j(\bv{s})=\exp(-\|\bv{s}-\bv{c}_j\|^2/2)$ centered at the
vertices of $[-1,1]^3$. The linear terms capture drag and the input-gain error, the products the
gyroscopic mismatch, the RBFs the remaining smooth nonlinearity. The adaptive terms are
$\bv{u}_{ad}=-\hat {\bv{W}}_o^\top\bv{\phi}_o$ and $\bv{\alpha}_{ad}=-\hat {\bv{W}}_i^\top\bv{\phi}_i$.

\subsection{Composite-Learning Adaptive Law With Realized Inputs}\label{sec:cl}
Let $\bv{e}=\bv{z}-\bv{z}_m$ and $\bv{P}=\bv{P}^\top>0$ solve $\bv{A}_m^\top \bv{P}+\bv{P}\bv{A}_m=-\bv{Q}_L$ with $\bv{A}_m=\bv{A}-\bv{B}\bv{K}$ (per axis, $\bv{Q}_L=\bv{I}_2$).
For three-axis expressions, $\bv{P}_3=\bv{P}\otimes \bv{I}_3$ and $\bv{B}_3=[0;\bv{I}_3]$; let $\tilde {\bv{W}}=\hat {\bv{W}}-\bv{W}^\ast$. Each loop uses an isotropic gain $\bv{\Gamma}=\gamma \bv{I}$, with its own scalar $\gamma>0$.
The controller runs the nominal motor model $\tau_{m0}\dot{\hat{\bv{\Omega}}}=\bv{\Omega}_c-\hat{\bv{\Omega}}$, from which
the realized thrust and torques $[\hat T;\hat{\bv{\tau}}]=\bv{M}\hat{\bv{\Omega}}^{\circ2}$ follow, and defines the
\emph{realized inputs}
\begin{equation}
\bv{u}_{\mathrm{eff}} = \frac{\bv{R}(\bv{q})\bv{e}_3\hat T}{m_0}-g\bv{e}_3,\qquad
\bv{\alpha}_{\mathrm{eff}} = \bv{J}_0^{-1}\big(\hat{\bv{\tau}}-\bv{\omega}\times \bv{J}_0\bv{\omega}\big), \label{eq:ueff}
\end{equation}
i.e.\ the accelerations produced by the \emph{measured} attitude and the \emph{predicted} rotor speeds. They are model-based estimates, not measured accelerations. They differ from $\bv{u}_{\mathrm{eff}}^{\rm true}$ and $\bv{\alpha}_{\mathrm{eff}}^{\rm true}$ when the predicted thrust and torques differ from the actual ones, for example under a motor-constant mismatch, and through noise in the measured attitude. These differences are not assumed to vanish: they enter the prediction residual $\bv{\xi}$ of Theorem~\ref{thm:1}, and Appendix~\sref{app:mismatch} bounds the part caused by the motor constant.
A derivative-free prediction error is built with a first-order filter $H(s)=1/(\tau_f s+1)$,
$\tau_f=50$\,ms:
\begin{equation}
\bv{y}_f = \frac{\bv{v} - \bv{v}_f}{\tau_f} - H[\bv{u}_{\mathrm{eff}}]\ \ \big(=H[\dot {\bv{v}} - \bv{u}_{\mathrm{eff}}]\big),\qquad
\bv{\phi}_f = H[\bv{\phi}], \label{eq:filt}
\end{equation}
with $\bv{v}_f=H[\bv{v}]$, and analogously $\bv{y}_f=(\bv{\omega}-\bv{\omega}_f)/\tau_f-H[\bv{\alpha}_{\mathrm{eff}}]$ in the inner
loop. By \eqref{eq:transunc}--\eqref{eq:rotunc}, $\bv{y}_f=H[\bv{d}^{\rm phys}]$ up to the motor-model error, i.e.\
the composite target removes $\bv{\delta}$ only with exact motor-state prediction and noiseless attitude, whereas a prediction error
built from the commanded inputs would equal $H[\bv{d}^{\rm phys}+\bv{\delta}]$. This does not imply that $\hat {\bv{W}}$
becomes a model of $\bv{d}^{\rm phys}$: the tracking-error term of \eqref{eq:law} also acts on $\hat {\bv{W}}$ and
dominates at large adaptation gains (Section~\ref{sec:extension}). Because $\hat{\bv{\Omega}}$ follows $\bv{\Omega}_c$ through a first-order lag, $\bv{u}_{\rm eff}$ and $\bv{\alpha}_{\rm eff}$ are, near an operating point, first-order low-pass filtered versions of the commanded inputs with constant $\tau_{m0}$. The velocity branch $(\bv{v}-\bv{v}_f)/\tau_f$ does not pass through this extra lag, so the filter delays the input branch toward the acceleration it produces and acts as a linear actuator-lag model. The ablation CL-LPF of Section~\ref{sec:extension} implements this filter directly on the commanded thrust and torques, $\tau_{\rm LP}\dot{\hat{\bv{\chi}}}=[T_c;\bv{\tau}_c]-\hat{\bv{\chi}}$ with $\hat{\bv{\chi}}=[\hat T_{\rm LP};\hat{\bv{\tau}}_{\rm LP}]$, so that $\bv{u}_{\rm eff}=\bv{R}(\bv{q})\bv{e}_3\hat T_{\rm LP}/m_0-g\bv{e}_3$ still uses the measured attitude and $\bv{\alpha}_{\rm eff}=\bv{J}_0^{-1}(\hat{\bv{\tau}}_{\rm LP}-\bv{\omega}\times \bv{J}_0\bv{\omega})$, and reaches the same admissible gain and design cost, whereas a larger shared constant $\tau_f$, which filters both branches equally, does not. The realized-input target is therefore used here as a lag-aligned target; its nonlinear motor model is not required, and the CL-LPF form needs only the commanded thrust and torques and one constant $\tau_{\rm LP}\approx\tau_{m0}$. The excitation matrices with
exponential forgetting $\lambda$,
\begin{equation}
\dot {\bv{Q}} = \lambda(\bv{\phi}_f\bv{\phi}_f^\top - \bv{Q}),\qquad \dot{\bv{\mathcal R}} = \lambda(\bv{\phi}_f \bv{y}_f^\top - \bv{\mathcal R}), \label{eq:QR}
\end{equation}
store the recent data, and the weights are updated by
\begin{equation}
\dot{\hat {\bv{W}}} = \mathrm{Proj}\Big\{\bv{\Gamma}\big[\bv{\phi}\, \bv{e}^\top \bv{P}_3 \bv{B}_3 - k_c\,(\bv{Q}\hat {\bv{W}}-\bv{\mathcal R})\big]\Big\}. \label{eq:law}
\end{equation}
In continuous time, $\mathrm{Proj}$ is the tangent-cone projection onto the ball $\{\|\bv{W}\|_F\le W_{\max}\}$: if $\|\hat{\bv{W}}\|_F=W_{\max}$ and the update $\bv{Y}$ satisfies $\langle\hat{\bv{W}},\bv{Y}\rangle_F>0$, it returns $\bv{Y}-\langle\hat{\bv{W}},\bv{Y}\rangle_F\hat{\bv{W}}/W_{\max}^2$, and otherwise $\bv{Y}$. For any $\bv{W}^\ast$ in the ball this gives $\langle\tilde{\bv{W}},\mathrm{Proj}\{\bv{Y}\}-\bv{Y}\rangle_F\le0$, the property used in Theorem~\ref{thm:1}; the discrete update \eqref{eq:implicit} below projects the new iterate instead. The first term is the MRAC (Lyapunov) term; the second is the composite-learning term
\cite{lavretsky2009,pan2016,cho2018}, $\bv{Q}\hat {\bv{W}}-\bv{\mathcal R}=\int_0^t\lambda e^{-\lambda(t-s)}\bv{\phi}_f
(\bv{\phi}_f^\top\hat {\bv{W}}-\bv{y}_f^\top)\,ds$, i.e.\ the filtered prediction error integrated over an exponentially weighted history. The tracking error contains the actual input deficit, whereas the prediction error contains the difference between true and modeled realized inputs; this is accounted for in Theorem~\ref{thm:1}.

Filters are initialized at zero; for a nonzero initial velocity the derivative-free identity includes the decaying initial-condition term $e^{-t/\tau_f}\bv{v}(0)/\tau_f$. The true initial velocities are zero here, but in the noisy scenarios the measured $\bv{v}(0)$ is not, and that term decays with time constant $\tau_f$. The discrete Euler filter is a sampled approximation, not an exact continuous-time convolution. Commands $\bv{u}$ before tilt limiting, the limited command $\bv{u}_{\rm lim}$, and the actual acceleration differ. The logged lumped estimate uses $\dot {\bv{v}}-\bv{u}_{\rm lim}$; the ideal tracking-error equation uses the pre-limit command and therefore also requires the tilt-limiting residual. In CL-ucmd the outer target subtracts $H[\bv{u}_{\rm lim}]$, so its residual contains the deficit $\bv{u}_{\rm eff}^{\rm true}-\bv{u}_{\rm lim}$, and the pre-limit tracking error contains $\bv{u}_{\rm lim}-\bv{u}$ as well.

\emph{Discrete implementation.} With $T_s=1$\,ms, the tracking injection, $\bv{Q}_k$ and $\bv{\mathcal R}_k$ are evaluated at $t_k$, and only the composite damping is taken at the new iterate:
\[
\hat{\bv{W}}^-_{k+1}=\hat{\bv{W}}_k+T_s\gamma\big[\bv{\phi}_k\bv{e}_k^\top\bv{P}_3\bv{B}_3-k_c(\bv{Q}_k\hat{\bv{W}}^-_{k+1}-\bv{\mathcal R}_k)\big].
\]
Collecting $\hat{\bv{W}}^-_{k+1}$ gives $(\bv{I}+T_s\gamma k_c\bv{Q}_k)\hat{\bv{W}}^-_{k+1}=\bv{X}_k$, whose matrix is positive definite whenever $\bv{Q}_k\succeq0$. The solution is then projected onto the weight ball:
\begin{equation}
\begin{aligned}
\bv{X}_k &= \hat {\bv{W}}_k+T_s\gamma(\bv{\phi}_k \bv{e}_k^\top \bv{P}_3\bv{B}_3+k_c\bv{\mathcal R}_k),\\
\hat {\bv{W}}_{k+1} &= \mathcal P\big[(\bv{I}+T_s\gamma k_c \bv{Q}_k)^{-1}\bv{X}_k\big].
\end{aligned}\label{eq:implicit}
\end{equation}
where $\mathcal P$ projects onto $\{\|\bv{W}\|_F\le W_{\max}\}$.
\begin{proposition}[Implicit composite damping]\label{prop:implicit}
For $\bv{Q}_k\succeq0$, $T_s>0$ and $\gamma k_c\ge0$, $\bv{M}_k=(\bv{I}+T_s\gamma k_c\bv{Q}_k)^{-1}$ satisfies $0\prec \bv{M}_k\preceq \bv{I}$. For a fixed $\bv{W}^\ast$ in the projection ball and an exact frozen composite target $\bv{\mathcal R}_k=\bv{Q}_k\bv{W}^\ast$, the damping-only update obeys
\begin{multline*}
\|\mathcal P[\bv{W}^\ast+\bv{M}_k(\hat {\bv{W}}_k-\bv{W}^\ast)]-\bv{W}^\ast\|_F\\
\le\|\bv{M}_k\|_2\|\hat {\bv{W}}_k-\bv{W}^\ast\|_F.
\end{multline*}
\end{proposition}
\begin{IEEEproof}
Diagonalize the symmetric $\bv{Q}_k$. The eigenvalues and singular values of $\bv{M}_k$ are $(1+T_s\gamma k_c\lambda_j(\bv{Q}_k))^{-1}\in(0,1]$. Projection onto a closed convex ball is non-expansive and fixes $\bv{W}^\ast$, which gives the inequality. Strict contraction requires $T_s\gamma k_c>0$ and $\bv{Q}_k\succ0$; semidefinite $\bv{Q}_k$ generally gives only non-expansiveness.
\end{IEEEproof}
For this frozen, exact-target step, the implicit form removes the explicit-Euler damping restriction $T_s\gamma k_c\lambda_{\max}(\bv{Q}_k)<2$. In \eqref{eq:implicit}, tracking injection and prediction residual remain forcing terms; the proposition does not prove unconditional stability of the complete update or sampled cascade. Explicit filters still restrict sampling, and Euler information updates preserve $\bv{Q}_k\succeq0$ when $0\le T_s\lambda\le1$.

\subsection{Stability Analysis}
A bound on the motor-model contribution to the residual $\bv{\xi}$ of Theorem~\ref{thm:1} for $\tau_m\neq\tau_{m0}$ is given in Appendix~\sref{app:mismatch}, Proposition~\sref{prop:mismatch}.

\begin{theorem}\label{thm:1}
Consider the ideal, continuous-time translational matched loop with known input gain, bounded regressor, fixed ideal weights and an initial estimate $\hat{\bv{W}}(0)$ inside the projection ball, bounded approximation/input residual $\bv{r}=\bv{\varepsilon}+\bv{\delta}$, and bounded prediction residual $\bv{\xi}$ satisfying $\bv{Q}\hat {\bv{W}}-\bv{\mathcal R}=\bv{Q}\tilde {\bv{W}}+\bv{\xi}$. For $\bv{\Gamma}=\gamma \bv{I}$, $\gamma>0$, $\ell\ge0$ and $k_c\ge0$, the reference model \eqref{eq:crmo} and projected law \eqref{eq:law} yield ultimately bounded weights and tracking errors. If additionally $k_c>0$, then on any interval on which $\bv{Q}(t)\succeq\sigma \bv{I}$, $\sigma>0$, the Lyapunov function contracts exponentially to a residual bound. This statement does not establish stability of the nonlinear attitude loop or the sampled cascade.
\end{theorem}
\begin{IEEEproof}
Let $\bv{P}_3=\bv{P}\otimes \bv{I}_3$, $\bv{B}_3=[0;\bv{I}_3]$, $a=\lambda_{\min}(\bv{I}_6+2\ell \bv{P}_3)>0$, and
$V=\bv{e}^\top \bv{P}_3\bv{e}+\gamma^{-1}\|\tilde {\bv{W}}\|_F^2$. Projection and cancellation of the tracking cross term give
\begin{equation}
\begin{aligned}
\dot V\le{}&-a\|\bv{e}\|^2+2\|\bv{P}_3\bv{B}_3\|\bar r\|\bv{e}\|\\
&-2k_c\operatorname{tr}(\tilde {\bv{W}}^\top \bv{Q}\tilde {\bv{W}})+2k_c\|\tilde {\bv{W}}\|_F\bar\xi.
\end{aligned}
\end{equation}
Without excitation, projection bounds $\tilde {\bv{W}}$; the Hurwitz error dynamics driven by bounded $\tilde {\bv{W}}^\top\bv{\phi}$ and $\bv{r}$ imply ultimate boundedness of $\bv{e}$. If $k_c>0$ and $\bv{Q}\succeq\sigma \bv{I}$, Young's inequality yields
\begin{equation}
\dot V\le-\tfrac a2\|\bv{e}\|^2-k_c\sigma\|\tilde {\bv{W}}\|_F^2+c,
\quad c=\frac{2\|\bv{P}_3\bv{B}_3\|^2\bar r^2}{a}+\frac{k_c\bar\xi^2}{\sigma}.
\end{equation}
Thus $\dot V\le-\kappa V+c$ with
$\kappa=\min\{a/(2\lambda_{\max}(\bv{P}_3)),\gamma k_c\sigma\}$, and
$V(t)\le e^{-\kappa(t-t_0)}V(t_0)+(c/\kappa)(1-e^{-\kappa(t-t_0)})$
while the lower bound holds. Absolute plant position need not be bounded on an infinite climbing reference; the assertion concerns errors and controller states on the operating horizon.
\end{IEEEproof}
Theorem~\ref{thm:1} bounds the error to the reference-model state $\bv{z}_m$. The gap $\bv{h}=\bv{z}_m-\bv{z}_d$ obeys $\dot{\bv{h}}=(\bv{A}_m\otimes\bv{I}_3)\bv{h}+\ell\bv{e}$ and is therefore bounded whenever $\bv{e}$ is. The implemented information matrix has exponential forgetting. A finite exciting interval alone does not guarantee a positive lower bound thereafter: for a constant filtered regressor $\bv{\phi}_0$, $\bv{Q}(t)=e^{-\lambda(t-t_0)}\bv{Q}(t_0)+(1-e^{-\lambda(t-t_0)})\bv{\phi}_0\bv{\phi}_0^\top$. Information in directions orthogonal to $\bv{\phi}_0$ decays exponentially. Unlike the retained adequate information matrix in \cite{cho2018}, no history-selection mechanism is implemented here. In the reported runs, $\bv{Q}$ stays far from full rank and no useful uniform lower bound $\bv{Q}\succeq\sigma \bv{I}$ with $\sigma>0$ was found (Section~\ref{sec:excitation}). The runs therefore do not establish a uniform positive excitation bound, and the contraction part of Theorem~\ref{thm:1} cannot be invoked for them; the boundedness part holds only for the idealized continuous-time translational loop of its assumptions, not for the sampled cascade. Gaussian sensor noise is unbounded in distribution, so the deterministic bounded-residual result is not a stochastic stability guarantee. Motor-model mismatch, sensor noise, filter initialization and discretization contribute to $\bv{\xi}$; it cannot generally be declared independent of actuator lag. A joint nonlinear cascade proof remains open.

The known-gain assumption of Theorem~\ref{thm:1} can be relaxed for a constant mass error in the same idealized translational loop.
\begin{corollary}[Unknown constant input gain]\label{cor:lambda}
In the setting of Theorem~\ref{thm:1}, let the specific-force gain be an unknown constant $\Lambda=m_0/m\in[\Lambda_{\min},\Lambda_{\max}]\subset(0,\infty)$, so that $\dot {\bv{v}}=\Lambda(\bv{u}+g\bv{e}_3)-g\bv{e}_3+\bv{d}$ with $\bv{u}=\bv{u}_{bl}-\hat {\bv{W}}^\top\bv{\phi}$, and let the realized input of the target be $\bv{u}_{\rm eff}=\bv{u}$. Assume ideal weights $\bv{W}^\ast$ in the projection ball with $\Lambda \bv{W}^{\ast\top}\bv{\phi}=(\Lambda-1)(\bv{u}_{bl}+g\bv{e}_3)+\bv{d}-\Lambda\bv{\varepsilon}$, $\|\bv{\varepsilon}\|\le\bar\varepsilon$; the bias and the normalized command $\bar {\bv{u}}_{bl}=\bv{u}_{bl}/5$ in $\bv{\phi}_o$ make such weights exist whenever $\bv{d}$ is representable, but whether they lie in the ball depends on $W_{\max}$, so this is an assumption. Then, with $V_\Lambda=\bv{e}^\top \bv{P}_3\bv{e}+\Lambda\gamma^{-1}\|\tilde {\bv{W}}\|_F^2$, the conclusions of Theorem~\ref{thm:1} hold with $\bv{r}$ replaced by $\Lambda\bv{\varepsilon}$ and $\bv{\xi}$ replaced by $\bv{\xi}+\bv{\xi}_\Lambda$. On an interval with $\bv{Q}\succeq\sigma\bv{I}$ the decay rate is at least $\min\{a/(2\lambda_{\max}(\bv{P}_3)),\,\gamma\Lambda_{\min}k_c\sigma\}$, and
\[
\|\bv{\xi}_\Lambda\|_F\le4|\Lambda-1|\,W_{\max}\bar\phi\,\Phi_f ,
\]
$\bar\phi$ bounds $\|\bv{\phi}\|_2$ and $\Phi_f\le\bar\phi$ bounds $\|\bv{\phi}_f\|_2$ (the filter $H$ has unit induced $L_\infty$ gain).
\end{corollary}
\begin{IEEEproof}
Substituting $\bv{W}^\ast$ gives $\dot {\bv{e}}=\bv{A}_m\bv{e}-\Lambda \bv{B}_3(\tilde {\bv{W}}^\top\bv{\phi}-\bv{\varepsilon})$ for the reference model \eqref{eq:crmo} with $\ell=0$; the CRM term enters as in Theorem~\ref{thm:1}. The tracking term of \eqref{eq:law} contributes $2\Lambda\operatorname{tr}(\tilde {\bv{W}}^\top\bv{\phi} \bv{e}^\top \bv{P}_3\bv{B}_3)$ to $\dot V_\Lambda$, which cancels the cross term $-2\Lambda \bv{e}^\top \bv{P}_3\bv{B}_3\tilde {\bv{W}}^\top\bv{\phi}$. For the target, $\dot {\bv{v}}-\bv{u}=(\Lambda-1)(\bv{u}+g\bv{e}_3)+\bv{d}=\Lambda \bv{W}^{\ast\top}\bv{\phi}+\Lambda\bv{\varepsilon}-(\Lambda-1)\hat {\bv{W}}^\top\bv{\phi}=\bv{W}^{\ast\top}\bv{\phi}-(\Lambda-1)\tilde {\bv{W}}^\top\bv{\phi}+\Lambda\bv{\varepsilon}$. The filter maps $\bv{W}^{\ast\top}\bv{\phi}$ to $\bv{W}^{\ast\top}\bv{\phi}_f$ because $\bv{W}^\ast$ is constant, whereas $H[\tilde {\bv{W}}^\top\bv{\phi}](s)$ depends on the weight history up to $s$. Writing $\hat {\bv{W}}(t)=\bv{W}^\ast+\tilde {\bv{W}}(t)$ inside $\bv{Q}(t)\hat {\bv{W}}(t)$ gives $\bv{Q}\hat {\bv{W}}-\bv{\mathcal R}=\Lambda \bv{Q}\tilde {\bv{W}}(t)+\bv{\xi}+\bv{\xi}_\Lambda$, where $\bv{\xi}$ collects the terms of Theorem~\ref{thm:1} and
\begin{multline*}
\bv{\xi}_\Lambda(t)=(\Lambda-1)\int_0^t\lambda e^{-\lambda(t-s)}\bv{\phi}_f(s)\\
\times\big(H[\tilde {\bv{W}}^\top\bv{\phi}](s)-\tilde {\bv{W}}(t)^\top\bv{\phi}_f(s)\big)^\top ds
\end{multline*}
is the commutation error of the filter. Both terms in the bracket are bounded by $2W_{\max}\bar\phi$, since $\|\tilde {\bv{W}}\|_F\le2W_{\max}$ and $\Phi_f\le\bar\phi$, and the kernel mass is at most one, which gives the stated bound. The composite term then contributes $-2\Lambda^2k_c\operatorname{tr}(\tilde {\bv{W}}^\top \bv{Q}\tilde {\bv{W}})+2\Lambda k_c\|\tilde {\bv{W}}\|_F(\bar\xi+\bar\xi_\Lambda)$, and the remaining steps of the proof of Theorem~\ref{thm:1} apply with the stated substitutions. With $\bv{Q}\succeq\sigma\bv{I}$, the Young inequality gives
\begin{multline*}
\dot V_\Lambda\le-\tfrac a2\|\bv{e}\|^2-\Lambda^2k_c\sigma\|\tilde {\bv{W}}\|_F^2\\
+\frac{2\|\bv{P}_3\bv{B}_3\|^2\Lambda^2\bar\varepsilon^2}{a}+\frac{k_c(\bar\xi+\bar\xi_\Lambda)^2}{\sigma}.
\end{multline*}
Because $V_\Lambda$ weights $\|\tilde {\bv{W}}\|_F^2$ by $\Lambda\gamma^{-1}$, the weight part decays at rate $\gamma\Lambda k_c\sigma\ge\gamma\Lambda_{\min}k_c\sigma$, and using $\Lambda_{\max}$ in the $\bar\varepsilon$ term makes the residual bound uniform.
\end{IEEEproof}
The corollary keeps the idealizations of Theorem~\ref{thm:1}: realized and commanded inputs coincide in the target, attitude is tracked exactly, and the bound concerns the translational loop only. Time-varying parts of $\bv{d}$ that fixed weights cannot represent remain in $\bv{\varepsilon}$. A mass error therefore enlarges the residual through $\bv{\xi}_\Lambda$ and scales the composite damping by $\Lambda$; it does not change the sign of either term.

\begin{proposition}[Absolute reference-model boundedness]\label{prop:reference}
Consider \eqref{eq:crmi} with $k_{1i}>0$, $k_{2i}>0$, $\ell_i\ge0$, unit $\bv{q}_m(0)$ and bounded measured $\|\bv{\omega}\|\le\bar\omega$. Since $\|\bv{\epsilon}(\bv{q}_a,\bv{q}_b)\|\le2$, the continuous reference satisfies $\|\bv{q}_m(t)\|=1$ and
\[
\begin{aligned}
\|\bv{\omega}_m(t)\|&\le e^{-at}\|\bv{\omega}_m(0)\|+C(1-e^{-at}),\\
C&=\frac{2k_{1i}+\ell_i\bar\omega}{a},\quad a=k_{2i}+\ell_i.
\end{aligned}
\]
\end{proposition}
\begin{IEEEproof}
Quaternion kinematics preserve norm. Rewrite the rate equation as $\dot{\bv{\omega}}_m=-a\bv{\omega}_m+\bv{f}$, with $\|\bv{f}\|\le2k_{1i}+\ell_i\bar\omega$, and apply variation of constants. For $\tilde {\bv{q}}_m=\bv{q}_d^{-1}\otimes \bv{q}_m$, constant $\bv{q}_d$ and $\ell_i=0$, away from the sign-switch surface the potential
$V_m=4k_{1i}(1-|\tilde q_{m0}|)+\frac12\|\bv{\omega}_m\|^2$
satisfies $\dot V_m=-k_{2i}\|\bv{\omega}_m\|^2$ because the implemented attitude error is $2\,\mathrm{sgn}_+(\tilde q_{m0})\tilde {\bv{q}}_{mv}$.
\end{IEEEproof}
This is a bounded-state/rate result under bounded measured rate, not a global ISS theorem for moving attitude commands or the switching quaternion error. The potential $V_m$ shows only that the reference model does not gain energy for a constant $\bv{q}_d$ with $\ell_i=0$ away from the sign-switch surface; the proposition gives boundedness of $\bv{q}_m$ and $\bv{\omega}_m$, not asymptotic stability of the attitude-tracking error. Numerical quaternion renormalization, sampling and the CRM coupling do not inherit an unproved continuous-time cascade guarantee.

\subsection{Algorithm}
At every sampling instant $t_k$: (1) read $\bv{p},\bv{v},\bv{q},\bv{\omega}$; (2) evaluate $\bv{p}_d,\dot {\bv{p}}_d,\ddot {\bv{p}}_d,\psi_d$;
(3) compute $\bv{u}_{bl}$, $\bv{\phi}_o$, $\bv{u}=\bv{u}_{bl}-\hat {\bv{W}}_o^\top\bv{\phi}_o$, the tilt-limited thrust vector, $T$ and
$\bv{q}_d$; (4) compute $\bv{\epsilon}(\bv{q}_d,\bv{q})$, $\bv{\alpha}_{bl}$, $\bv{\phi}_i$, $\bv{\alpha}=\bv{\alpha}_{bl}-\hat {\bv{W}}_i^\top\bv{\phi}_i$,
$\bv{\tau}$ and $\bv{\Omega}_c=\sqrt{\mathrm{sat}(\bv{M}^{-1}[T;\bv{\tau}])}$; (5) use the current motor-model state to compute
$\bv{u}_{\mathrm{eff}}$, $\bv{\alpha}_{\mathrm{eff}}$ \eqref{eq:ueff}; (6) update both weight matrices with
\eqref{eq:implicit}, the filters \eqref{eq:filt}, the matrices \eqref{eq:QR} and the reference models \eqref{eq:crmo}, \eqref{eq:crmi}, then propagate the motor model with the current clipped rotor command. The computational load is dominated by the $18\times18$ linear solve
of the inner implicit update (execution times in Section~\ref{sec:effort}).

\subsection{Compared Configurations and Selection}\label{sec:fair}
All controllers share the plant, sensors, nominal gains and actuator limits. The proposed framework has two primary configurations: \textbf{CL-Inner} uses measured attitude and commanded thrust in the outer target, with predicted motor torque in the inner target; \textbf{CL-Both} uses the motor predictor in both targets. \textbf{CL-Outer} reverses CL-Inner's predictor switches. \textbf{CL-att} uses measured attitude but commanded thrust/torque in both targets. \textbf{CL-LPF} replaces the motor model in both targets by a linear first-order filter with constant $\tau_{\rm LP}$ applied to the commanded thrust and torques, so its target needs no rotor-speed map and no allocation inverse. All five retain composite updates in both loops.

The remaining comparators are \textbf{LQR} (nominal laws), \textbf{MRAC} (linear regressor, no composite term), \textbf{NN-MRAC} (RBF regressor, no composite term), and \textbf{CL-ucmd} (commanded-input composite targets, with the tilt-limited command $\bv{u}_{\rm lim}$ in the outer target). \textbf{B1} is an imitation-trained FFNN/Elman-RNN surrogate inspired by \cite{madebo2024}, not a faithful reproduction of its plant-identifier-based RNN-STR. Its architecture and training are documented in the supplementary package.

\textbf{GEO-DOB} uses $\bv{e}_R=\frac12(\bv{R}_d^\top \bv{R}-\bv{R}^\top \bv{R}_d)^\vee$ and $\bv{e}_\omega=\bv{\omega}-\bv{R}^\top \bv{R}_d\bv{\omega}_d$, with transported reference-rate feedforward \cite{lee2010geometric}. Its derivative-free disturbance estimate is $(\bv{v}-\bv{v}_f)/\tau-\bv{u}_f$, where $\dot {\bv{v}}_f=(\bv{v}-\bv{v}_f)/\tau$ and $\dot {\bv{u}}_f=(\bv{u}_{\rm eff}-\bv{u}_f)/\tau$, with an analogous angular-rate estimate. Estimates are clipped at 5 m/s$^2$ and 50 rad/s$^2$. It uses the same measured attitude and nominal motor predictions as the realized-input configurations and never the true acceleration, disturbance or rotor state.
With proportional scales $\kappa_p,\kappa_a$ and compensation gain $\beta$, the comparator commands are
\begin{align}
\bv{u}_G&=\ddot {\bv{p}}_d-\kappa_p k_{1o}(\bv{p}-\bv{p}_d)
-\sqrt{\kappa_p}k_{2o}(\bv{v}-\dot {\bv{p}}_d)-\beta\hat {\bv{d}}_o,\\
\bv{\alpha}_G&=-\kappa_a k_{1i}\bv{e}_R-\sqrt{\kappa_a}k_{2i}\bv{e}_\omega
-\bv{\omega}\times \bv{R}^\top \bv{R}_d\bv{\omega}_d \nonumber\\
&\quad+\bv{R}^\top \bv{R}_d\dot{\bv{\omega}}_d-\beta\hat {\bv{d}}_i,\qquad
\bv{\tau}_G=\bv{J}_0\bv{\alpha}_G+\bv{\omega}\times \bv{J}_0\bv{\omega} .
\end{align}
The thrust/attitude construction, tilt limit and allocation are the same as for the neural controllers; the discrete desired-rate filter that defines the comparator is given in Appendix~\sref{app:comp}.

\textbf{INDI} applies incremental nonlinear dynamic inversion \cite{smeur2016,tal2021} in both loops, with the thrust/attitude construction, tilt limit, allocation, measured attitude and nominal motor predictor of GEO-DOB. Accelerations are not measured: $\bv{a}_f=(\bv{v}-\bv{v}_f)/\tau$ and $\dot{\bv{\omega}}_f=(\bv{\omega}-\bv{\omega}_f)/\tau$, and the realized specific force $\bv{u}_{\rm eff}$ and the predicted torque $\hat{\bv{\tau}}$ pass through the same first-order filter $H$. The commands are
\begin{equation}
\bv{u}_I=H[\bv{u}_{\rm eff}]+\bv{\nu}_o-\bv{a}_f,\qquad \bv{\tau}_I=H[\hat{\bv{\tau}}]+\bv{J}_0(\bv{\nu}_i-\dot{\bv{\omega}}_f),\label{eq:indi}
\end{equation}
with the scaled LQR virtual controls $\bv{\nu}_o=\ddot {\bv{p}}_d-\kappa_pk_{1o}(\bv{p}-\bv{p}_d)-\sqrt{\kappa_p}k_{2o}(\bv{v}-\dot {\bv{p}}_d)$ and $\bv{\nu}_i=-\kappa_ak_{1i}\bv{\epsilon}-\sqrt{\kappa_a}k_{2i}\bv{\omega}$. Because the filters are synchronized, $\bv{u}_I=\bv{\nu}_o-\hat {\bv{d}}_o$ with the derivative-free estimate of GEO-DOB; in the outer loop INDI is therefore a unit-gain disturbance observer on the quaternion LQR cascade, without estimate clipping and without geometric feedforward. Its grid is $\kappa_p\in\{0.5,1,2,4\}$, $\kappa_a\in\{0.25,0.5,1,2\}$ and $\tau\in\{10,20,50,100,200\}$\,ms (80 candidates in three dimensions; Section~\ref{sec:tuning} adds 27 refinement candidates). \textbf{INDI-2nd} replaces the first-order filter by the second-order low-pass $\omega_c^2/(s^2+2\zeta\omega_c s+\omega_c^2)$, $\zeta=0.707$, applied identically to $\bv{v}$, $\bv{\omega}$, $\bv{u}_{\rm eff}$ and $\hat{\bv{\tau}}$, whose second state gives $\bv{a}_f$ and $\dot{\bv{\omega}}_f$, as in \cite{smeur2016}; its grid is the 16 $(\kappa_p,\kappa_a)$ pairs of the INDI grid with $f_c=\omega_c/2\pi\in\{0.5,1,2,4,8,16\}$\,Hz (96 candidates).

{\bfseries $\mathcal{L}_1$} augments the same LQR cascade with a piecewise-constant augmentation adapted from the $\mathcal{L}_1$ architecture of \cite{wu2025l1quad}. State predictors $\dot{\hat {\bv{v}}}=\bv{u}_{\rm eff}+\hat{\bv{\sigma}}_o-a_o(\hat {\bv{v}}-\bv{v})$ and $\dot{\hat{\bv{\omega}}}=\bv{\alpha}_{\rm eff}+\hat{\bv{\sigma}}_i-a_i(\hat{\bv{\omega}}-\bv{\omega})$ are driven by the realized inputs \eqref{eq:ueff}; the estimate $\hat{\bv{\sigma}}=-a\,e^{-aT_s}(1-e^{-aT_s})^{-1}(\hat {\bv{x}}-\bv{x})$ is held over each period, and the compensation is $\bv{u}=\bv{\nu}_o-C_o(s)\hat{\bv{\sigma}}_o$, $\bv{\alpha}=\bv{\nu}_i-C_i(s)\hat{\bv{\sigma}}_i$ with $C(s)=\omega_c/(s+\omega_c)$. The grid is $a_o\in\{10,40\}$\,s$^{-1}$ with $a_i=5a_o$, $\omega_{c,o}\in\{2,5,10,20\}$\,rad/s, $\omega_{c,i}\in\{10,20,50,100\}$\,rad/s and a common nominal-gain scale in $\{0.5,1,2\}$ (96 candidates in four dimensions; Section~\ref{sec:tuning} adds 72 refinement candidates). The baseline law is the quaternion LQR cascade rather than geometric control, and no performance bound is claimed for this sampled implementation.

The neural-family gain grid is $\gamma_o\in\{5,10,20,\ldots,5120\}$, $\gamma_i=10\gamma_o$; composite families test $\ell_o\in\{0,5\}$, $\ell_i=5\ell_o$, $k_c\in\{0.25,0.5,1\}$ and $\lambda\in\{0.5,2\}$ s$^{-1}$. The shared cost is the mean root-mean-square (RMS) position error RMSE$_5$ (Section~\ref{sec:clean}) on S2--S4, constrained by full-run S7 activity. MRAC, NN-MRAC and each composite family have 11, 22 and 132 candidates, respectively. GEO-DOB has 1074 candidates in five dimensions. Different comparator architectures have different tuning dimensions; every family uses the same cost, constraint and seed. Selection conditions and the empirical multi-seed guard are specified together below.

\section{Simulation Results and Discussion}\label{sec:results}
\subsection{Protocol, Scenarios and Metrics}\label{sec:clean}
The plant is integrated by fixed-step RK4 at 1 kHz, with sampled disturbances/parameters held over each integration step; control is at 1 kHz unless stated otherwise. Table~\ref{tab:scen} defines S1--S10; details are in Appendix~\sref{app:scen}. Only S2--S4 tracking costs and the S7 activity condition enter selection. S8--S10 do not enter tuning or B1 training; B1 training uses S1, S2 and S6, so those are not all held out for B1. Sequential comparator refinements and reuse of the same random-parameter draws are documented in \suppref{} (see Section~\ref{sec:limits}).

\begin{table}[!t]
\centering\caption{Simulation scenarios (D: design, V: evaluation, C: constraint)}\label{tab:scen}
\renewcommand{\arraystretch}{1.08}\footnotesize
\begin{tabular}{@{}>{\raggedright\arraybackslash}p{0.75cm}>{\raggedright\arraybackslash}p{3.1cm}>{\raggedright\arraybackslash}p{3.6cm}@{}}
\toprule
 & Trajectory & Changes / disturbances / noise\\ \midrule
S1 V & helix of \cite{madebo2024} & nominal\\
S2 D & photography path, 9\,m, 2\,m/s & nominal\\
S3 D & helix & $\bv{J}=2\bv{J}_0$, $c_d=(1+0.1t)c_{d0}$, $c_a=(1+0.01t)c_{a0}$ \cite{madebo2024}\\
S4 D & photography path & forces/torques with $\sin3t$, $\sin0.03t$ \cite{madebo2024}, gust at 10\,s\\
S5 V & photography path & S3 changes, $m=1.2m_0$, S4 disturbances, noise\\
S6 V & 1\,m set-point step & nominal\\
S7 C & noisy set-point & 1\,m step on each axis; activity constraint\\
S8 V & figure-eight, 5\,m, yaw $0.3\sin0.2t$ & $m=1.1m_0$, $\bv{J}=1.5\bv{J}_0$, $1.5\times$ dist., $\sin2t$, gust at 20\,s ($-$), noise\\
S9 V & photography path & $m=0.9m_0$, $\bv{J}=1.5\bv{J}_0$, $c_d=2c_{d0}$, $1.5\times$ dist., $\sin5t$, gust at 25\,s ($-$), noise\\
S10 V & fast helix, $r=2$\,m, 0.6\,rad/s & $m=1.15m_0$, $\bv{J}=2.5\bv{J}_0$, growing drag, $0.8\times$ dist., $\sin4t$, gust at 15\,s, noise\\ \bottomrule
\end{tabular}
\end{table}

The principal error is the three-dimensional RMS position error after the first 5\,s,
\[
\mathrm{RMSE}_5=\Big(N_5^{-1}\textstyle\sum_{t_k>5\,\mathrm{s}}\|\bv{p}(t_k)-\bv{p}_d(t_k)\|^2\Big)^{1/2},
\]
computed from the true plant position on the 10\,ms log grid, with $N_5$ samples. Command activity $\Delta\Omega=\big(\sum_{k,i}(\Omega_{c,i}(k)-\Omega_{c,i}(k-1))^2/(4N)\big)^{1/2}$ is the RMS sample-to-sample rotor-command change at the 1\,kHz control rate over the complete run, with $N$ increments, including transients; S7 is a noisy step, not stationary hover. Total variation per second is $\mathrm{TV}=T^{-1}\sum_{k,i}|\Omega_{c,i}(k)-\Omega_{c,i}(k-1)|$. Both activity metrics depend on sampling and sensor noise. Every controller is scored by the same code, RMSE$_5$ from the logged plant position and $\Delta\Omega$ from the rotor commands sent to the plant, so the budget constrains the same signal for all controllers and not a controller's own estimate of its effort. The limit of 1.05$\times$LQR activity in S7 is a design choice that serves as a proxy for command chatter; it is not imposed on maneuver activity.
\val{sl}{all}{n} matched-prefix MATLAB--Simulink comparisons (S2/S5) of seven controllers (LQR, MRAC, B1, NN-MRAC, CL-ucmd, CL-att and seed-1 CL-Both) have maximum position difference $\val{sl}{all}{maxdp}$ m; the guarded profiles, CL-LPF and the non-neural comparators were not run in Simulink. Shared functions make this an implementation-consistency check, not independent physical validation; software provenance and complete logs are in the supplementary package.

\subsection{Unified Activity-Constrained Selection}\label{sec:tuning}

\begin{table}[t]\centering\caption{Tuning outcomes of the main comparison}\label{tab:tune}
\small\setlength{\tabcolsep}{3pt}\begin{tabular}{llrrrl}\toprule Profile & Rule & $\gamma_o$ & $J$ & grid & pass \\\midrule
MRAC & S & 40 & 86.451 & 11 (1) & -- \\
NN-MRAC & S & 40 & 82.551 & 22 (2) & -- \\
CL-att G & G & 80 & 49.551 & 3 (1) & 100/100 \\
CL-Inner G & G & 320 & 41.982 & 3 (1) & 100/100 \\
{\bfseries CL-Both G} & G & 1280 & 37.294 & 3 (1) & 100/100 \\
{\bfseries CL-LPF G} & G & 1280 & {\bfseries 37.214} & 3 (1) & 100/100 \\
GEO-DOB G & G & -- & 69.580 & 1074 (5) & 100/100 \\
INDI G & G & -- & 101.583 & 107 (3) & 100/100 \\
INDI-RPM G & G & -- & 102.502 & 107 (3) & 100/100 \\
INDI-2nd G & G & -- & 70.745 & 96 (3) & 100/100 \\
\bottomrule\end{tabular}\par\smallskip\parbox{\columnwidth}{\footnotesize $J$: S2--S4 mean RMSE$_5$ (mm); pass: fixed-budget S7 checks on validation seeds 5001--5100; grid: candidates (dimensions). S: seed-1 rule (lowest $J$ with seed-1 S7 activity within budget); G: guarded rule (seeds 4001--4020) over three gains for the composite families and over the ten lowest-cost admissible seed-1 candidates for GEO-DOB and the INDI variants. --: not assessed or not applicable. GEO-DOB and the INDI variants use their own gain parameterizations (Section~\ref{sec:fair}); CL-LPF uses $\tau_{\rm LP}=20$\,ms.}\end{table}

Table~\ref{tab:tune} summarizes the selection; the seed-1 profiles of every composite family and the full candidate grids are in \suppref{} (Table~\sref{tab:tune-full}). The seed-1 rule selects, for each family, the lowest design cost $J$ among the candidates whose seed-1 S7 activity stays within the budget of \val{rR}{base}{budget}\,rad/s. It selects $\gamma_o=2560$ for CL-Both against 40 for MRAC, but on noise seeds 3001--3100 the seed-1 CL-Both gain passes only \val{rR}{base}{bothpass}/100 checks and the seed-1 CL-Inner gain \val{rR}{base}{innerpass}/100.

A common empirical guard therefore uses 20 training seeds 4001--4020: the larger of maximum activity and mean plus three sample standard deviations must not exceed the budget. For the composite families it is applied to three gains with the other hyperparameters fixed ($\{40,80,160\}$ for CL-att and CL-Outer, $\{160,320,640\}$ for CL-Inner, $\{640,1280,2560\}$ for CL-Both) and selects \val{rR}{base}{gsel}; all four profiles pass 100/100 validation seeds 5001--5100 against both the fixed budget and 1.05 times paired LQR activity (each exact 95\,\% interval \val{rR}{base}{cilo}--100\,\%). The main text reports the guarded profiles (G) of the composite families and comparators and the seed-1 profiles of MRAC and NN-MRAC.

INDI and $\mathcal{L}_1$ follow the seed-1 rule, with local refinements where the best admissible candidate lay on a grid edge (Appendix~\sref{app:comp}). Only \val{tune}{INDI}{adm} of \val{tune}{INDI}{grid} INDI and \val{tune}{L1}{adm} of \val{tune}{L1}{grid} $\mathcal{L}_1$ candidates meet the budget, because both act on differentiated or high-bandwidth estimates of $\bv{v}$ and $\bv{\omega}$; on seeds 3001--3100 they pass \val{tune}{INDI}{pass}/100 and \val{tune}{L1}{pass}/100 checks. For the comparators the guard is applied to the \val{rIIIE}{INDI}{ncand} lowest-cost admissible seed-1 candidates of each family. INDI keeps its seed-1 candidate (bound \val{rIIIE}{INDI}{gbound}\,rad/s). The seed-1 GEO-DOB candidate ($J=$\val{rIIIE}{GEODOB}{sJ}\,mm, bound \val{rIIIE}{GEODOB}{sbound}\,rad/s) fails, and the guard selects the same design with a \val{rR}{geo}{rtau}\,ms instead of \val{rR}{geo}{srtau}\,ms desired-rate filter ($\kappa_p=\val{rR}{geo}{kp}$, $\kappa_a=\val{rR}{geo}{ka}$, $\tau=\val{rR}{geo}{otau}$\,ms, $\beta=\val{rR}{geo}{beta}$; bound \val{rIIIE}{GEODOB}{gbound}\,rad/s, $J=$\val{rIIIE}{GEODOB}{gJ}\,mm). No $\mathcal{L}_1$ candidate passes (smallest bound \val{rIIIE}{L1}{minbound}\,rad/s), so $\mathcal{L}_1$ has no guarded profile and is left out of the guarded comparison and the fresh Monte Carlo test. Evaluated outside that comparison, its seed-1 candidate reaches $J=\val{rR}{Lonea}{J}$\,mm and Eval6 $=\val{rR}{Lonea}{eval6}$\,mm, below the \val{rIIIE}{best}{eval6}\,mm of guarded GEO-DOB, but does not complete \val{rR}{Lonea}{incS2} of 50 noisy-S2 and \val{rR}{Lonea}{incS4} of 50 noisy-S4 fresh draws (Table~\sref{tab:r2-l1} in \suppref{}). Guarded GEO-DOB and INDI pass \val{rIIIE}{GEODOB}{pass}/100 and \val{rIIIE}{INDI}{pass}/100 validation seeds against both budgets. INDI-2nd admits \val{rB}{indi2}{adm} of its \val{rB}{indi2}{grid} candidates; the seed-1 rule selects the INDI gains ($\kappa_p=\val{rB}{indi2}{po}$, $\kappa_a=\val{rB}{indi2}{pi}$) with $f_c=\val{rB}{indi2}{fc}$\,Hz, the largest cut-off of its grid, which the guard keeps (bound \val{rB}{indi2}{bound}\,rad/s, $J=$\val{rB}{indi2}{J}\,mm, \val{rB}{indi2}{vpass}/100 validation seeds). INDI-RPM, which feeds back measured rotor speeds (true speeds plus Gaussian noise of 1\,\% of $\Omega_{\max}$, an assumed sensor), selects the same gains with $J=$\val{rIIIE}{INDIRPM}{gJ}\,mm; along each INDI tuning dimension through the selected point, every setting with lower $J$ exceeds the budget (Fig.~\sref{fig:indi} in \suppref{}).

With 40 training seeds (4001--4040) the pass/fail decision of all \val{rR}{gd}{ncand} candidates is unchanged (Table~\sref{tab:r2-guard} in \suppref{}). The bootstrap standard error of the guard bound is \val{rR}{gd}{selo}--\val{rR}{gd}{sehi}\,rad/s; the closest passing candidate (CL-LPF, $\gamma_o=1280$) lies \val{rR}{gd}{passmargin}\,rad/s, or \val{rR}{gd}{passnse} standard errors, below the budget, and the closest failing candidate \val{rR}{gd}{failmin}\,rad/s above it.

\subsection{Deterministic Tracking and Predictor Mechanism}\label{sec:extension}

\begin{table*}[t]\centering\caption{Deterministic RMSE$_5$ (mm) of the main comparison}\label{tab:main}\small\setlength{\tabcolsep}{3pt}
\begin{tabular}{lrrrrrrrrrrrr}\toprule Controller & S1 & S2 & S3 & S4 & S5 & S6 & S7 & S8 & S9 & S10 & Design & Eval6 \\\midrule
LQR & 161.2 & 230.9 & 450.2 & 339.0 & 1126.5 & {\bfseries 0.0} & 1.2 & 465.8 & 552.7 & 591.1 & 340.0 & 482.9 \\
MRAC & 31.3 & 76.6 & 88.8 & 93.9 & 138.4 & 5.5 & 3.6 & 68.0 & 147.6 & 112.2 & 86.5 & 83.8 \\
NN-MRAC & 37.0 & 55.5 & 113.6 & 78.6 & 153.8 & 1.7 & 1.6 & 92.6 & 125.9 & 96.3 & 82.6 & 84.5 \\
CL-att G & 25.5 & 44.4 & 49.2 & 55.0 & 151.4 & 0.4 & 1.2 & 74.0 & 69.4 & 42.1 & 49.6 & 60.5 \\
CL-Inner G & 24.9 & 36.1 & 43.1 & 46.8 & 123.8 & 0.3 & {\bfseries 1.1} & 62.8 & 67.3 & 32.2 & 42.0 & 51.9 \\
{\bfseries CL-Both G} & 22.1 & {\bfseries 29.4} & {\bfseries 37.2} & 45.2 & 114.7 & 0.4 & 1.3 & 53.6 & 59.7 & 28.2 & 37.3 & 46.4 \\
{\bfseries CL-LPF G} & 22.1 & {\bfseries 29.4} & {\bfseries 37.2} & {\bfseries 45.0} & {\bfseries 110.4} & 0.4 & {\bfseries 1.1} & 53.6 & {\bfseries 59.3} & {\bfseries 28.0} & {\bfseries 37.2} & {\bfseries 45.6} \\
GEO-DOB G & {\bfseries 16.2} & 53.9 & 69.8 & 85.0 & 267.0 & 0.5 & 1.4 & 91.6 & 145.9 & 202.0 & 69.6 & 120.5 \\
INDI G & 34.4 & 75.1 & 122.1 & 107.5 & 276.6 & 0.3 & 1.5 & 81.1 & 173.7 & 845.2 & 101.6 & 235.2 \\
INDI-RPM G & 34.4 & 75.1 & 124.8 & 107.6 & 275.3 & 0.3 & 4.0 & 82.0 & 174.6 & 861.2 & 102.5 & 238.0 \\
INDI-2nd G & 29.2 & 76.9 & 48.1 & 87.2 & 217.2 & 1.4 & 1.8 & {\bfseries 41.0} & 91.5 & 52.7 & 70.7 & 72.2 \\
\bottomrule\end{tabular}\par\smallskip\parbox{\textwidth}{\footnotesize S2--S4 select gains; S7 constrains activity. G: empirical dispersion-guarded selection with the same rule for every family (seeds 4001--4020). Means use unrounded source values; S7 is excluded from Eval6. Seed-1 profiles are in Table~\sref{tab:main-full}.}\end{table*}

\begin{figure*}[t]\centering\includegraphics[width=.9\textwidth]{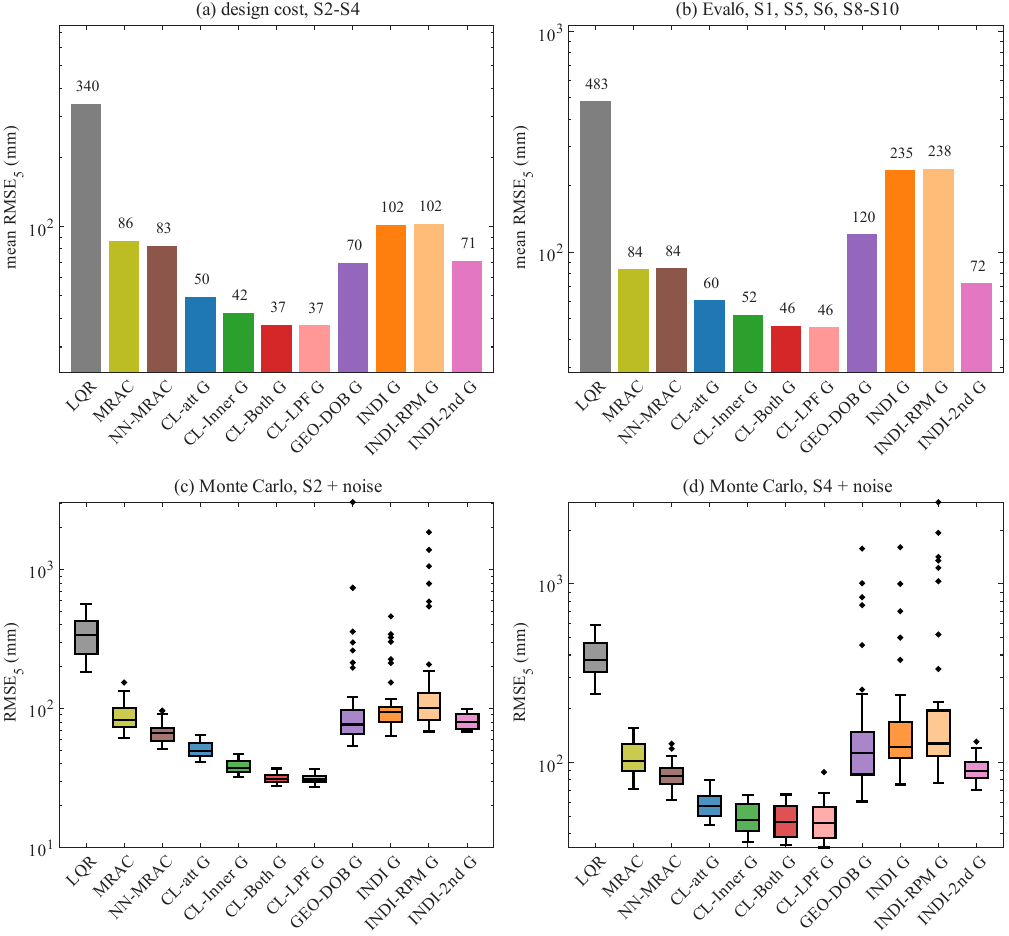}
\caption{Main comparison. (a) Design cost (S2--S4) and (b) Eval6 (S1, S5, S6, S8--S10), deterministic, logarithmic scale. (c), (d) Fresh Monte Carlo RMSE$_5$ distributions (50 noisy draws each, completed runs; incomplete runs are counted in Table~\ref{tab:mc}). G: guarded selection; colors are the same in every panel.}\label{fig:main-results}\end{figure*}

Table~\ref{tab:main} and Fig.~\ref{fig:main-results}(a), (b) report two aggregates: the design cost (mean over S2--S4) and Eval6, the mean over the six evaluation scenarios S1, S5, S6 and S8--S10. Guarded CL-Both reaches \val{det}{GUARDboth}{design}/\val{det}{GUARDboth}{eval6}\,mm, against \val{det}{GUARDinner}{design}/\val{det}{GUARDinner}{eval6}\,mm for guarded CL-Inner, \val{det}{GUARDnone}{design}/\val{det}{GUARDnone}{eval6}\,mm for guarded CL-att and \val{det}{MRAC}{design}/\val{det}{MRAC}{eval6}\,mm for MRAC. Among the guarded non-neural comparators, GEO-DOB gives \val{rIIIE}{GEODOB}{design}/\val{rIIIE}{GEODOB}{eval6}\,mm, INDI \val{rIIIE}{INDI}{design}/\val{rIIIE}{INDI}{eval6}\,mm and INDI-RPM \val{rIIIE}{INDIRPM}{design}/\val{rIIIE}{INDIRPM}{eval6}\,mm; the filtered-target variant CL-LPF G reaches \val{rIE}{bench}{design}/\val{rIE}{bench}{eval6}\,mm. Guarded GEO-DOB is the most accurate controller in nominal S1 (\val{rIIIE}{GEODOB}{Sone} versus \val{det}{GUARDboth}{S1}\,mm for guarded CL-Both), and INDI follows it among the guarded non-neural comparators (\val{det}{INDI}{S1}\,mm); both degrade under the parameter changes and disturbances of S3--S5 and S8--S10, INDI most in the fast helix S10 (\val{det}{INDI}{S10}\,mm). With second-order synchronized filters, INDI-2nd G reaches Eval6 $=\val{rB}{indi2}{eval6}$\,mm and \val{rB}{indi2}{S10}\,mm in S10, the lowest Eval6 of the guarded non-neural comparators, but still above the guarded composite profiles. First-order INDI does not diverge in S10: its tilt reaches the command limit in \val{rIIIE}{diag}{atlim}\,\% of the samples at the low gains that the budget admits (Appendix~\sref{app:comp}). The seed-1 CL-Both profile reaches \val{det}{PROPOSED}{design}/\val{det}{PROPOSED}{eval6}\,mm (Table~\sref{tab:main-full}). Repeating the noisy evaluation scenarios S5 and S8--S10 with ten further noise seeds changes Eval6 little: \val{rV}{CLLPFG}{esmean}$\pm$\val{rV}{CLLPFG}{essd}\,mm for CL-LPF G, \val{rV}{CLBothG}{esmean}$\pm$\val{rV}{CLBothG}{essd}\,mm for CL-Both G and \val{rV}{GEODOBG}{esmean}$\pm$\val{rV}{GEODOBG}{essd}\,mm for GEO-DOB G (mean$\pm$SD); the ranking of the profiles is the same for every seed, and every run completes.

The activity budget applies only to S7. During maneuvers, guarded CL-Both commands more rotor activity than the other neural controllers: its S4 activity $\Delta\Omega$ is \val{det}{GUARDboth}{DWS4}\,rad/s, against \val{det}{GUARDnone}{DWS4} for guarded CL-att, \val{det}{NNMRAC}{DWS4} for NN-MRAC and \val{det}{INDI}{DWS4} for INDI. Part of the tracking gain is therefore paid for with maneuver activity that the S7 constraint does not limit; Section~\ref{sec:budget} adds an explicit in-flight constraint.

Common-gain factorial tests distinguish target effects from gain effects. At $\gamma_o=2560$, CL-att, CL-Outer, CL-Inner and CL-Both give S7 activity 16.99, 16.90, 13.12 and 12.98\,rad/s, with design costs 35.51, 35.60, 35.90 and 35.76\,mm. The inner predictor thus provides most of the activity reduction, and adding the outer predictor brings the configuration under the budget (Fig.~\sref{fig:unified-factorial} in \suppref{}). At a fixed gain the four design costs lie within 0.4\,mm, so the predictors act through the larger gain they make admissible.

Higher gain lowers the tracking error but raises the error of the physical-uncertainty estimate (Table~\sref{tab:gain} in \suppref{}): at high gain the tracking term dominates the composite term and absorbs lag and fast disturbances. CL-Both's tracking advantage therefore comes from the admissible gain rather than from a more accurate model of the physical uncertainty. Seed-1 trajectories and error traces are shown in \suppref{} (Figs.~\sref{fig:traj} and~\sref{fig:err}).

\emph{Motor model versus target filtering.} Two further tests separate the motor model from the low-pass filtering it implies (Fig.~\sref{fig:e1} in \suppref{}). The variant CL-LPF replaces the motor model in both targets by a linear first-order filter with constant $\tau_{\rm LP}$ applied to the commanded thrust and torques. Under the seed-1 rule at 1.05 times LQR, CL-LPF reaches $J=$\val{rIE}{b}{LPF}\,mm with $\tau_{\rm LP}=$\val{rIE}{b}{tau}\,ms and $\gamma_o=$\val{rIE}{b}{go}, against \val{rIE}{b}{Both}\,mm for CL-Both and \val{rIE}{b}{att}\,mm for CL-att; with $\tau_{\rm LP}=5$ or 10\,ms the budget admits only $\gamma_o=$\val{rIE}{tau5}{go}. A first-order target filter with the nominal motor constant therefore reproduces the benefit of the realized-input target on the nominal plant. Without sensor noise and at a common gain, the realized-input target lowers $J$ by at most \val{rIE}{t20g160}{dJ}\,mm on the nominal plant ($\gamma_o=160$) and by \val{rIE}{t60g160}{dJ}\,mm when the plant motor constant is 60\,ms. For that slow plant, the best admissible motor-model constant $\tau_{m0}$ on the grid of Fig.~\sref{fig:e1}(a) is \val{rIE}{c}{optms}\,ms ($J=$\val{rIE}{c}{optJ}\,mm), while the matched constant of 60\,ms gives \val{rIE}{c60}{J}\,mm. Filtering both branches of the target equally, by raising the shared constant $\tau_f$ of CL-att from 50 up to 200\,ms, does not reproduce the CL-LPF benefit: the best admissible design cost is \val{rV}{six}{Jsym}\,mm ($\tau_f=$\val{rV}{six}{tauf}\,ms, $\gamma_o=$\val{rV}{six}{go}), \val{rV}{six}{gap}\,mm above CL-LPF, because the budget still admits only small gains. The benefit therefore requires delaying the input branch relative to the velocity branch, which is what a first-order actuator model does; the nonlinear motor model is not needed.

A one-axis linearization shows where the targets differ (Appendix~\sref{app:r2-lin}). With actuator lag $A(s)=1/(\tau_m s+1)$ and target filter $F(s)$, the commanded-input target contains the residual $-F(s)\,(1-A(s))\,\alpha$ in addition to the filtered uncertainty $F(s)\,d$. Since $1-A(s)=\tau_m s\,A(s)$, this residual is a filtered derivative of the command, which carries the sensor noise of the nominal law. In this linear model, delaying only the input branch by $A$ removes it exactly when $A$ matches the actuator. A longer common filter attenuates it too, and more than the slow part of $d$, but it delays the whole target. Because the residual and the measurement noise in the target are correlated through the feedback, the analysis alone does not show that removing the residual lowers the noise reaching the update; the simulations show it for the tested gains. In a simulation of one roll axis with the inner-loop constants of this paper, the lag-aligned target admits a \val{rR}{lin}{gratio} times larger gain than the commanded target at 1.05 times the nominal activity ($\gamma=\val{rR}{lin}{glag}$ against \val{rR}{lin}{gcmd}), and its attitude error at that budget (\val{rR}{lin}{elag}\,mrad) is lower than with symmetric filtering at any $\tau_f$ from 70 to 200\,ms (at best \val{rR}{lin}{esym}\,mrad) or with the commanded target (\val{rR}{lin}{ecmd}\,mrad).

\subsection{Excitation and Weight Motion}\label{sec:excitation}
The logged information matrices of the guarded composite profiles show that the trajectories excite only a few weight directions: after the first 5\,s, the median number of eigenvalues of $\bv{Q}_o$ above $10^{-3}\lambda_{\max}(\bv{Q}_o)$ is \val{exc}{GUARDbothS4}{MedRankQo} of 15 for CL-Both in S4, and of $\bv{Q}_i$ \val{exc}{GUARDbothS4}{MedRankQi} of 18. The condition $\bv{Q}\succeq\sigma \bv{I}$ of Theorem~\ref{thm:1} therefore holds with no useful $\sigma>0$. The weights nevertheless move quickly, and faster at higher gain (Fig.~\sref{fig:excitation} in \suppref{}). The controllers act as high-gain adaptive compensators that track the current input deficit along the excited directions, not as estimators that converge to a fixed model of the uncertainty. Consistent with this, removing the eight RBF neurons from both regressors (Appendix~\sref{app:lin}) raises the design cost $J$ of the two profiles with lag-aligned inner-loop targets by about 0.63\,mm and lowers that of CL-att by 2.5\,mm, because a larger gain then passes the guard.

\subsection{Paired Random-Parameter Comparisons}\label{sec:robust}

\begin{table*}[t]\centering\caption{Fresh Monte Carlo study: RMSE$_5$ (mm) and activity (rad/s)}\label{tab:mc}\small\setlength{\tabcolsep}{3pt}
\begin{tabular}{lrrrrrrrrrrrrrr}\toprule & \multicolumn{7}{c}{S2 + noise} & \multicolumn{7}{c}{S4 + noise}\\\cmidrule(lr){2-8}\cmidrule(lr){9-15}Controller & mean & median & P95 & $\Delta\Omega$ & fail & $>10$m & $\tilde\Delta$ ($w/n$) & mean & median & P95 & $\Delta\Omega$ & fail & $>10$m & $\tilde\Delta$ ($w/n$) \\\midrule
LQR & 348.3 & 336.8 & 560.7 & 12.25 & 0 & 0 & 306.4 (50/50) & 392.9 & 373.4 & 529.5 & 12.33 & 0 & 0 & 326.6 (50/50) \\
MRAC & 89.1 & 82.5 & 134.0 & 20.57 & 0 & 0 & 52.0 (50/50) & 107.8 & 102.5 & 148.7 & 24.42 & 0 & 0 & 55.1 (50/50) \\
NN-MRAC & 66.8 & 66.6 & 91.1 & 13.69 & 0 & 0 & 33.2 (50/50) & 85.5 & 84.2 & 108.5 & 14.59 & 0 & 0 & 36.5 (50/50) \\
CL-att G & 50.6 & 49.2 & 60.7 & 16.06 & 0 & 0 & 18.5 (50/50) & 59.0 & 57.4 & 76.4 & 19.48 & 0 & 0 & 11.1 (49/50) \\
CL-Inner G & 38.1 & 37.5 & 43.8 & 17.84 & 0 & 0 & 6.9 (50/50) & 49.9 & 47.7 & 64.7 & 21.98 & 0 & 0 & 2.1 (41/50) \\
{\bfseries CL-Both G} & 31.6 & 31.1 & 36.2 & 18.81 & 0 & 0 & -- & {\bfseries 48.0} & 46.6 & {\bfseries 64.3} & 23.22 & 0 & 0 & -- \\
{\bfseries CL-LPF G} & {\bfseries 31.3} & {\bfseries 30.9} & {\bfseries 35.6} & 18.71 & 0 & 0 & -0.3 (0/50) & {\bfseries 48.0} & {\bfseries 46.0} & 65.8 & 23.47 & 0 & 0 & -0.7 (7/50) \\
GEO-DOB G & 184.2 & 76.8 & 739.8 & 11.81 & 1 & 0 & 44.3 (49/49) & 197.9 & 113.5 & 842.1 & 12.09 & 0 & 0 & 57.4 (50/50) \\
INDI G & 118.2 & 94.6 & 326.8 & 12.37 & 2 & 0 & 62.9 (48/48) & 199.5 & 122.0 & 702.8 & 11.97 & 0 & 0 & 71.9 (50/50) \\
INDI-RPM G & 219.9 & 101.7 & 1091.4 & 12.38 & 2 & 0 & 67.7 (48/48) & 330.7 & 127.9 & 1442.2 & 11.93 & 1 & 0 & 81.2 (49/49) \\
INDI-2nd G & 81.6 & 80.3 & 97.5 & 9.03 & 0 & 0 & 50.1 (50/50) & 91.5 & 90.1 & 112.5 & 9.15 & 0 & 0 & 40.3 (50/50) \\
\bottomrule\end{tabular}\par\smallskip\parbox{\textwidth}{\footnotesize One run of 50 noisy draws per scenario; draws 9001--9100 were not used for any design decision. $\tilde\Delta$: median paired difference to CL-Both G (mm, positive: CL-Both G better; 95\,\% bootstrap intervals in Table~\sref{tab:mc-effects}); $w/n$: draws in which CL-Both G has the lower RMSE$_5$ out of $n$ completed pairs without ties. Every Holm-adjusted two-sided exact sign test within a scenario gives $p\le\val{rIVE}{all}{pmax}$. Fail counts incomplete runs; $>10$m counts completed accuracy failures. Mean, median and P95 condition on completed runs with RMSE$_5\le10$ m. $\Delta\Omega$ is mean maneuver activity over all runs. G: guarded selection. Profiles and test plan were fixed before the runs; the written plan is in the supplementary package.}\end{table*}

Table~\ref{tab:mc} reports a fresh Monte Carlo study that was planned before execution and run once: 100 new draws (seeds 9001--9100; 50 noisy S2 and 50 noisy S4) from the distribution used during development, which varies mass over $[0.9,1.2]m_0$, inertia over $[0.8,2]\bv{J}_0$, drag over $[0.7,1.5]c_{d0}$, motor constant over 10--80 ms, disturbance amplitude over $[0.5,1.5]$ times its base scale, frequency over 2--5 rad/s, phase over $[0,2\pi)$, gust time over 5--40 s, equiprobable gust sign and sensor-noise seed. Continuous parameters are drawn uniformly, and controllers keep the nominal motor constant 20 ms. Profiles, statistics and tests were fixed before the runs, and no gain was changed afterwards; the development draws used in the earlier analyses are reported in Table~\sref{tab:mc-dev} in \suppref{}. Table~\ref{tab:mc} and Fig.~\ref{fig:main-results}(c), (d) distinguish incomplete runs from completed runs with RMSE above 10\,m; means, medians and 95th percentiles (P95) condition on completed runs with RMSE$_5\le10$\,m and should be read together with the failure counts.

Guarded CL-Both has the lower error than guarded CL-Inner in \val{rIVE}{S2CLInnerG}{wins}/\val{rIVE}{S2CLInnerG}{n} noisy-S2 draws (mean \val{rIVE}{S2CLBothG}{mean} versus \val{rIVE}{S2CLInnerG}{mean} mm) and in \val{rIVE}{S4CLInnerG}{wins}/\val{rIVE}{S4CLInnerG}{n} noisy-S4 draws (mean \val{rIVE}{S4CLBothG}{mean} versus \val{rIVE}{S4CLInnerG}{mean} mm; median paired difference \val{rIVE}{S4CLInnerG}{med} mm, 95\,\% bootstrap interval \val{rIVE}{S4CLInnerG}{cilo}--\val{rIVE}{S4CLInnerG}{cihi} mm). The S4 tails are similar (P95 \val{rIVE}{S4CLBothG}{p95} versus \val{rIVE}{S4CLInnerG}{p95} mm), and CL-Inner commands less maneuver activity (\val{rIVE}{S4CLInnerG}{meandw} versus \val{rIVE}{S4CLBothG}{meandw} rad/s in S4); on the development draws the two configurations were not separated in S4. CL-LPF G has the lower error than CL-Both G in most draws (CL-Both G is lower in \val{rIVE}{S2CLLPFG}{wins}/\val{rIVE}{S2CLLPFG}{n} S2 and \val{rIVE}{S4CLLPFG}{wins}/\val{rIVE}{S4CLLPFG}{n} S4 draws), but the median paired differences of \val{rIVE}{S2CLLPFG}{med} and \val{rIVE}{S4CLLPFG}{med} mm have bootstrap intervals inside the practical margin of $\pm1$\,mm fixed before the computation, so the two profiles are practically equivalent (Table~\sref{tab:mc-effects} in \suppref{}). Consistent paired signs make the sign tests sensitive to differences far smaller than this margin, so differences are interpreted by size.

Against the guarded non-neural comparators, CL-Both G has the lower error in every completed pair: \val{rIVE}{S2GEODOBG}{wins}/\val{rIVE}{S2GEODOBG}{n} and \val{rIVE}{S4GEODOBG}{wins}/\val{rIVE}{S4GEODOBG}{n} against GEO-DOB, \val{rIVE}{S2INDIG}{wins}/\val{rIVE}{S2INDIG}{n} and \val{rIVE}{S4INDIG}{wins}/\val{rIVE}{S4INDIG}{n} against INDI and \val{rIVE}{S2INDIRPMG}{wins}/\val{rIVE}{S2INDIRPMG}{n} and \val{rIVE}{S4INDIRPMG}{wins}/\val{rIVE}{S4INDIRPMG}{n} against INDI-RPM and \val{rB}{indi2}{Siiw}/\val{rB}{indi2}{Siin} and \val{rB}{indi2}{Sivw}/\val{rB}{indi2}{Sivn} against INDI-2nd, which completes every draw with the narrowest error distribution of the comparators (P95 \val{rB}{indi2}{Siip95} and \val{rB}{indi2}{Sivp95}\,mm; noisy S2/S4). Every paired comparison of Table~\ref{tab:mc} remains significant after Holm correction within its scenario (largest adjusted $p=$\val{rIVE}{all}{pmax}). GEO-DOB fails to complete \val{rIVE}{S2GEODOBG}{inc} and \val{rIVE}{S4GEODOBG}{inc} draws, INDI \val{rIVE}{S2INDIG}{inc} and \val{rIVE}{S4INDIG}{inc} and INDI-RPM \val{rIVE}{S2INDIRPMG}{inc} and \val{rIVE}{S4INDIRPMG}{inc}, against none for the composite profiles; GEO-DOB and INDI also have heavy tails (S4 P95 \val{rIVE}{S4GEODOBG}{p95} and \val{rIVE}{S4INDIG}{p95} mm). These comparators command less maneuver activity (mean $\Delta\Omega$ \val{rIVE}{S4INDIG}{meandw}\,rad/s in S4 for INDI, against \val{rIVE}{S4CLBothG}{meandw} for CL-Both G), which is the trade-off examined in Section~\ref{sec:budget}.

Development-draw results and all seed-1 profiles (Tables~\sref{tab:mc-dev} and~\sref{tab:mc-full}) are given in \suppref{}; on the development draws, reducing the CL-Both gain from 2560 to 1280 lowers the noisy-S4 P95 error from 161.2 to 80.8\,mm (49.9\,\%). Failures are reported as counts rather than replaced by arbitrary large errors.

\subsection{Activity Budget, Pareto Fronts and Sensitivities}\label{sec:budget}
\begin{figure*}[t]\centering\includegraphics[width=.9\textwidth]{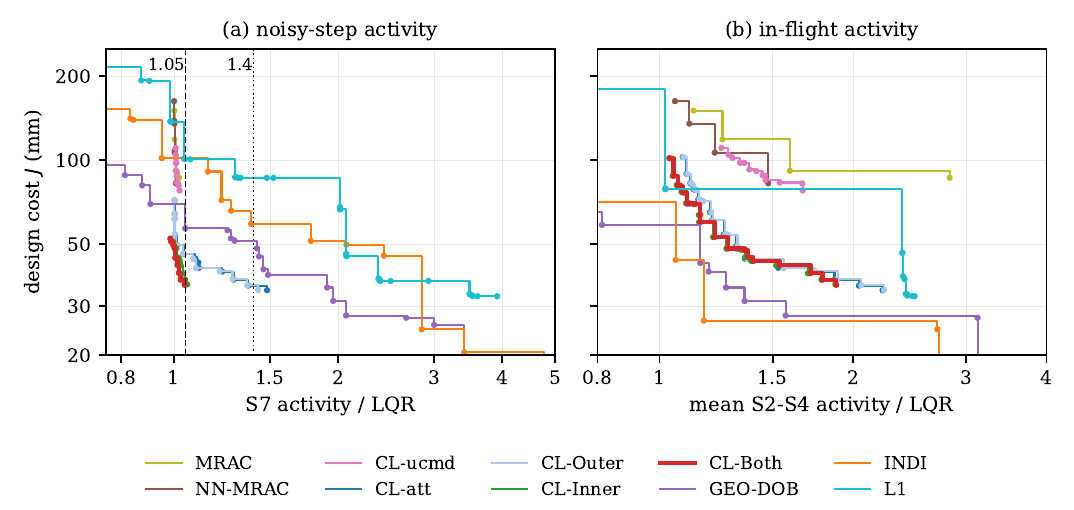}
\caption{Tracking--activity Pareto fronts over the full tuning grids (seed 1). Each line is the lower envelope of design cost $J$ for one family against (a) S7 activity and (b) mean S2--S4 maneuver activity, both relative to LQR. Dashed and dotted lines: budgets $1.05\times$ and $1.4\times$ LQR. CL-LPF, INDI-RPM and INDI-2nd are not shown; CL-LPF is included in Fig.~\ref{fig:dual}.}\label{fig:pareto}\end{figure*}
\begin{figure*}[t]\centering\includegraphics[width=.9\textwidth]{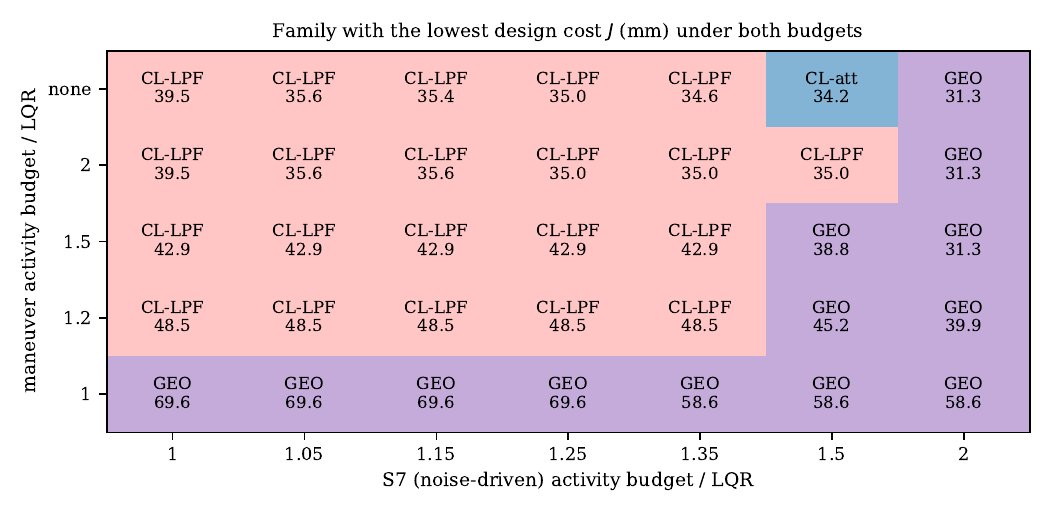}
\caption{Family with the lowest design cost $J$ (mm) on a grid of two activity budgets, both relative to LQR: S7 (noise-driven) activity and mean S2--S4 (maneuver) activity, from the full seed-1 tuning grids including CL-LPF. GEO: GEO-DOB.}\label{fig:dual}\end{figure*}

Fig.~\ref{fig:pareto} shows the trade-off between design cost and activity on the seed-1 tuning grids of the plotted families. Against the noisy-step activity of S7 (panel a), CL-Both has the lowest design cost of the families shown for budgets from \val{par}{all}{bothlo} to \val{par}{all}{bothhi} times LQR activity. At 1.4 times, CL-att is slightly better (\val{par}{1p40}{Jatt} versus \val{par}{1p40}{Jboth}\,mm), at 1.5 times CL-att reaches \val{par}{1p50}{Jatt}\,mm, and at 2.0 times GEO-DOB is best (\val{par}{2p00}{Jbest}\,mm). On these grids, the motor predictor gives the lowest design cost only for S7 thresholds up to \val{par}{all}{bothhi} times LQR; Fig.~\ref{fig:dual} adds the maneuver limit.

Against maneuver activity (panel b) the ordering changes. Without the S7 constraint, INDI reaches a design cost of \val{flt}{1p2}{INDI}\,mm with mean S2--S4 activity below 1.2 times that of LQR, against \val{flt}{1p2}{CLBoth}\,mm for CL-Both and \val{flt}{1p2}{GEODOB}\,mm for GEO-DOB, but its S7 activity is then \val{flt}{1p2}{INDIS7} times that of LQR. INDI and GEO-DOB thus trade low maneuver activity for high noise-driven activity, and the composite neural controllers the reverse. The comparison in Tables~\ref{tab:main} and~\ref{tab:mc} is a comparison under the S7 chatter constraint, and its outcome depends on that choice.

Adding an in-flight constraint to the S7 budget (Table~\sref{tab:dual} in \suppref{}) reduces the CL-Both advantage without removing it. With mean S2--S4 activity limited to that of tuned NN-MRAC (\val{dual}{all}{nnflight}\,rad/s), CL-Both selects $\gamma_o=160$ and $J=$\val{dual}{CLBoth}{nnJ}\,mm, against \val{dual}{CLInner}{nnJ} for CL-Inner and \val{dual}{CLatt}{nnJ}\,mm for CL-att. With the limit $1.2\times$LQR (\val{dual}{all}{lqrflight12}\,rad/s), CL-Both and CL-Inner both give \val{dual}{CLBoth}{lqr12J}\,mm, against \val{dual}{CLatt}{lqr12J}\,mm for CL-att and \val{dual}{GEODOB}{lqr12J}\,mm for GEO-DOB.

Fig.~\ref{fig:dual} combines both limits on one grid. A composite profile has the lowest design cost in every cell with an S7 budget of at most 1.35 times LQR and a maneuver budget of at least 1.2 times LQR; at the selection budget without a maneuver limit, CL-LPF reaches \val{rIIE}{b1p05minf}{CLLPF}\,mm and CL-Both \val{rIIE}{b1p05minf}{CLBoth}\,mm. GEO-DOB has the lowest cost when maneuver activity is held at the LQR level (\val{rIIE}{b1p05m1}{GEODOB}\,mm at an S7 budget of 1.05) and for S7 budgets of 1.5 to 2 times LQR. Most of the S7 command variance lies above the motor corner frequency, and two alternative activity metrics rank the profiles as $\Delta\Omega$ does (Appendix~\sref{app:act}).

A measurement-path variant tests the dependence on the additive Gaussian noise of S7 (Appendix~\sref{app:r3}; amplitudes are assumptions, not measured data): gyro vibration at the hover rotor frequency and its second harmonic, a 40\,Hz second-order gyro low-pass and a 10\,ms delay of position and velocity, with all six families retuned under the same rules. The admissible gains fall and the gain advantage of the lag-aligned targets nearly disappears (guarded $\gamma_o=\val{rB}{n1}{bothcand}$ for CL-Both, CL-Inner and CL-LPF against \val{rB}{n1}{attcand} for CL-att), yet these three profiles keep the lowest Eval6 (\val{rB}{n1}{innereval}--\val{rB}{n1}{botheval}\,mm, within \val{rB}{n1}{compspread}\,mm of each other, CL-Inner lowest), ahead of CL-att (\val{rB}{n1}{atteval}\,mm), GEO-DOB (\val{rB}{n1}{geoeval}\,mm) and INDI (\val{rB}{n1}{indieval}\,mm). With a 20\,ms delay and unchanged gains the order is the same, but guarded CL-Inner exceeds the activity budget (\val{rB}{n2}{inneract} times LQR). INDI-2nd was not run with this variant.

The motor-lag and filter sensitivities are in \suppref{} (Tables~\sref{tab:taum}, \sref{tab:a4} and~\sref{tab:r2-taum}). When the plant motor constant differs from the 20\,ms assumed by the controllers, the guarded profiles keep their tracking up to \val{rR}{tm}{lpfkeep}\,ms for CL-LPF and \val{rR}{tm}{bothkeep}\,ms for CL-Both (design cost at most \val{rR}{tm}{lpfJmax} and \val{rR}{tm}{bothJmax}\,mm); CL-LPF diverges in S3 at \val{rR}{tm}{lpfdiv}\,ms and CL-Both at \val{rR}{tm}{bothdiv}\,ms, whereas CL-att G completes every run up to \val{rR}{tm}{attkeep}\,ms. The activity benefit is lost earlier: with a 40\,ms plant the S7 activity of CL-Both G and CL-LPF G is \val{rR}{tm}{bothact40} and \val{rR}{tm}{lpfact40} times that of LQR, against \val{rR}{tm}{attact40} for CL-att G, so the benefit is already lost at 40\,ms, the first plant constant tested above the nominal one. Retuning the motor-model constant $\tau_{m0}$ of CL-Both for a 60\,ms plant gives \val{rIE}{c}{optms}\,ms ($J=$\val{rIE}{c}{optJ}\,mm) on the grid of Fig.~\sref{fig:e1}(a) and 40\,ms ($J=$\val{fa60}{lp40}{J105}\,mm) on the coarser grid of Table~\sref{tab:a4}, which lacks 50\,ms; the matched 60\,ms gives \val{rIE}{c60}{J}\,mm. Proposition~\sref{prop:mismatch} in \suppref{} bounds one component of the residual only.

\subsection{Sampling, Local Linearization and Computation}\label{sec:effort}
With controller sampling at 2 and 4\,ms and unchanged gains, every deterministic S2/S4 run of seed-1 CL-Both stays bounded, and the S4 error rises from \val{ts1S4}{PROPOSED}{RMSE5} to \val{ts4S4}{PROPOSED}{RMSE5}\,mm (Table~\sref{tab:ts} in \suppref{}). The one-period map of the sampled cascade, linearized at hover by central differences, has spectral radii of at most \val{lin}{all}{rhomax} for LQR and the guarded composite profiles at $T_s=1$, 2 and 4\,ms, with frozen weights, and the same bound holds for the adapting map after \val{lin}{all}{nunit} eigenvalues of unit modulus are excluded; with them its spectral radius is one to within $4\times10^{-13}$. The count \val{lin}{all}{nunit} equals the number of weight directions orthogonal to a constant hover regressor, $3\,(14+17)$, but the eigenvectors were not checked (Table~\sref{tab:linear}). The controller step, dominated by an $18\times18$ symmetric solve, takes \val{time}{PROPOSED}{mex}\,$\mu$s in a C MEX loop on a desktop computer, against \val{time}{LQR}{mex}\,$\mu$s for LQR.

\subsection{PX4 Software-in-the-Loop Check}\label{sec:sitl}
The archived LQR and guarded CL-Both ($\gamma_o=1280$) controllers were ported to a double-precision C++ module and run in PX4 v1.15.4 software-in-the-loop (SITL) simulation with the built-in simulation-in-hardware (SIH) rigid-body simulator. The module executes on every 1\,kHz gyro sample in the PX4 rate-control work queue, reads the position, velocity and attitude estimates of the PX4 extended Kalman filter (EKF2) and the filtered body rate, and publishes rotor commands to the PX4 output driver, so the PX4 position, attitude, rate and allocation controllers are bypassed. Replaying 4000 steps of a noisy MATLAB S5 run through the port reproduces the MATLAB rotor commands within $3.8\times10^{-12}$\,rad/s. Gains keep their archived values.
SIH uses the mass, inertia, thrust, yaw-torque and drag coefficients of Section~\ref{sec:model}, a 20\,ms first-order thrust lag and an X rotor layout with the same arm length; both controllers allocate with this layout. With ground-truth state feedback on the S2 path, the errors of LQR and CL-Both are \val{sitlGTSii}{LQR}{RMSE5} and \val{sitlGTSii}{PROPOSED}{RMSE5}\,mm, close to the MATLAB values of \val{det}{LQR}{S2} and \val{det}{GUARDboth}{S2}\,mm in Table~\ref{tab:main}; with a wind profile they are \val{sitlGTWind}{LQR}{RMSE5} and \val{sitlGTWind}{PROPOSED}{RMSE5}\,mm. This agreement checks the port and the SIH configuration under the remaining plant differences (X layout, Euler integration, thrust-domain lag and output quantization).

With EKF2 feedback (five noise seeds per case; Table~\sref{tab:sitl} and Appendix~\sref{app:sitl}), CL-Both has the lower ground-truth tracking error on \val{sitlSii}{all}{wins} of the 5 tested seeds in S2 and \val{sitlWind}{all}{wins} of 5 with wind. This concerns tracking only, and the EKF2 position error (\val{sitlSii}{LQR}{esterr} and \val{sitlSii}{PROPOSED}{esterr}\,mm RMS) is comparable in size to the ground-truth error of both controllers. The activity budget is not shown to transfer: no tested gain, with or without filtering of the estimates seen by the adaptive law, keeps hover activity within 1.05 times that of LQR on every seed. On the Windows Subsystem for Linux (WSL2) host used here, the simulated barometer and magnetometer report wall-clock timeouts in most runs, and runs with timeouts tended to command more activity; longer settling, a higher process priority and a slower simulation clock did not remove them, so a per-seed activity comparison in PX4 remains open. The controller step inside the PX4 process takes \val{sitlSii}{PROPOSED}{cpu}\,$\mu$s for CL-Both and \val{sitlSii}{LQR}{cpu}\,$\mu$s for LQR (mean, wall clock, 24-core x86 host). The supplementary package contains the module source, the PX4 patch, one summary row per SITL run (including runs with timeouts) and the aggregation scripts; the raw PX4 logs are not included.

\subsection{Limitations}\label{sec:limits}
\emph{Evidence.} The evidence is simulation-only; no hardware-in-the-loop (HIL) or flight test was performed. The MATLAB and Simulink implementations share the plant and controller functions, so their agreement checks the implementation rather than the physics. The MATLAB model omits ground effect, rotor aerodynamics and estimator dynamics; SIH adds the estimator but still omits rotor aerodynamics and ground effect. The PX4 check of Section~\ref{sec:sitl} uses the SIH rigid-body simulator rather than Gazebo, flies only LQR and guarded CL-Both on the S2 path and in hover with five seeds (ten in the follow-up), keeps the MATLAB gains (a re-selection in SITL found no tested gain within the budget) and reports host rather than autopilot timings. Most SITL runs report a simulated barometer or magnetometer timeout caused by the host, which raises command activity, and too few timeout-free pairs remain to test the budget per seed. The activity-constrained ranking is therefore established only under the additive noise model of the MATLAB study. Execution times are wall-clock measurements on an x86 host (the PX4 timings with SITL running in WSL2), not embedded worst-case execution times.

\emph{Model dependence and reuse of data.} Tuning depends on the simulation model. The development draws were reused across analyses and are not an independent holdout; the fresh draws of Table~\ref{tab:mc} were run once with profiles fixed in advance, but they come from the same simulation model and parameter distribution. Sensor noise is additive and Gaussian in the main comparison; the vibration, gyro-filter and delay variant uses assumed amplitudes, retunes only the gain line of the composite families and omits notch filters and INDI-2nd. The activity benefit of the realized-input target is lost at a plant motor constant of 40\,ms, twice the modeled value and the first mismatch tested above it (Section~\ref{sec:budget}).

\emph{Activity measure and selection.} The activity budget is a chatter proxy measured in one noisy step and does not demonstrate an energy, heating or hardware-life benefit; under a maneuver-activity constraint the ranking changes (Section~\ref{sec:budget}). The multi-seed activity guard is an engineering heuristic over a local candidate set, not a probabilistic guarantee, and each grid search covers one design family without establishing a global optimum for it. Bootstrap intervals and signed-rank tests on the development draws are pointwise and exploratory, not simultaneous or equivalence intervals.

\emph{Comparators and ablations.} The B1 surrogate is our re-implementation, and its failures do not establish failure of the published method. INDI and $\mathcal{L}_1$ are implemented on the quaternion LQR cascade with derivative-free acceleration estimates and tuned on finite grids under the same rule. INDI with measured rotor speeds ranks as INDI does, under an assumed rotor-speed sensor; INDI with measured accelerations and $\mathcal{L}_1$ on a geometric baseline were not tested and may rank differently. INDI-2nd cut-offs above 16\,Hz, tested up to 128\,Hz, exceed the budget. The comparison without RBF neurons retunes only the gain and keeps the other hyperparameters of the RBF profiles.

\emph{Theory.} Theorem~\ref{thm:1} and Corollary~\ref{cor:lambda} cover only the continuous-time translational loop; the corollary admits a constant unknown mass ratio but not time-varying or attitude-loop input-gain errors. The excitation logs show that the composite information matrices stay far from full rank, so the parameter-convergence part of Theorem~\ref{thm:1} does not apply to the reported runs. The nonlinear sampled cascade has no stability proof, and because the Gaussian sensor noise is unbounded, the deterministic bounds are not stochastic guarantees; the local linearization of Section~\ref{sec:effort} is numerical evidence for the tested gains and equilibria only.

\section{Conclusion}\label{sec:concl}
Under a tight noise-driven activity budget, a composite-learning target that passes the commanded inputs through a first-order actuator model admits larger adaptation gains in a quadrotor neural MRAC. A linear model with the motor time constant gives the same benefit as the nonlinear motor model, and filtering both sides of the prediction error equally does not: the commanded-input target contains a filtered derivative of the command, which, in the one-axis linear model with a matched motor constant, the input-branch lag removes exactly, whereas a longer common filter attenuates it only by delaying the whole target. An implementation therefore needs only a first-order filter with the motor time constant on the commanded thrust and torques before they enter the prediction error. Its activity benefit is lost at a plant motor constant twice the modeled value, the first mismatch tested above it. In the tested profiles the benefit does not depend on the RBF neurons: with a linear regressor and the gain retuned, the design cost of the two guarded profiles with lag-aligned targets in both loops rises by about 0.63\,mm, and their median paired fresh-draw errors in S2 and S4 rise by 0.54 to 0.62\,mm.

The result is conditional on the budget. With the S7 budget of 1.05 times LQR, the guarded configurations with lag-aligned targets in both loops have lower design and evaluation costs than MRAC, INDI, INDI with second-order synchronized filters (the best guarded non-neural comparator) and GEO-DOB, and the lower error in every completed fresh random-parameter pair against them, and the guard decisions are unchanged with 40 training seeds; $\mathcal{L}_1$ misses the guard by less than 1\,\%, and its seed-1 candidate, evaluated outside the guarded comparison, also tracks worse than the composite profiles. GEO-DOB remains the most accurate controller in the nominal scenario S1, has the lowest design cost when maneuver activity is held at the LQR level, and INDI attains lower design cost when only maneuver activity is limited. The logged information matrices show that the gain works as high-gain compensation rather than as identification. With gyro vibration, gyro filtering and a 10\,ms measurement delay, the admissible gains fall to nearly those of CL-att, but the three lag-aligned profiles still have the lowest evaluation cost of the retuned families, which exclude INDI-2nd, within \val{rB}{n1}{compspread}\,mm of each other.

In PX4 SITL with EKF2 feedback, guarded CL-Both has lower ground-truth S2 error than LQR on each of the five tested seeds, with and without wind, but the activity budget is not shown to hold on every seed: no tested gain meets it, filtering the estimates used by the adaptive law reduces the excess without removing it per seed, and simulated sensor timeouts on the host confound the per-seed ratios.

The next steps follow from these limits: SITL on a host without sensor timeouts, so that the activity threshold can be tested per seed; HIL and flight tests; identification of the motor time constant, since the activity benefit is lost at twice the modeled value; and a cascade analysis with input-gain uncertainty.

\appendices

\section{Scenarios and Comparator Details}\label{app:setup}

\subsection{Simulation Scenarios}\label{app:scen}

S1: helix $\bv{p}_d=[\cos t,\sin t,t]$\,m, $\psi_d=-0.5$\,rad, 30\,s, start at the origin.
S2: take-off to 9\,m in 6\,s (quintic), then lawn-mower legs of 10\,m at 2\,m/s with U-turns of
radius 1.5\,m (speed ramp 2\,s), 60\,s. S3: S1 with the parametric changes of \cite{madebo2024}: $\bv{J}=2\bv{J}_0$,
$c_d=(1+0.1t)c_{d0}$, $c_a=(1+0.01t)c_{a0}$. S4: S2 with
$\bv{d}_F=[0.3,0.3,0.2]^\top\!\odot[\sin0.03t+0.5\sin3t+g_s,\ \sin0.03t+0.5\cos3t-0.5g_s,\ 0.5\sin3t]^\top$\,N
and $\bv{d}_\tau=[0.01,0.01,0.005]^\top\!\odot[\sin3t+0.5g_s,\ \cos3t-0.5g_s,\ \sin3t]^\top$\,N\,m, where $g_s$
is a unit step at $t=10$\,s (gust). S5: S2 with the parametric changes of S3, a 20\,\% mass increase
($m=1.2m_0$), the disturbances of S4 and sensor noise. S6: set-point step from $(0,0,1)$ to $(1,1,2)$\,m. S7: the S6 step with sensor noise over 20\,s (actuator-activity constraint, including the step transient). S8: figure-eight $\bv{p}_d=[4\sin\theta,\,4\sin\theta\cos\theta,\,5]$\,m,
$\dot\theta$ ramped to 0.4\,rad/s in 3\,s, 40\,s. S9: S2 path. S10: helix
$\bv{p}_d=[2\cos0.6t,\,2\sin0.6t,\,0.5t]$\,m, $\psi_d=0.3$\,rad, start at $(2,0,0)$\,m. The disturbances of
S8--S10 have the structure of S4 with the amplitude scale, fast frequency, phase, slow frequency, gust
time and gust sign of Table~\mref{tab:scen} (S8: slow 0.05\,rad/s, phase 0.7; S9: slow 0.05\,rad/s;
S10: slow 0.1\,rad/s, phase 1.2, drag rate 0.05\,s$^{-1}$, $c_a$ rate 0.01\,s$^{-1}$).

\subsection{Comparator Details}\label{app:comp}

\emph{GEO-DOB desired rate.} Desired rate is obtained from
$\operatorname{vee}(\operatorname{skew}[\bv{R}_{d,k}^\top(\bv{R}_{d,k}-\bv{R}_{d,k-1})/T_s])$
and filtered by a first-order Euler filter of constant $\tau_R$; its finite-difference derivative is filtered again with the same constant. Both are initially zero. The comparator is defined by this discrete implementation; its model and gain dimensions differ from those of the adaptive laws.

\begin{figure*}[t]\centering
	\includegraphics[width=.9\textwidth]{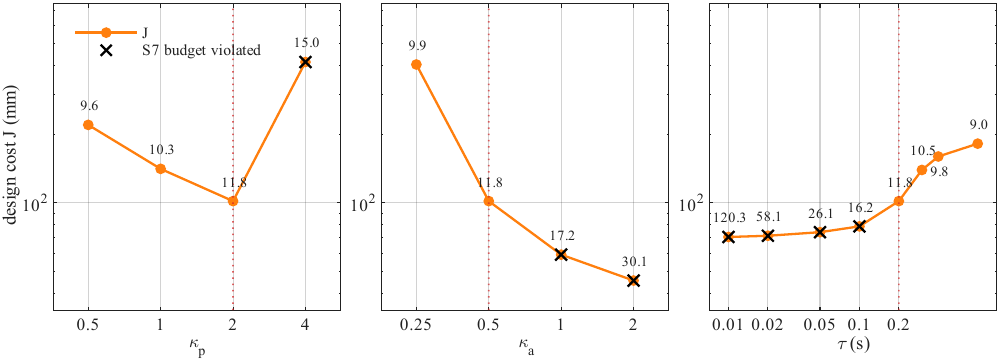}
	\caption{INDI design cost along each tuning dimension through the selected point $(\kappa_p,\kappa_a,\tau)=(2,0.5,0.2\,\mathrm{s})$; labels give S7 activity (rad/s), crosses mark candidates above the budget.}\label{fig:indi}
\end{figure*}

\emph{Seed-1 INDI and $\mathcal{L}_1$ selections.} INDI and $\mathcal{L}_1$ follow the seed-1 rule. On the initial grids their best admissible candidates lay on grid edges (INDI filter constant 200\,ms; $\mathcal{L}_1$ predictor pole 10\,s$^{-1}$ and inner filter 10\,rad/s), so local refinements of 27 and 72 candidates were added, in the same way as for GEO-DOB; the selected candidates are interior to the refined grids. INDI selects $\kappa_p=2$, $\kappa_a=0.5$ and $\tau=200$\,ms with $J=$\val{tune}{INDI}{J}\,mm, and $\mathcal{L}_1$ selects $a_o=10$\,s$^{-1}$, $\omega_{c,o}=5$, $\omega_{c,i}=10$\,rad/s and unit scale with $J=$\val{tune}{L1}{J}\,mm. Only \val{tune}{INDI}{adm} of \val{tune}{INDI}{grid} INDI candidates and \val{tune}{L1}{adm} of \val{tune}{L1}{grid} $\mathcal{L}_1$ candidates meet the budget: both act on differentiated or high-bandwidth estimates of $\bv{v}$ and $\bv{\omega}$, which pass sensor noise to the rotor commands. INDI passes \val{tune}{INDI}{pass}/100 and $\mathcal{L}_1$ \val{tune}{L1}{pass}/100 S7 checks on seeds 3001--3100; the selected $\mathcal{L}_1$ candidate lies 0.3\,\% below the budget on seed~1. Fig.~\ref{fig:indi} shows the INDI design cost along each tuning dimension through the selected point: every setting with a lower cost exceeds the budget.

\emph{INDI in S10.} In S10, INDI never commands a rotor at its limit, and its error stays bounded (maximum \val{rIIIE}{diag}{emax}\,m, RMS \val{rIIIE}{diag}{efirst}\,mm over 5--15\,s and \val{rIIIE}{diag}{elast}\,mm over the last 10\,s), but the measured tilt reaches the \val{rIIIE}{diag}{tiltlim}$^\circ$ command limit in \val{rIIIE}{diag}{atlim}\,\% of the samples (maximum \val{rIIIE}{diag}{tiltmax}$^\circ$), against \val{rIIIE}{diagB}{atlim}\,\% for guarded CL-Both (maximum \val{rIIIE}{diagB}{tiltmax}$^\circ$). The S10 loss of INDI is therefore tilt-limited tracking at the low gains that the budget admits, not divergence; INDI-RPM gives \val{rIIIE}{INDIRPM}{Sten}\,mm.

\section{Motor-Mismatch Residual}\label{app:mismatch}

Proposition~\ref{prop:mismatch} bounds the contribution of a motor-model mismatch $\tau_m\neq\tau_{m0}$ to the residual $\bv{\xi}$ of Theorem~\mref{thm:1}.

\begin{proposition}[Motor-mismatch residual]\label{prop:mismatch}
Let the plant motor and nominal predictor have constant positive time constants $\tau_m,\tau_{m0}$ and the same clipped command $\bv{c}(t)\in[0,\Omega_{\max}]^4$, with both initial rotor states in that box. Set $\widetilde{\bv{\Omega}}=\bv{\Omega}-\hat{\bv{\Omega}}$. If $\|\dot{\hat{\bv{\Omega}}}\|_2\le D_\Omega$ on $[0,t]$, then
\begin{equation}
\|\widetilde{\bv{\Omega}}(t)\|_2\le
e^{-t/\tau_m}\|\widetilde{\bv{\Omega}}(0)\|_2+
|\tau_m-\tau_{m0}|D_\Omega(1-e^{-t/\tau_m}).
\label{eq:motorbound}
\end{equation}
With exact measured attitude, the outer realized-input error satisfies $\|\bv{u}_{\rm eff}^{\rm true}-\bv{u}_{\rm eff}\|_2\le L_o\|\widetilde{\bv{\Omega}}\|_2$, where $L_o=4k\Omega_{\max}/m_0$. For equal initial rotor states, a zero-initialized stable first-order filter and bounded $\|\bv{\phi}_f\|_2\le\Phi_f$, the motor contribution to the information residual is bounded by
\begin{equation}
\|\bv{\xi}_{\rm motor}(t)\|_F\le
\Phi_f L_o|\tau_m-\tau_{m0}|D_\Omega .
\label{eq:ximotor}
\end{equation}
\end{proposition}
\begin{IEEEproof}
Subtracting the two motor equations gives
$\tau_m\dot{\widetilde{\bv{\Omega}}}+\widetilde{\bv{\Omega}}
=-(\tau_m-\tau_{m0})\dot{\hat{\bv{\Omega}}}$.
Variation of constants yields \eqref{eq:motorbound}. The clipped-input first-order dynamics preserve the rotor box. Consequently
$\|\bv{\Omega}^{\circ2}-\hat{\bv{\Omega}}^{\circ2}\|_2\le2\Omega_{\max}\|\widetilde{\bv{\Omega}}\|_2$.
The thrust map has norm $k\|\mathbf1_4^\top\|_2/m_0=2k/m_0$, giving $L_o$. The filter has unit induced $L_\infty$ gain. The motor component of the information residual $\bv{\xi}$ defined below is
$-\int_0^t\lambda e^{-\lambda(t-s)}\bv{\phi}_f(s)\bv{\zeta}_{\rm motor}(s)^\top ds$,
where $\bv{\zeta}_{\rm motor}=H[\bv{u}_{\rm eff}^{\rm true}-\bv{u}_{\rm eff}]$. Bounding the integrand and using the kernel mass $1-e^{-\lambda t}\le1$ proves \eqref{eq:ximotor}.
\end{IEEEproof}
The total residual is $\bv{\xi}=\bv{\xi}_{\rm approx}+\bv{\xi}_{\rm motor}+\bv{\xi}_{\rm meas}+\bv{\xi}_{\rm init}+\bv{\xi}_{\rm disc}$ when these components are defined by the filtered regression. Only the motor contribution vanishes with time-constant mismatch; NN approximation, measurement, initialization and discretization effects need not vanish. The bounds concern continuous motor/filter models; sampled predictor/filter errors add $\bv{\xi}_{\rm disc}$ even when $\tau_m=\tau_{m0}$. In the inner torque target the same argument uses Lipschitz constant $2\Omega_{\max}\|\bv{J}_0^{-1}\bv{M}_\tau\|_2$, with $\bv{M}_\tau$ the torque rows of the allocation map. Gyroscopic and other unmodeled effects remain outside this motor term. For comparison, if the clipped motor command is absolutely continuous with $\|\dot {\bv{c}}\|_2\le D_c$, the command-state error obeys
\[
\|\bv{\Omega}(t)-\bv{c}(t)\|_2\le e^{-t/\tau_m}\|\bv{\Omega}(0)-\bv{c}(0)\|_2
+\tau_m D_c(1-e^{-t/\tau_m}).
\]
This follows from $\tau_m\frac{d}{dt}(\bv{\Omega}-\bv{c})+(\bv{\Omega}-\bv{c})=-\tau_m\dot {\bv{c}}$. Thus a motor-only commanded-input discrepancy involves the full time constant, whereas the predictor discrepancy in \eqref{eq:motorbound} involves its mismatch. The derivative bounds are conditional assumptions: they do not automatically hold for discontinuous noisy sampled commands, and they do not by themselves bound the full CL-ucmd learning residual.
No general scalar $\tau_{\rm att}$ exists for the nonlinear attitude cascade, so a universal CL-ucmd bound proportional to $\tau_m+\tau_{\rm att}$ is not asserted.

\section{Full Selection and Comparison Results}\label{app:full}

\begin{figure}[!tbp]
\centering\includegraphics[width=\columnwidth]{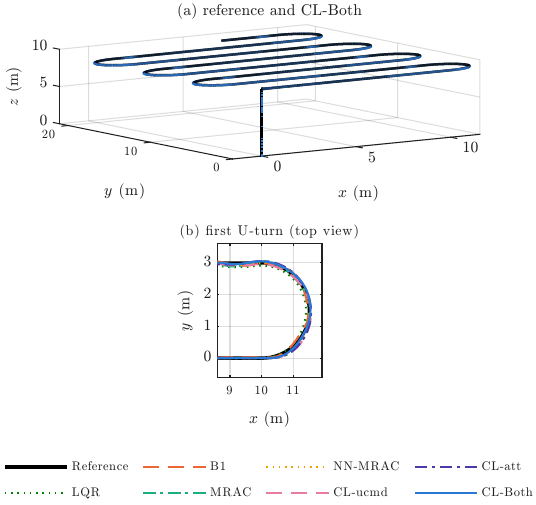}
\caption{S2 seed-1 trajectories: (a) reference and CL-Both; (b) first U-turn. CL-Both uses gain 2560; B1 is a baseline-inspired surrogate.}
\label{fig:traj}
\end{figure}

\begin{figure}[!tbp]
\centering\includegraphics[width=\columnwidth]{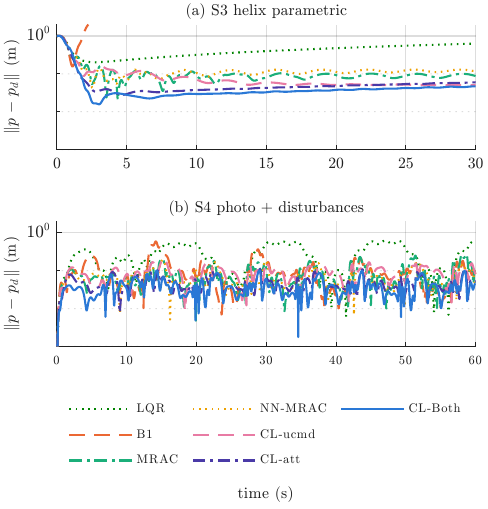}
\caption{Seed-1 position-error norms: (a) S3 parameter changes; (b) S4 disturbances. CL-Both uses gain 2560.}
\label{fig:err}
\end{figure}

This appendix collects the profiles that the main text summarizes in one sentence each: the seed-1 profiles of the composite families, CL-ucmd, CL-Outer and the B1 surrogate, together with INDI and $\mathcal{L}_1$. Tables~\ref{tab:tune-full} and~\ref{tab:main-full} extend Tables~\mref{tab:tune} and~\mref{tab:main}. Seed-1 CL-Both (gain 2560) reaches \val{det}{PROPOSED}{design}/\val{det}{PROPOSED}{eval6}\,mm design/Eval6 cost and passes \val{rR}{base}{bothpass}/100 S7 checks; seed-1 CL-Inner (gain 1280) \val{det}{PREDinner}{design}/\val{det}{PREDinner}{eval6}\,mm and \val{rR}{base}{innerpass}/100; CL-Outer behaves like CL-att because the outer predictor alone barely changes S7 activity; CL-ucmd, with commanded inputs in both targets, admits only gain 20 and reaches \val{det}{CLucmd}{design}/\val{det}{CLucmd}{eval6}\,mm; B1 has no feasible all-design candidate. Fig.~\ref{fig:traj} shows the seed-1 trajectories in S2 and Fig.~\ref{fig:err} the position-error norms in S3 and S4.

\begin{table}[t]\centering\caption{Tuning outcomes of all neural profiles and GEO-DOB}\label{tab:tune-full}\small\setlength{\tabcolsep}{3pt}\begin{tabular}{llrrl}\toprule Profile & Rule & $\gamma_o$ & $J$ & pass \\\midrule
MRAC & seed 1 & 40 & 86.451 & -- \\
NN-MRAC & seed 1 & 40 & 82.551 & -- \\
CL-ucmd & seed 1 & 20 & 77.797 & -- \\
CL-att & seed 1 & 160 & 46.155 & 95/100 \\
CL-Outer & seed 1 & 160 & 46.215 & 99/100 \\
CL-Inner & seed 1 & 1280 & 37.299 & 63/100 \\
{\bfseries CL-Both} & seed 1 & 2560 & {\bfseries 35.763} & 92/100 \\
GEO-DOB & seed 1 & -- & 56.986 & -- \\
CL-att G & G & 80 & 49.551 & 100/100 \\
CL-Outer G & G & 80 & 49.602 & 100/100 \\
CL-Inner G & G & 320 & 41.982 & 100/100 \\
{\bfseries CL-Both G} & G & 1280 & 37.294 & 100/100 \\
\bottomrule\end{tabular}\par\smallskip\parbox{\columnwidth}{\footnotesize $J$: S2--S4 mean RMSE$_5$ (mm); pass: independent fixed-budget S7 noise checks. G: local empirical guarded selection. Seed-1 pass uses seeds 3001--3100; G uses 5001--5100. --: not assessed/not applicable. B1 has no feasible all-design candidate. GEO-DOB uses a distinct geometric gain parameterization.}\end{table}

\begin{table*}[t]\centering\caption{Deterministic RMSE$_5$ (mm) of all profiles}\label{tab:main-full}\small\setlength{\tabcolsep}{3pt}
\begin{tabular}{lrrrrrrrrrrrr}\toprule Controller & S1 & S2 & S3 & S4 & S5 & S6 & S7 & S8 & S9 & S10 & Design & Eval6 \\\midrule
LQR & 161.2 & 230.9 & 450.2 & 339.0 & 1126.5 & {\bfseries 0.0} & 1.2 & 465.8 & 552.7 & 591.1 & 340.0 & 482.9 \\
MRAC & 31.3 & 76.6 & 88.8 & 93.9 & 138.4 & 5.5 & 3.6 & 68.0 & 147.6 & 112.2 & 86.5 & 83.8 \\
B1 & 36.2 & 85.6 & -- & 136.4 & -- & 15.2 & 8.7 & -- & -- & -- & -- & -- \\
NN-MRAC & 37.0 & 55.5 & 113.6 & 78.6 & 153.8 & 1.7 & 1.6 & 92.6 & 125.9 & 96.3 & 82.6 & 84.5 \\
CL-ucmd S & 21.8 & 74.0 & 63.3 & 96.1 & 132.7 & 25.5 & 15.0 & 97.7 & 137.8 & 77.3 & 77.8 & 82.1 \\
CL-att S & 25.4 & 41.5 & 46.1 & 50.9 & 142.2 & 0.3 & 1.2 & 69.5 & 67.7 & 36.7 & 46.2 & 57.0 \\
CL-Outer S & 25.4 & 41.5 & 46.1 & 51.0 & 142.3 & 0.4 & 1.4 & 69.6 & 68.0 & 36.9 & 46.2 & 57.1 \\
CL-Inner S & 22.1 & 29.4 & 37.3 & 45.2 & {\bfseries 112.6} & 0.3 & {\bfseries 1.1} & 53.5 & 60.1 & 28.1 & 37.3 & 46.1 \\
{\bfseries CL-Both S} & 21.4 & {\bfseries 27.5} & {\bfseries 35.8} & {\bfseries 44.0} & 113.6 & 0.4 & 1.3 & {\bfseries 51.2} & {\bfseries 57.0} & {\bfseries 27.3} & {\bfseries 35.8} & {\bfseries 45.1} \\
CL-att G & 25.5 & 44.4 & 49.2 & 55.0 & 151.4 & 0.4 & 1.2 & 74.0 & 69.4 & 42.1 & 49.6 & 60.5 \\
CL-Outer G & 25.5 & 44.5 & 49.3 & 55.1 & 151.3 & 0.4 & 1.4 & 74.2 & 69.6 & 42.3 & 49.6 & 60.5 \\
CL-Inner G & 24.9 & 36.1 & 43.1 & 46.8 & 123.8 & 0.3 & {\bfseries 1.1} & 62.8 & 67.3 & 32.2 & 42.0 & 51.9 \\
{\bfseries CL-Both G} & 22.1 & 29.4 & 37.2 & 45.2 & 114.7 & 0.4 & 1.3 & 53.6 & 59.7 & 28.2 & 37.3 & 46.4 \\
GEO-DOB & {\bfseries 16.9} & 38.6 & 69.7 & 62.7 & 255.6 & 0.9 & 1.5 & 80.7 & 111.2 & 460.8 & 57.0 & 154.3 \\
INDI & 34.4 & 75.1 & 122.1 & 107.5 & 276.6 & 0.3 & 1.5 & 81.1 & 173.7 & 845.2 & 101.6 & 235.2 \\
$\mathcal{L}_1$ & 53.8 & 80.2 & 130.0 & 92.1 & 329.4 & 0.6 & 1.5 & 96.3 & 119.9 & 67.9 & 100.8 & 111.3 \\
\bottomrule\end{tabular}\par\smallskip\parbox{\textwidth}{\footnotesize Includes seed-1 profiles, CL-ucmd, CL-Outer and B1. S: seed-1 selection; G: empirical dispersion-guarded selection. --: incomplete run or invalid aggregate. Means use unrounded source values; S7 is excluded from Eval6. B1 is a baseline-inspired surrogate.}\end{table*}

\subsection{Monte Carlo Results}

Table~\ref{tab:mc-effects} gives the median paired difference of each profile to guarded CL-Both in the fresh Monte Carlo study of Table~\mref{tab:mc}, with bootstrap intervals; only the interval of CL-LPF G lies inside the practical margin. Table~\ref{tab:mc-dev} reports the main comparison on the development draws, which were reused across the earlier analyses, and Table~\ref{tab:mc-full} extends it to all profiles.

\begin{table}[t]\centering\caption{Fresh Monte Carlo: median paired RMSE$_5$ difference to CL-Both G (mm)}\label{tab:mc-effects}\small\setlength{\tabcolsep}{3pt}
\begin{tabular}{lrrrr}\toprule & \multicolumn{2}{c}{S2 + noise} & \multicolumn{2}{c}{S4 + noise}\\\cmidrule(lr){2-3}\cmidrule(lr){4-5}Controller & $\tilde\Delta$ & 95\,\% CI & $\tilde\Delta$ & 95\,\% CI \\\midrule
LQR & 306.39 & [259.86, 357.81] & 326.56 & [299.49, 364.84] \\
MRAC & 52.03 & [46.96, 57.05] & 55.10 & [48.98, 71.63] \\
NN-MRAC & 33.22 & [31.93, 36.83] & 36.47 & [31.85, 40.34] \\
CL-att G & 18.45 & [16.70, 20.37] & 11.08 & [9.64, 12.00] \\
CL-Inner G & 6.91 & [5.77, 7.48] & 2.06 & [1.29, 2.78] \\
{\bfseries CL-LPF G} & -0.25$^{\approx}$ & [-0.33, -0.20] & -0.66$^{\approx}$ & [-0.81, -0.54] \\
GEO-DOB G & 44.29 & [38.43, 57.86] & 57.43 & [47.69, 84.74] \\
INDI G & 62.90 & [57.55, 67.01] & 71.87 & [62.73, 96.58] \\
INDI-RPM G & 67.73 & [60.90, 77.91] & 81.16 & [69.20, 113.91] \\
\bottomrule\end{tabular}\par\smallskip\parbox{\columnwidth}{\footnotesize Other minus CL-Both G, with 95\,\% percentile bootstrap interval (10000 resamples) over completed pairs; $\approx$: interval inside the practical margin of $\pm1$\,mm fixed before the computation. INDI-2nd G: medians in Table~\mref{tab:mc}.}
\end{table}

\begin{table*}[t]\centering\caption{Paired Monte Carlo results on the development draws}\label{tab:mc-dev}\small\setlength{\tabcolsep}{4pt}
\begin{tabular}{lrrrrrrrrrrrr}\toprule & \multicolumn{6}{c}{S2 + noise} & \multicolumn{6}{c}{S4 + noise}\\\cmidrule(lr){2-7}\cmidrule(lr){8-13}Controller & mean & median & P95 & $\Delta\Omega$ & fail & $>10$m & mean & median & P95 & $\Delta\Omega$ & fail & $>10$m \\\midrule
LQR & 343.1 & 330.2 & 522.5 & 12.2 & 0 & 0 & 396.3 & 404.1 & 559.9 & 12.3 & 0 & 0 \\
MRAC & 93.0 & 91.8 & 124.1 & 21.6 & 0 & 0 & 116.6 & 111.9 & 157.2 & 25.1 & 2 & 0 \\
NN-MRAC & 68.6 & 67.4 & 84.1 & 13.7 & 0 & 0 & 86.3 & 84.4 & 100.5 & 15.0 & 0 & 0 \\
CL-att G & 50.7 & 49.5 & 61.6 & 16.3 & 0 & 0 & 57.5 & 55.7 & 71.6 & 19.2 & 0 & 0 \\
CL-Inner G & 38.0 & 37.0 & 44.8 & 18.1 & 0 & 0 & {\bfseries 49.3} & 48.7 & {\bfseries 67.9} & 22.4 & 0 & 0 \\
{\bfseries CL-Both G} & {\bfseries 31.5} & {\bfseries 30.9} & {\bfseries 35.4} & 19.2 & 0 & 0 & 49.8 & {\bfseries 47.6} & 80.8 & 23.6 & 0 & 0 \\
GEO-DOB & 226.9 & 67.5 & 1153.6 & 14.5 & 3 & 0 & 692.2 & 96.5 & 3251.0 & 15.1 & 3 & 0 \\
INDI & 134.8 & 90.6 & 257.9 & 11.7 & 3 & 0 & 342.0 & 146.9 & 1485.3 & 11.8 & 2 & 0 \\
$\mathcal{L}_1$ & 103.7 & 89.7 & 117.6 & 13.0 & 5 & 0 & 100.4 & 100.7 & 127.1 & 13.0 & 11 & 0 \\
\bottomrule\end{tabular}\par\smallskip\parbox{\textwidth}{\footnotesize Noise seeds 1001--1050 and 2001--2050, reused across the earlier analyses; main comparison, 50 noisy draws per scenario, RMSE$_5$ in mm, activity in rad/s. Fail counts incomplete runs; $>10$m counts completed accuracy failures. Summaries condition on completion. $\Delta\Omega$ is mean maneuver activity, not an S7-budget pass count. S: seed-1, G: guarded selection.}\end{table*}

\begin{table*}[t]\centering\caption{Paired Monte Carlo results of all profiles (development draws)}\label{tab:mc-full}\small\setlength{\tabcolsep}{4pt}
\begin{tabular}{lrrrrrrrrrrrr}\toprule & \multicolumn{6}{c}{S2 + noise} & \multicolumn{6}{c}{S4 + noise}\\\cmidrule(lr){2-7}\cmidrule(lr){8-13}Controller & mean & median & P95 & $\Delta\Omega$ & fail & $>10$m & mean & median & P95 & $\Delta\Omega$ & fail & $>10$m \\\midrule
LQR & 343.1 & 330.2 & 522.5 & 12.2 & 0 & 0 & 396.3 & 404.1 & 559.9 & 12.3 & 0 & 0 \\
MRAC & 93.0 & 91.8 & 124.1 & 21.6 & 0 & 0 & 116.6 & 111.9 & 157.2 & 25.1 & 2 & 0 \\
B1 & 345.2 & 400.5 & 838.4 & 28.0 & 43 & 0 & 214.3 & 162.4 & 406.3 & 18.3 & 40 & 0 \\
NN-MRAC & 68.6 & 67.4 & 84.1 & 13.7 & 0 & 0 & 86.3 & 84.4 & 100.5 & 15.0 & 0 & 0 \\
CL-ucmd S & 75.6 & 76.0 & 79.3 & 16.8 & 0 & 0 & 100.3 & 98.1 & 123.5 & 19.6 & 0 & 0 \\
CL-att S & 47.1 & 45.7 & 56.9 & 16.7 & 0 & 0 & 54.8 & 53.7 & 69.1 & 20.0 & 0 & 0 \\
CL-Outer S & 47.1 & 45.7 & 56.9 & 16.7 & 0 & 0 & 55.0 & 53.9 & 68.8 & 20.0 & 0 & 0 \\
CL-Inner S & 31.2 & 30.6 & 35.2 & 19.3 & 0 & 0 & 53.5 & 48.3 & 121.3 & 24.1 & 0 & 0 \\
{\bfseries CL-Both S} & {\bfseries 29.8} & {\bfseries 29.4} & {\bfseries 34.0} & 19.5 & 0 & 0 & 62.4 & 49.0 & 161.2 & 24.2 & 0 & 0 \\
CL-att G & 50.7 & 49.5 & 61.6 & 16.3 & 0 & 0 & 57.5 & 55.7 & 71.6 & 19.2 & 0 & 0 \\
CL-Outer G & 50.8 & 49.6 & 61.5 & 16.2 & 0 & 0 & 57.7 & 55.9 & 72.0 & 19.2 & 0 & 0 \\
CL-Inner G & 38.0 & 37.0 & 44.8 & 18.1 & 0 & 0 & {\bfseries 49.3} & 48.7 & {\bfseries 67.9} & 22.4 & 0 & 0 \\
{\bfseries CL-Both G} & 31.5 & 30.9 & 35.4 & 19.2 & 0 & 0 & 49.8 & {\bfseries 47.6} & 80.8 & 23.6 & 0 & 0 \\
GEO-DOB & 226.9 & 67.5 & 1153.6 & 14.5 & 3 & 0 & 692.2 & 96.5 & 3251.0 & 15.1 & 3 & 0 \\
INDI & 134.8 & 90.6 & 257.9 & 11.7 & 3 & 0 & 342.0 & 146.9 & 1485.3 & 11.8 & 2 & 0 \\
$\mathcal{L}_1$ & 103.7 & 89.7 & 117.6 & 13.0 & 5 & 0 & 100.4 & 100.7 & 127.1 & 13.0 & 11 & 0 \\
\bottomrule\end{tabular}\par\smallskip\parbox{\textwidth}{\footnotesize Draws as in Table~\ref{tab:mc-dev}. Fail counts incomplete runs; $>10$m counts completed accuracy failures. Summaries condition on completion. $\Delta\Omega$ is mean maneuver activity, not an S7-budget pass count. S: seed-1, G: guarded selection.}\end{table*}

\subsection{Guard With 40 Training Seeds}

In a check that changes no selection (planned in writing before the runs; the plan is in the supplementary package), every candidate of the guarded selection was run on 20 further S7 seeds (4021--4040); seeds 4001 and 4002 were rerun as a reproduction check and match the archived activity exactly. Table~\ref{tab:r2-guard} lists the guard bound $\max(\max\Delta\Omega,\ \mathrm{mean}+3\,\mathrm{SD})$ over 20 and 40 seeds with a bootstrap standard error (2000 resamples of the seeds). No pass/fail decision changes, so no selected profile changes. The largest change of a bound is 0.031\,rad/s.

\begin{table}[!t]\centering\caption{Guard bound (rad/s) over 20 and 40 training seeds}\label{tab:r2-guard}
\footnotesize\setlength{\tabcolsep}{3pt}\begin{tabular}{@{}llrrrrc@{}}\toprule Family & cand. & bound$_{20}$ & bound$_{40}$ & SE$_{40}$ & margin$_{40}$ & pass \\\midrule
CL-att & 40 & 12.592 & 12.604 & 0.016 & +0.423 & yes \\
CL-att & 80 & 12.712 & 12.719 & 0.014 & +0.307 & yes \\
CL-att & 160 & 13.100 & 13.113 & 0.014 & -0.087 & no \\
CL-Outer & 40 & 12.573 & 12.584 & 0.016 & +0.443 & yes \\
CL-Outer & 80 & 12.680 & 12.688 & 0.014 & +0.339 & yes \\
CL-Outer & 160 & 13.061 & 13.080 & 0.016 & -0.053 & no \\
CL-Inner & 160 & 12.713 & 12.727 & 0.014 & +0.300 & yes \\
CL-Inner & 320 & 12.862 & 12.886 & 0.016 & +0.141 & yes \\
CL-Inner & 640 & 13.040 & 13.071 & 0.020 & -0.044 & no \\
CL-Both & 640 & 12.805 & 12.835 & 0.020 & +0.192 & yes \\
CL-Both & 1280 & 12.925 & 12.951 & 0.025 & +0.076 & yes \\
CL-Both & 2560 & 13.142 & 13.160 & 0.035 & -0.133 & no \\
CL-LPF & 640 & 12.822 & 12.852 & 0.019 & +0.175 & yes \\
CL-LPF & 1280 & 12.956 & 12.981 & 0.025 & +0.046 & yes \\
CL-LPF & 2560 & 13.163 & 13.179 & 0.036 & -0.153 & no \\
GEO-DOB & 654 & 13.153 & 13.138 & 0.013 & -0.111 & no \\
GEO-DOB & 924 & 11.365 & 11.357 & 0.012 & +1.670 & yes \\
INDI & 50 & 13.077 & 13.070 & 0.025 & -0.043 & no \\
INDI & 71 & 11.947 & 11.944 & 0.017 & +1.083 & yes \\
INDI-RPM & 50 & 13.077 & 13.070 & 0.024 & -0.044 & no \\
INDI-RPM & 71 & 11.947 & 11.945 & 0.017 & +1.082 & yes \\
$\mathcal{L}_1$ & 36 & 13.078 & 13.078 & 0.015 & -0.051 & no \\
\bottomrule\end{tabular}\par\smallskip\parbox{\columnwidth}{\footnotesize Budget 13.027\,rad/s. Composite families: all three gains $\gamma_o$; comparators: the guarded candidate and the failing candidate closest to the budget (index into the seed-1 grid). SE: bootstrap standard error. Margin: budget minus bound; positive passes. Pass decisions over 20 and 40 seeds are identical for all 55 candidates (full list in the supplementary package).}\end{table}

\subsection{\texorpdfstring{$\mathcal{L}_1$}{L1} Near the Budget}

$\mathcal{L}_1$ has no guarded profile because none of its ten candidates passes the guard. Table~\ref{tab:r2-l1} evaluates the seed-1 candidate (index 35) and the candidate with the smallest guard bound (index 36) with the protocols of the main comparison; they are not admissible under the guard and are not part of the guarded ranking. Both reach a lower Eval6 than guarded GEO-DOB (120.5\,mm) and a higher one than every composite profile; both fail to complete some fresh Monte Carlo draws, with more failures than any guarded profile.

\begin{table}[!t]\centering\caption{$\mathcal{L}_1$ candidates near the budget}\label{tab:r2-l1}
\footnotesize\setlength{\tabcolsep}{4pt}\begin{tabular}{@{}lrr@{}}\toprule & idx 35 (seed-1) & idx 36 \\\midrule
Guard bound (20 / 40 seeds) & 13.117 / 13.116 & 13.078 / 13.078 \\
Design cost $J$ (S2--S4) & 100.8 & 101.2 \\
Eval6, seed 1 & 111.3 & 113.2 \\
Eval6, seeds 6001--6010 (mean$\pm$SD) & 111.3$\pm$0.05 & 113.2$\pm$0.05 \\
S7 validation pass (fixed budget) & 95/100 & 100/100 \\
S7 within $1.05\times$ paired LQR & 99/100 & 100/100 \\
Median paired S7 ratio to LQR & 1.046 & 1.043 \\
Fresh MC S2: incomplete / 50 & 6 & 2 \\
Fresh MC S2: median, P95 & 92.5, 116.7 & 93.3, 131.8 \\
Fresh MC S4: incomplete / 50 & 3 & 1 \\
Fresh MC S4: median, P95 & 98.1, 119.7 & 99.5, 126.2 \\
\bottomrule\end{tabular}\par\smallskip\parbox{\columnwidth}{\footnotesize Not admissible under the guard. RMSE in mm, activity in rad/s. S7 validation: seeds 5001--5100 (the 92/100 of Section~\mref{sec:tuning} refers to seeds 3001--3100). Fresh Monte Carlo: the draws 9001--9100 of Table~\mref{tab:mc}, run after that study; medians and P95 over completed runs (P95 by the MATLAB convention). Reference values in Table~\mref{tab:mc}: CL-Both G median 31.1/46.6\,mm, GEO-DOB G 76.8/113.5\,mm (S2/S4).}\end{table}

\section{Activity Budget and Gain Effects}\label{app:budgetdetails}

\subsection{Budget Sweeps}

The budget sweep changes the admissibility threshold while retaining the original candidate grids. Table~\ref{tab:budget} lists the resulting best design cost of MRAC, NN-MRAC, CL-ucmd, CL-att and CL-Both for S7 budgets from 1.00 to 1.50 times LQR activity and complements Fig.~\mref{fig:pareto}(a). Table~\ref{tab:dual} adds an in-flight activity constraint to the selection (Section~\mref{sec:budget}).

\begin{table}[!t]
\centering\caption{Best design cost $J$ (mm) versus the S7 activity budget $b\times$ LQR}\label{tab:budget}
\renewcommand{\arraystretch}{1.1}\footnotesize\setlength{\tabcolsep}{4pt}
\begin{tabular}{@{}cccccc@{}}
\toprule
$b$ & MRAC & NN-MRAC & CL-ucmd & CL-att & {\bfseries CL-Both}\\ \midrule
1.00 & \val{pb100}{MRAC}{J} & \val{pb100}{NN-MRAC}{J} & \val{pb100}{CL-ucmd}{J} & \val{pb100}{CL-att}{J} & {\bfseries \val{pb100}{PROPOSED}{J}} \\
1.05 & \val{pb105}{MRAC}{J} & \val{pb105}{NN-MRAC}{J} & \val{pb105}{CL-ucmd}{J} & \val{pb105}{CL-att}{J} & {\bfseries \val{pb105}{PROPOSED}{J}} \\
1.15 & \val{pb115}{MRAC}{J} & \val{pb115}{NN-MRAC}{J} & \val{pb115}{CL-ucmd}{J} & \val{pb115}{CL-att}{J} & {\bfseries \val{pb115}{PROPOSED}{J}} \\
1.25 & \val{pb125}{MRAC}{J} & \val{pb125}{NN-MRAC}{J} & \val{pb125}{CL-ucmd}{J} & \val{pb125}{CL-att}{J} & {\bfseries \val{pb125}{PROPOSED}{J}} \\
1.35 & \val{pb135}{MRAC}{J} & \val{pb135}{NN-MRAC}{J} & \val{pb135}{CL-ucmd}{J} & \val{pb135}{CL-att}{J} & {\bfseries \val{pb135}{PROPOSED}{J}} \\
1.40 & \val{pb140}{MRAC}{J} & \val{pb140}{NN-MRAC}{J} & \val{pb140}{CL-ucmd}{J} & {\bfseries \val{pb140}{CL-att}{J}} & \val{pb140}{PROPOSED}{J} \\
1.50 & \val{pb150}{MRAC}{J} & \val{pb150}{NN-MRAC}{J} & \val{pb150}{CL-ucmd}{J} & {\bfseries \val{pb150}{CL-att}{J}} & \val{pb150}{PROPOSED}{J} \\
\bottomrule
\multicolumn{6}{@{}l}{\footnotesize --: no admissible configuration. The paper uses $b=1.05$.}\\
\multicolumn{6}{@{}l}{\footnotesize $b=1.00$, CL-Both: $\gamma_o=$~\val{pb100cfg}{PROPOSED}{go}, $k_c=$~\val{pb100cfg}{PROPOSED}{kc}, $\lambda=$~\val{pb100cfg}{PROPOSED}{lam}\,s$^{-1}$, $\ell_o=$~\val{pb100cfg}{PROPOSED}{lo},}\\
\multicolumn{6}{@{}l}{\footnotesize $\Delta\Omega=$~\val{pb100cfg}{PROPOSED}{act}\,rad/s (LQR \val{pb100cfg}{all}{lqr}\,rad/s).}
\end{tabular}
\end{table}

\begin{table}[t]\centering\caption{Selection with an additional in-flight activity constraint}\label{tab:dual}
\small\setlength{\tabcolsep}{3pt}\begin{tabular}{lrrrrrr}\toprule & \multicolumn{2}{c}{S7 only} & \multicolumn{2}{c}{+NN-MRAC} & \multicolumn{2}{c}{+1.2 LQR}\\\cmidrule(lr){2-3}\cmidrule(lr){4-5}\cmidrule(lr){6-7}Family & $J$ & $\bar{\Delta\Omega}$ & $J$ & $\bar{\Delta\Omega}$ & $J$ & $\bar{\Delta\Omega}$ \\\midrule
MRAC & 86.5 & 2.40 & 118.5 & 1.06 & 150.3 & 0.96 \\
NN-MRAC & 82.6 & 1.25 & 82.6 & 1.25 & 134.9 & 0.94 \\
CL-ucmd S & 77.8 & 1.42 & 84.8 & 1.24 & -- & -- \\
CL-att S & 46.2 & 1.26 & 49.6 & 1.13 & 65.0 & 1.02 \\
CL-Outer & 46.2 & 1.27 & 49.6 & 1.14 & 71.2 & 0.99 \\
CL-Inner & 37.3 & 1.51 & 43.6 & 1.18 & {\bfseries 60.0} & 0.98 \\
{\bfseries CL-Both} & {\bfseries 35.8} & 1.60 & {\bfseries 43.5} & 1.19 & {\bfseries 60.0} & 0.98 \\
GEO-DOB & 57.0 & 1.11 & 57.0 & 1.11 & 69.6 & 0.70 \\
INDI & 101.6 & 0.50 & 101.6 & 0.50 & 101.6 & 0.50 \\
$\mathcal{L}_1$ & 100.8 & 0.88 & 100.8 & 0.88 & 100.8 & 0.88 \\
\bottomrule\end{tabular}\par\smallskip\parbox{\columnwidth}{\footnotesize $J$ (mm) and mean S2--S4 activity $\bar{\Delta\Omega}$ (rad/s) of the best candidate with S7 activity $\le1.05\times$LQR alone, and additionally with $\bar{\Delta\Omega}\le$ 1.25 rad/s (tuned NN-MRAC) or $\le$ 1.02 rad/s ($1.2\times$LQR). Full grids of Section~\mref{sec:fair}, seed 1. --: no candidate satisfies both constraints.}\end{table}

\subsection{Common-Gain Test and Gain--Accuracy Trade-Off}

\begin{figure}[!t]\centering\includegraphics[width=\columnwidth]{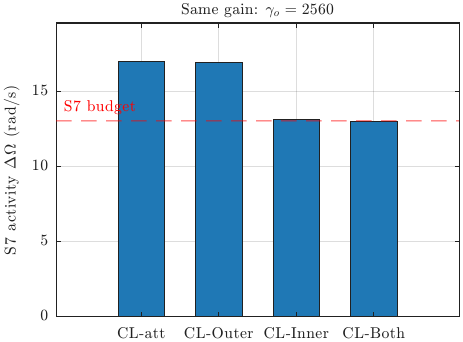}
\caption{Common-gain S7 predictor test. The inner predictor supplies most activity reduction; both predictors cross the budget at gain 2560.}\label{fig:unified-factorial}\end{figure}

Fig.~\ref{fig:unified-factorial} shows the common-gain S7 test of the four target configurations: the inner predictor provides most of the activity reduction, and with both predictors the activity crosses the budget at gain 2560. Table~\ref{tab:gain} lists the gain--accuracy trade-off referred to in Section~\mref{sec:extension}. At high gain the tracking term dominates the composite term and absorbs lag and fast disturbances; lower-gain CL-att estimates the physical uncertainty more accurately, and rapid disturbance components that do not depend on the state are only partly rejected.

\begin{table}[!t]
\centering\caption{Gain--accuracy trade-off versus $\gamma_o$}\label{tab:gain}
\renewcommand{\arraystretch}{1.1}\footnotesize
\begin{tabular}{@{}llccccc@{}}
\toprule
 & $\gamma_o$ & 40 & 160 & 640 & 2560 & 5120\\ \midrule
\multirow{2}{*}{RMSE$_5$ (mm)} & CL-att & \val{gt40}{CL-att}{RMSE5} & \val{gt160}{CL-att}{RMSE5} & \val{gt640}{CL-att}{RMSE5} & \val{gt2560}{CL-att}{RMSE5} & \val{gt5120}{CL-att}{RMSE5}\\
 & CL-Both & \val{gt40}{PROPOSED}{RMSE5} & \val{gt160}{PROPOSED}{RMSE5} & \val{gt640}{PROPOSED}{RMSE5} & \val{gt2560}{PROPOSED}{RMSE5} & \val{gt5120}{PROPOSED}{RMSE5}\\
\multirow{2}{*}{$\hat {\bv{d}}$ error (m/s$^2$)} & CL-att & \val{gt40}{CL-att}{DHATP} & \val{gt160}{CL-att}{DHATP} & \val{gt640}{CL-att}{DHATP} & \val{gt2560}{CL-att}{DHATP} & \val{gt5120}{CL-att}{DHATP}\\
 & CL-Both & \val{gt40}{PROPOSED}{DHATP} & \val{gt160}{PROPOSED}{DHATP} & \val{gt640}{PROPOSED}{DHATP} & \val{gt2560}{PROPOSED}{DHATP} & \val{gt5120}{PROPOSED}{DHATP}\\
\multirow{2}{*}{$\Delta\Omega$ S7 (rad/s)} & CL-att & \val{gt40}{CL-att}{DW} & \val{gt160}{CL-att}{DW} & \val{gt640}{CL-att}{DW} & \val{gt2560}{CL-att}{DW} & \val{gt5120}{CL-att}{DW}\\
 & CL-Both & \val{gt40}{PROPOSED}{DW} & \val{gt160}{PROPOSED}{DW} & \val{gt640}{PROPOSED}{DW} & \val{gt2560}{PROPOSED}{DW} & \val{gt5120}{PROPOSED}{DW}\\ \bottomrule
\multicolumn{7}{@{}l}{\footnotesize Budget: $\Delta\Omega\le$~\val{tune}{all}{budget}\,rad/s in S7.}
\end{tabular}
\par\smallskip\parbox{\columnwidth}{\footnotesize S4 tracking and $\bv{d}^{\rm phys}$-estimation error and S7 activity; $\gamma_i=10\gamma_o$, other parameters as tuned.}
\end{table}

\subsection{Excitation and Weight Motion}

\begin{figure*}[t]\centering\includegraphics[width=.85\textwidth]{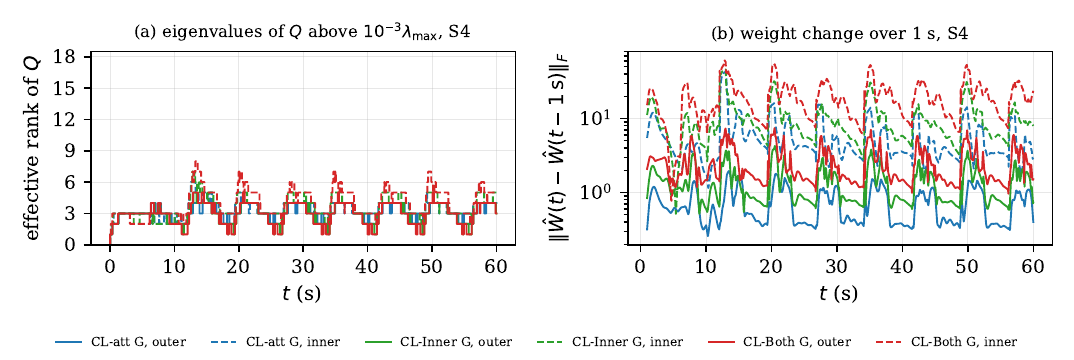}
\caption{Excitation and learning in S4 for the guarded composite profiles. (a) Number of eigenvalues of $\bv{Q}_o$ (of 15) and $\bv{Q}_i$ (of 18) above $10^{-3}\lambda_{\max}$. (b) Weight change over the preceding second, $\|\hat {\bv{W}}(t)-\hat {\bv{W}}(t-1\,\mathrm{s})\|_F$.}\label{fig:excitation}\end{figure*}

To separate learning from high-gain compensation, the information matrices and weights of the guarded CL-att, CL-Inner and CL-Both profiles were logged in S2 and S4 (Fig.~\ref{fig:excitation}). After the first 5\,s, the median number of eigenvalues of $\bv{Q}_o$ above $10^{-3}\lambda_{\max}(\bv{Q}_o)$ is \val{exc}{GUARDbothS4}{MedRankQo} of 15 for CL-Both in S4 (maximum \val{exc}{GUARDbothS4}{MaxRankQo}), and of $\bv{Q}_i$ \val{exc}{GUARDbothS4}{MedRankQi} of 18 (maximum \val{exc}{GUARDbothS4}{MaxRankQi}); $\lambda_{\min}(\bv{Q}_o)$ never exceeds $\val{exc}{GUARDbothS4}{MaxLamQo}$, and the other profiles and S2 behave alike. With a 0.5\,s forgetting window and 15- and 18-dimensional regressors, the trajectories excite only a few weight directions, so the condition $\bv{Q}\succeq\sigma \bv{I}$ of Theorem~\mref{thm:1} holds with no useful $\sigma>0$, and the composite term damps only a low-dimensional subspace. The weights nevertheless move quickly: over one second the inner-loop change reaches \val{exc}{GUARDbothS4}{MaxDWi} for CL-Both in S4, against \val{exc}{GUARDinnerS4}{MaxDWi} for CL-Inner and \val{exc}{GUARDnoneS4}{MaxDWi} for CL-att, and it grows with the adaptation gain. The data therefore support reading the controllers as high-gain adaptive compensators that track the current input deficit along the excited directions, not as estimators that converge to a fixed model of the uncertainty. This agrees with Table~\ref{tab:gain}.

\subsection{Activity Spectra and Alternative Metrics}\label{app:act}

\begin{figure}[t]\centering\includegraphics[width=\columnwidth]{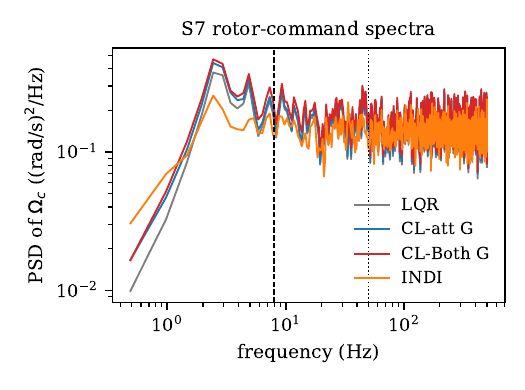}
\caption{Welch power spectral density of the rotor commands in S7 (seed 1, 1\,kHz) for LQR and selected profiles.}\label{fig:psd}\end{figure}

For all main-text profiles, \val{rIIE}{psd}{fcmin}--\val{rIIE}{psd}{fcmax}\,\% of the S7 command variance lies above the motor corner frequency of 7.96\,Hz and \val{rIIE}{psd}{fiftymin}--\val{rIIE}{psd}{fiftymax}\,\% above 50\,Hz (Fig.~\ref{fig:psd}), so the budget measures noise transmission that the motors largely filter. Total variation after down-sampling to 250\,Hz and the RMS derivative of the motor-filtered command rank the profiles as $\Delta\Omega$ does (Spearman correlation \val{rIIE}{act}{spearmin}). The second of these metrics is the RMS rotor acceleration demanded by the commands, a proxy for the inertial part of the motor torque (drag torque and the ESC also shape the motor current); relative to LQR it is \val{rV}{act}{bothLPF} for CL-Both G, \val{rV}{act}{lpfLPF} for CL-LPF G and \val{rV}{act}{geoLPF} for GEO-DOB, against \val{rIIE}{act}{bothDW} for CL-Both G on $\Delta\Omega$.

\section{Motor Model, Target Filtering and Sampling}\label{app:motor}

\subsection{Motor Model Versus Target Filtering}

\begin{figure*}[t]\centering\includegraphics[width=.9\textwidth]{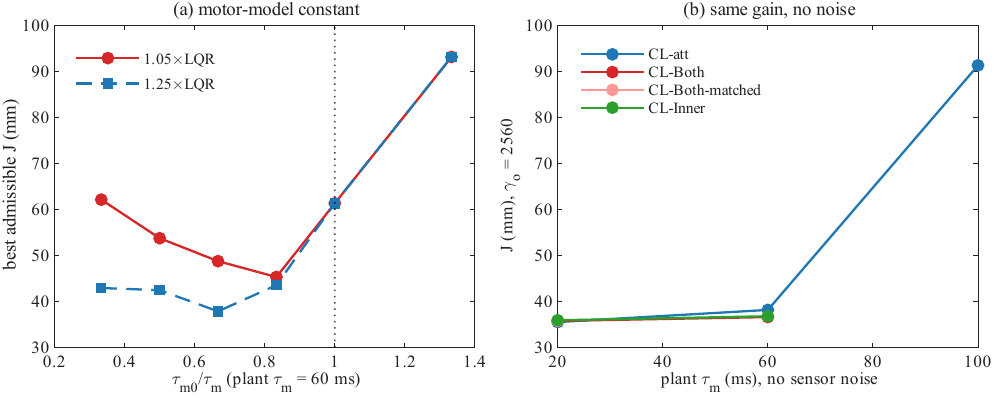}
\caption{Motor model versus target filtering (seed 1). (a) Plant motor constant 60\,ms: best admissible design cost against the ratio of the controller motor-model constant to the plant constant, for budgets $1.05\times$ and $1.25\times$ LQR. (b) Nominal and slow plants without sensor noise: design cost at $\gamma_o=2560$; missing points diverged.}\label{fig:e1}\end{figure*}

Fig.~\ref{fig:e1} separates the motor model from the low-pass filtering it implies (Section~\mref{sec:extension}): panel (a) varies the motor-model constant of the controller for a plant with a 60\,ms motor constant, and panel (b) compares the nominal and slow plants without sensor noise. Table~\ref{tab:a4} sweeps the motor-model constant $\tau_{m0}$ of CL-Both for plants with 20 and 60\,ms, where $\tau_{m0}=0$ is CL-att, and Table~\ref{tab:taum} gives the S4 error of MRAC, NN-MRAC, CL-ucmd, CL-att and CL-Both when the plant motor constant varies from 10 to 80\,ms.

\begin{table}[!t]
\centering\caption{CL-Both with motor-model constant $\tau_{m0}$}\label{tab:a4}
\renewcommand{\arraystretch}{1.1}\footnotesize\setlength{\tabcolsep}{4pt}
\begin{tabular}{@{}cccccc@{}}
\toprule
 & \multicolumn{3}{c}{plant $\tau_m=20$\,ms} & \multicolumn{2}{c}{plant $\tau_m=60$\,ms}\\
\cmidrule(lr){2-4}\cmidrule(l){5-6}
$\tau_{m0}$ (ms) & $J_{1.05}$ & $J_{1.25}$ & $\Delta\Omega$ S7 & $J_{1.05}$ & $J_{1.25}$\\ \midrule
0 (CL-att) & \val{fa20}{lp0}{J105} & \val{fa20}{lp0}{J125} & \val{fa20}{lp0}{DWg} & \val{fa60}{lp0}{J105} & \val{fa60}{lp0}{J125}\\
5 & \val{fa20}{lp5}{J105} & \val{fa20}{lp5}{J125} & \val{fa20}{lp5}{DWg} & \val{fa60}{lp5}{J105} & \val{fa60}{lp5}{J125}\\
10 & \val{fa20}{lp10}{J105} & \val{fa20}{lp10}{J125} & \val{fa20}{lp10}{DWg} & \val{fa60}{lp10}{J105} & \val{fa60}{lp10}{J125}\\
15 & \val{fa20}{lp15}{J105} & \val{fa20}{lp15}{J125} & \val{fa20}{lp15}{DWg} & \val{fa60}{lp15}{J105} & \val{fa60}{lp15}{J125}\\
20 & \val{fa20}{lp20}{J105} & \val{fa20}{lp20}{J125} & \val{fa20}{lp20}{DWg} & \val{fa60}{lp20}{J105} & \val{fa60}{lp20}{J125}\\
25 & \val{fa20}{lp25}{J105} & \val{fa20}{lp25}{J125} & \val{fa20}{lp25}{DWg} & \val{fa60}{lp25}{J105} & \val{fa60}{lp25}{J125}\\
30 & \val{fa20}{lp30}{J105} & \val{fa20}{lp30}{J125} & \val{fa20}{lp30}{DWg} & \val{fa60}{lp30}{J105} & \val{fa60}{lp30}{J125}\\
40 & \val{fa20}{lp40}{J105} & \val{fa20}{lp40}{J125} & \val{fa20}{lp40}{DWg} & \val{fa60}{lp40}{J105} & \val{fa60}{lp40}{J125}\\
60 & \val{fa20}{lp60}{J105} & \val{fa20}{lp60}{J125} & \val{fa20}{lp60}{DWg} & \val{fa60}{lp60}{J105} & \val{fa60}{lp60}{J125}\\
80 & \val{fa20}{lp80}{J105} & \val{fa20}{lp80}{J125} & \val{fa20}{lp80}{DWg} & \val{fa60}{lp80}{J105} & \val{fa60}{lp80}{J125}\\
\bottomrule
\multicolumn{6}{@{}l}{\footnotesize $\tau_{m0}=20$\,ms: CL-Both design. 50\,ms is not on this grid;}\\
\multicolumn{6}{@{}l}{\footnotesize Fig.~\ref{fig:e1}(a) gives 50\,ms for the 60\,ms plant. Gain swept over}\\
\multicolumn{6}{@{}l}{\footnotesize the grid, $k_c=0.25$, $\lambda=2$, $\ell=0$.}
\end{tabular}
\par\smallskip\parbox{\columnwidth}{\footnotesize Plant $\tau_m$ as stated. Best admissible cost $J$ (mm) for budgets $1.05\times$ and $1.25\times$ LQR, and S7 activity at $\gamma_o=2560$.}
\end{table}

\begin{table}[!t]
\centering\caption{RMSE$_5$ in S4 (mm) versus the plant motor time constant}\label{tab:taum}
\renewcommand{\arraystretch}{1.1}\footnotesize
\begin{tabular}{@{}lccccc@{}}
\toprule
$\tau_m$ (ms) & 10 & 20 & 40 & 60 & 80\\ \midrule
MRAC     & \val{tm10}{MRAC}{RMSE5} & \val{tm20}{MRAC}{RMSE5} & \val{tm40}{MRAC}{RMSE5} & \val{tm60}{MRAC}{RMSE5} & \val{tm80}{MRAC}{RMSE5}\\
NN-MRAC  & \val{tm10}{NN-MRAC}{RMSE5} & \val{tm20}{NN-MRAC}{RMSE5} & \val{tm40}{NN-MRAC}{RMSE5} & \val{tm60}{NN-MRAC}{RMSE5} & \val{tm80}{NN-MRAC}{RMSE5}\\
CL-ucmd  & \val{tm10}{CL-ucmd}{RMSE5} & \val{tm20}{CL-ucmd}{RMSE5} & \val{tm40}{CL-ucmd}{RMSE5} & \val{tm60}{CL-ucmd}{RMSE5} & \val{tm80}{CL-ucmd}{RMSE5}\\
CL-att   & \val{tm10}{CL-att}{RMSE5} & \val{tm20}{CL-att}{RMSE5} & \val{tm40}{CL-att}{RMSE5} & \val{tm60}{CL-att}{RMSE5} & \val{tm80}{CL-att}{RMSE5}\\
CL-Both & \val{tm10}{PROPOSED}{RMSE5} & \val{tm20}{PROPOSED}{RMSE5} & \val{tm40}{PROPOSED}{RMSE5} & \val{tm60}{PROPOSED}{RMSE5} & \val{tm80}{PROPOSED}{RMSE5}\\ \bottomrule
\end{tabular}
\par\smallskip\parbox{\columnwidth}{\footnotesize Controllers assume 20\,ms.}
\end{table}

\subsection{Mis-Specified Motor Time Constant}

Table~\ref{tab:r2-taum} varies the plant motor constant $\tau_m$ while every controller keeps the nominal 20\,ms (in its motor model or, for CL-LPF, in $\tau_{\rm LP}$); seed 1, scenarios S2--S4 and S7 (planned in writing before the runs). Divergence occurs in S3 (doubled inertia and growing drag). The table also shows the gain 2560, at which an earlier run found CL-Both divergent with a 100\,ms plant: CL-LPF at 2560 also diverges there and already at 80\,ms.

\begin{table*}[!t]\centering\caption{Design cost and S7 activity for a mis-specified plant motor constant}\label{tab:r2-taum}
\footnotesize\setlength{\tabcolsep}{4pt}\begin{tabular}{@{}rrrrrrrrrrrr@{}}\toprule
$\tau_m$ (ms) & \multicolumn{2}{c}{CL-LPF 1280} & \multicolumn{2}{c}{CL-LPF 2560} & \multicolumn{2}{c}{CL-Both 1280} & \multicolumn{2}{c}{CL-Both 2560} & \multicolumn{2}{c}{CL-att 80} & LQR $J$\\
 & $J$ & S7 & $J$ & S7 & $J$ & S7 & $J$ & S7 & $J$ & S7 & \\\midrule
10 & 37.3 & 0.96 & 35.8 & 0.95 & 37.3 & 0.96 & 35.9 & 0.94 & 49.1 & 1.00 & 340.0 \\
20 & 37.2 & 1.03 & 35.6 & 1.05 & 37.3 & 1.03 & 35.8 & 1.05 & 49.6 & 1.01 & 340.0 \\
40 & 37.0 & 1.33 & 35.4 & 1.38 & 37.1 & 1.32 & 35.6 & 1.38 & 50.9 & 1.07 & 340.0 \\
60 & 37.7 & 1.55 & 36.0 & 1.71 & 38.0 & 1.56 & 36.6 & 1.77 & 52.5 & 1.21 & 340.1 \\
80 & -- & 1.71 & -- & 1.71 & 39.8 & 1.81 & 38.7 & 1.85 & 53.8 & 1.52 & 341.1 \\
100 & -- & 1.90 & -- & 1.98 & -- & 1.83 & -- & 2.18 & 56.4 & 1.59 & 903.5 \\
\bottomrule\end{tabular}\par\smallskip\parbox{\textwidth}{\footnotesize Controllers assume 20\,ms. $J$ in mm (--: divergence in S3); S7 activity relative to LQR on the same plant.}
\end{table*}

\subsection{Linear Analysis of the Target-Filter Ablation}\label{app:r2-lin}

\begin{figure*}[!t]\centering\includegraphics[width=.85\textwidth]{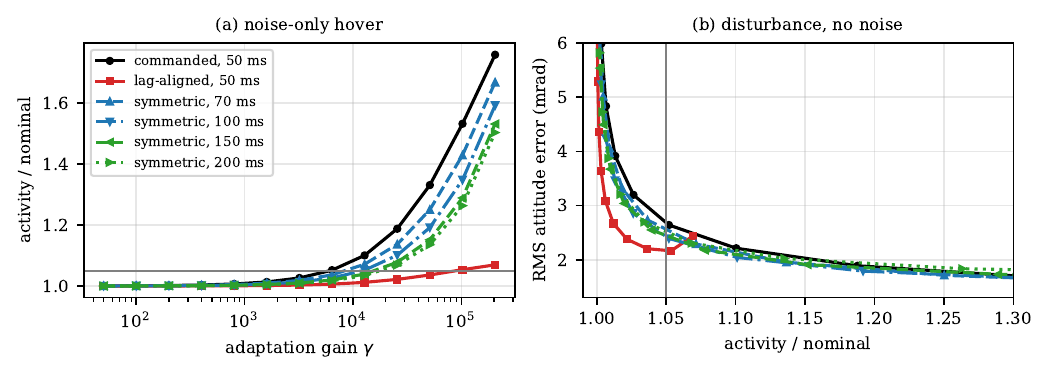}\caption{One-axis model of the target-filter ablation. (a) Activity relative to the nominal law versus adaptation gain. (b) RMS attitude error under a torque disturbance versus activity; the vertical line marks the budget of 1.05.}\label{fig:r2-lin}\end{figure*}

Consider one channel of the composite law with plant $\dot\nu=a_r+d$, first-order actuator $a_r=A(s)\,\alpha$, $A(s)=1/(\tau_m s+1)$, and the dirty-derivative target of Section~\mref{sec:cl}, $y=F(s)[s\nu_m]-F(s)[\alpha_{\rm eff}]$ with $F(s)=1/(\tau_f s+1)$ and measured $\nu_m=\nu+n$. For the commanded target ($\alpha_{\rm eff}=\alpha$),
\begin{equation}y=F(s)\,d-F(s)\,\bigl(1-A(s)\bigr)\alpha+F(s)\,s\,n,\label{eq:r2-res}\end{equation}
and, since $1-A(s)=\tau_m s\,A(s)$, the target contains a filtered derivative of the command in addition to the filtered uncertainty. The command contains the measurement noise transmitted by the nominal law ($\alpha\approx-k_2 n+\ldots$), so the residual returns that noise to the update, and through the update and $\hat W^\top\phi$ back to the command. A common filter $F$ multiplies both terms at each frequency. The residual is concentrated at high frequency and $d$ mostly is not, so a larger $\tau_f$ attenuates the residual more than $d$, but it also delays the whole target. With a lag model $A_0(s)$ in the target, $\alpha_{\rm eff}=A_0(s)\alpha$, the middle term of \eqref{eq:r2-res} becomes $F(s)\,(A(s)-A_0(s))\,\alpha$; it vanishes for a matched $A_0=A$, and only the measurement term $F\,s\,n$, common to all targets, remains. Since $\alpha\approx-k_2n+\ldots$, the residual and $F\,s\,n$ are correlated, and the noise that reaches the update depends on both; the simulation below checks the net effect.
To check that this mechanism reproduces the ordering of the target-filter ablation of Section~\mref{sec:extension}, one roll axis of the inner loop was simulated with the constants of this paper ($k_{1i}=100$, $k_{2i}=14.14$, $\tau_m=20$\,ms, $\lambda=2$\,s$^{-1}$, $k_c=0.25$, $1$\,kHz sampling, gyro and attitude noise of Section~\mref{sec:model}) and an affine regressor $[1,\ \omega/\omega_s,\ \alpha_{bl}/\alpha_s]$; activity is the RMS command increment in a 20\,s noisy hover and tracking the RMS attitude error under a torque disturbance without noise. At the budget of 1.05 times the activity of the nominal law (Fig.~\ref{fig:r2-lin}), the commanded target admits $\gamma=3200$ and reaches 3.20\,mrad; symmetric filtering with $\tau_f=70$, 100, 150 and 200\,ms admits 6400, 6400, 12800, 12800 and reaches 2.73, 2.86, 2.55, 2.70\,mrad; the lag-aligned target admits $\gamma=51200$ and reaches 2.20\,mrad, with an RMS lag residual of 0.002 against about 0.54\,rad/s$^2$ for the commanded targets. At a common gain the lag-aligned target tracks slightly worse than the commanded one, as in the factorial test of Section~\mref{sec:extension}: its benefit is the larger admissible gain. The model omits the cascade, the RBF terms and the nonlinear plant and serves only to explain the sign of that result.

\subsection{Sampling, Local Linearization and Computation}

For the seed-1 CL-Both profile, controller sampling at 2 and 4 ms with unchanged gains and the plant still at 1 kHz keeps every deterministic S2/S4 run bounded; the S4 error increases from \val{ts1S4}{PROPOSED}{RMSE5} to \val{ts4S4}{PROPOSED}{RMSE5} mm and the noisy-S4 median from \val{tsmc1}{PROPOSED}{med} to \val{tsmc4}{PROPOSED}{med} mm (Table~\ref{tab:ts}). A numerical local check complements these runs (Table~\ref{tab:linear}): the one-period map of the sampled cascade, plant RK4 included, is linearized at hover by central differences for LQR and the guarded composite profiles at $T_s=1$, 2 and 4\,ms. With weights frozen at the hover equilibrium, frozen at their values after S4 ($\|\hat {\bv{W}}\|_F$ up to \val{lin}{all}{wnorm}), or adapting, all spectral radii are at most \val{lin}{all}{rhomax}, a decay rate of \val{lin}{Ts1}{rate}\,s$^{-1}$ for the slowest mode at every $T_s$. With frozen weights the slowest mode decays at \val{lin}{Ts1}{ratefrozen}\,s$^{-1}$, the real part of the designed outer-loop pole pair; with adaptation a slower mode remains. The adaptive map has \val{lin}{all}{nunit} additional eigenvalues $\mu$ with $||\mu|-1|<10^{-6}$; including them, its spectral radius is one to within $4\times10^{-13}$. Their number equals $3\,(14+17)$, the number of weight directions orthogonal to the constant hover regressor, which would neither affect the control nor be damped by the composite term. The eigenvectors were not computed, so this identification rests on the count alone. The check is local and numerical (Section~\mref{sec:limits}).

\begin{table}[!t]
\centering\caption{Effect of the controller sampling period (seed-1 CL-Both)}\label{tab:ts}
\renewcommand{\arraystretch}{1.1}\footnotesize
\setlength{\tabcolsep}{4pt}
\begin{tabular}{@{}ccccccc@{}}
\toprule
 & \multicolumn{3}{c}{CL-Both} & \multicolumn{3}{c}{MC S4, median / wins}\\
\cmidrule(lr){2-4}\cmidrule(l){5-7}
$T_s$ (ms) & S2 & S4 & TV S4 & CL-Both & CL-att & wins$^\ast$\\ \midrule
1 & \val{ts1S2}{PROPOSED}{RMSE5} & \val{ts1S4}{PROPOSED}{RMSE5} & \val{ts1S4}{PROPOSED}{TV} & \val{tsmc1}{PROPOSED}{med} & \val{tsmc1}{CL-att}{med} & \val{MCS4}{CL-att}{wins}\\
2 & \val{ts2S2}{PROPOSED}{RMSE5} & \val{ts2S4}{PROPOSED}{RMSE5} & \val{ts2S4}{PROPOSED}{TV} & \val{tsmc2}{PROPOSED}{med} & \val{tsmc2}{CL-att}{med} & \val{tsmc2}{CL-att}{wins}\\
4 & \val{ts4S2}{PROPOSED}{RMSE5} & \val{ts4S4}{PROPOSED}{RMSE5} & \val{ts4S4}{PROPOSED}{TV} & \val{tsmc4}{PROPOSED}{med} & \val{tsmc4}{CL-att}{med} & \val{tsmc4}{CL-att}{wins}\\
\bottomrule
\multicolumn{7}{@{}l}{\footnotesize $^\ast$runs (of \val{MC}{all}{N}) in which CL-Both is better than CL-att.}
\end{tabular}
\par\smallskip\parbox{\columnwidth}{\footnotesize Gains tuned at 1\,ms. RMSE$_5$ (mm) and rotor-command total variation TV ((rad/s)/s); randomized S4 runs with noise.}
\end{table}

\begin{table}[t]\centering\caption{Local check at hover: spectral radius of the one-period Jacobian}\label{tab:linear}
\small\setlength{\tabcolsep}{3pt}\begin{tabular}{lrrrrr}\toprule Controller & $T_s$ (ms) & frozen & S4 weights & adaptive & $n_1$ \\\midrule
LQR & 1 & 0.99859 & -- & -- & 0 \\
LQR & 2 & 0.99717 & -- & -- & 0 \\
LQR & 4 & 0.99436 & -- & -- & 0 \\
CL-att G & 1 & 0.99859 & 0.99859 & 0.99875 & 93 \\
CL-att G & 2 & 0.99717 & 0.99717 & 0.99750 & 93 \\
CL-att G & 4 & 0.99436 & 0.99436 & 0.99500 & 93 \\
CL-Inner G & 1 & 0.99859 & 0.99859 & 0.99874 & 93 \\
CL-Inner G & 2 & 0.99717 & 0.99717 & 0.99749 & 93 \\
CL-Inner G & 4 & 0.99436 & 0.99436 & 0.99497 & 93 \\
CL-Both G & 1 & 0.99859 & 0.99859 & 0.99872 & 93 \\
CL-Both G & 2 & 0.99717 & 0.99717 & 0.99745 & 93 \\
CL-Both G & 4 & 0.99436 & 0.99436 & 0.99490 & 93 \\
\bottomrule\end{tabular}\par\smallskip\parbox{\columnwidth}{\footnotesize $T_s$: controller period. Frozen: $\hat {\bv{W}},\bv{Q},\bv{\mathcal R}$ fixed at the hover equilibrium; S4 weights: $\hat {\bv{W}}$ frozen at its value after S4; adaptive: full map without the $n_1$ eigenvalues with $||\mu|-1|<10^{-6}$ (with them, the radius is one to within $4\times10^{-13}$). Central-difference Jacobians of the sampled map (plant RK4 at 1 ms, zero-order hold); numerical evidence, not a proof.}\end{table}

The dominant operation is an $18\times18$ symmetric solve. Desktop step times are \val{time}{PROPOSED}{mat} $\mu$s in interpreted MATLAB and \val{time}{PROPOSED}{mex} $\mu$s in a C MEX loop, compared with \val{time}{MRAC}{mex} and \val{time}{LQR}{mex} $\mu$s for MRAC and LQR. These are minima over five batches of 20000 steps with fixed inputs on a desktop computer.

\section{Measurement-Path Variant and Second-Order INDI}\label{app:r3}

Both analyses were planned in writing before they were run; the plan is in the supplementary package.

\emph{Variant.} In the measurement seen by every controller, the gyro signal receives, in the noisy scenarios, vibration at the hover rotor frequency $f_1=\Omega_h/2\pi\approx98.8$\,Hz and at $2f_1$ with RMS amplitudes $2\sigma_\omega$ and $\sigma_\omega$ and random phases; the gyro then passes through a second-order Butterworth low-pass at 40\,Hz (all scenarios); position and velocity are delayed by 10\,ms (N1) or 20\,ms (N2). The amplitudes are assumptions chosen as a stress test; no flight data of this platform were available. With the variant switched off, the code reproduces archived activities to 13 digits. The budget is 1.05 times the S7 activity of LQR under N1 (\val{rB}{n1}{budget}\,rad/s). The composite families keep their guarded hyperparameters and are retuned on the gain line $\gamma_o\in\{10,\dots,5120\}$; GEO-DOB and INDI are retuned on their full seed-1 grids; the 20-seed guard follows the rules of Section~\mref{sec:tuning}. N2 evaluates the N1 profiles without retuning. Table~\ref{tab:r3-noise} gives the selection under N1 and the evaluation of the selected profiles under N1 and N2; the results are discussed in Section~\mref{sec:budget}.

\begin{table*}[!t]\centering\caption{Measurement-path variant: selection and evaluation}\label{tab:r3-noise}
\footnotesize\setlength{\tabcolsep}{4pt}\begin{tabular}{@{}lllrrrrrr@{}}\toprule
 & \multicolumn{3}{c}{selection under N1} & \multicolumn{3}{c}{N1} & \multicolumn{2}{c}{N2}\\\cmidrule(lr){2-4}\cmidrule(lr){5-7}\cmidrule(l){8-9}
Profile & seed 1 & guarded & guard & $J$ & Eval6 & S7 & Eval6 & S7\\\midrule
CL-att & 80 (4/10) & 80 & 3/3 & 50.6 & 59.2 & 1.028 & 60.7 & 1.041 \\
CL-Inner & 160 (5/10) & 160 & 3/3 & {\bfseries 46.4} & {\bfseries 54.2} & 1.034 & {\bfseries 56.3} & 1.055 \\
{\bfseries CL-Both} & 320 (6/10) & 160 & 2/3 & 46.5 & 54.4 & 1.025 & 56.5 & 1.039 \\
{\bfseries CL-LPF} & 320 (6/10) & 160 & 2/3 & 46.5 & 54.3 & 1.025 & 56.4 & 1.039 \\
GEO-DOB & 924 (463/1074) & 924 & 9/10 & 73.0 & 158.9 & 0.894 & 237.8 & 0.895 \\
INDI & 71 (24/107) & 71 & 9/10 & 113.7 & 275.1 & 0.943 & 347.9 & 0.943 \\
LQR & -- & -- & -- & 326.0 & 473.8 & 1.000 & 465.0 & 1.000 \\
\bottomrule\end{tabular}\par\smallskip\parbox{\textwidth}{\footnotesize Vibration, 40\,Hz gyro filter and position/velocity delay. Seed-1 selection (admissible/total), guarded selection and guard passes under N1 (10\,ms delay); design cost $J$, Eval6 (mm) and S7 activity relative to LQR under N1 and, with unchanged profiles, under N2 (20\,ms delay). GEO-DOB and INDI candidates are grid indices; composite candidates are $\gamma_o$.}
\end{table*}

\emph{INDI-2nd.} The second-order filter states replace $(\bv{v}-\bv{v}_f)/\tau$ and $(\bv{\omega}-\bv{\omega}_f)/\tau$; with the second-order option switched off, the INDI code reproduces the archived activities to 13 digits. The guarded INDI-2nd profile uses the INDI gains with $f_c=16$\,Hz. Because 16\,Hz was the largest cut-off of the planned grid, the grid was extended under a second written plan to $f_c\in\{32,64,128\}$\,Hz for all 16 gain pairs. None of the 48 new candidates meets the budget: their seed-1 S7 activities are 29.1--72.3\,rad/s at 32\,Hz, 147--310\,rad/s at 64\,Hz and 568--618\,rad/s at 128\,Hz, against 13.03\,rad/s, so the selection and its evaluation are unchanged. Its S1--S10 errors and fresh Monte Carlo statistics are in Tables~\mref{tab:main} and~\mref{tab:mc}. Its maneuver activity in the fresh draws is about 9\,rad/s, roughly half that of CL-Both G.

\section{PX4 SITL Details}\label{app:sitl}

\begin{figure}[!t]\centering\includegraphics[width=\columnwidth]{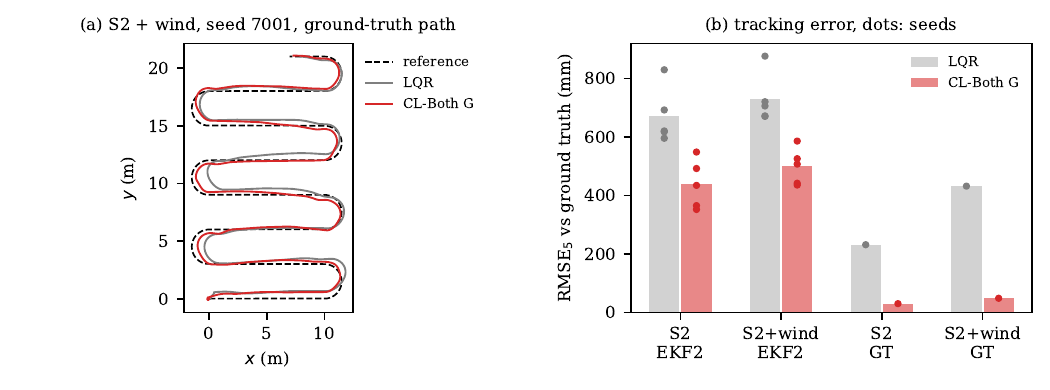}
\caption{PX4 SITL. (a) Ground-truth paths with wind, seed 7001. (b) S2 RMSE$_5$ against ground truth with EKF2 (five seeds, dots) and with ground-truth state feedback (GT).}\label{fig:sitl}\end{figure}

\emph{Setup.} Sensor noise enters through the SIH inertial, magnetometer, barometer and GPS models and EKF2, in place of the additive noise of S5. Three cases are flown with five new noise seeds (7001--7005) each: the S2 path, the S2 path with wind, and a 25\,s hover at 2\,m. The wind is a deterministic profile added to SIH (mean 2.5\,m/s, two sinusoids and a 2\,m/s gust between 30 and 40\,s) that acts through drag relative to the air. One run per path case with ground-truth state feedback separates controller effects from estimation error. Table~\ref{tab:sitl} summarizes this campaign, and Fig.~\ref{fig:sitl} shows the ground-truth paths with wind and the S2 errors of the individual seeds.

\begin{table}[!t]\centering
\caption{PX4 SITL results}\label{tab:sitl}
\renewcommand{\arraystretch}{1.1}\footnotesize\setlength{\tabcolsep}{2.5pt}
\resizebox{\columnwidth}{!}{\begin{tabular}{@{}llrrrr@{}}\toprule
Case & Controller & RMSE$_5$ & est.\ track & $\Delta\Omega$ & step \\ \midrule
S2, EKF2 & LQR & \val{sitlSii}{LQR}{RMSE5}$\pm$\val{sitlSii}{LQR}{RMSE5sd} & \val{sitlSii}{LQR}{RMSEest}$\pm$\val{sitlSii}{LQR}{RMSEestsd} & \val{sitlSii}{LQR}{DW}$\pm$\val{sitlSii}{LQR}{DWsd} & \val{sitlSii}{LQR}{cpu} \\
 & {\bfseries CL-Both G} & {\bfseries \val{sitlSii}{PROPOSED}{RMSE5}$\pm$\val{sitlSii}{PROPOSED}{RMSE5sd}} & {\bfseries \val{sitlSii}{PROPOSED}{RMSEest}$\pm$\val{sitlSii}{PROPOSED}{RMSEestsd}} & \val{sitlSii}{PROPOSED}{DW}$\pm$\val{sitlSii}{PROPOSED}{DWsd} & \val{sitlSii}{PROPOSED}{cpu} \\
\addlinespace
S2+wind, EKF2 & LQR & \val{sitlWind}{LQR}{RMSE5}$\pm$\val{sitlWind}{LQR}{RMSE5sd} & \val{sitlWind}{LQR}{RMSEest}$\pm$\val{sitlWind}{LQR}{RMSEestsd} & \val{sitlWind}{LQR}{DW}$\pm$\val{sitlWind}{LQR}{DWsd} & \val{sitlWind}{LQR}{cpu} \\
 & {\bfseries CL-Both G} & {\bfseries \val{sitlWind}{PROPOSED}{RMSE5}$\pm$\val{sitlWind}{PROPOSED}{RMSE5sd}} & {\bfseries \val{sitlWind}{PROPOSED}{RMSEest}$\pm$\val{sitlWind}{PROPOSED}{RMSEestsd}} & \val{sitlWind}{PROPOSED}{DW}$\pm$\val{sitlWind}{PROPOSED}{DWsd} & \val{sitlWind}{PROPOSED}{cpu} \\
\addlinespace
S2, GT & LQR & \val{sitlGTSii}{LQR}{RMSE5} & -- & \val{sitlGTSii}{LQR}{DW} & \val{sitlGTSii}{LQR}{cpu} \\
 & {\bfseries CL-Both G} & {\bfseries \val{sitlGTSii}{PROPOSED}{RMSE5}} & -- & \val{sitlGTSii}{PROPOSED}{DW} & \val{sitlGTSii}{PROPOSED}{cpu} \\
S2+wind, GT & LQR & \val{sitlGTWind}{LQR}{RMSE5} & -- & \val{sitlGTWind}{LQR}{DW} & \val{sitlGTWind}{LQR}{cpu} \\
 & {\bfseries CL-Both G} & {\bfseries \val{sitlGTWind}{PROPOSED}{RMSE5}} & -- & \val{sitlGTWind}{PROPOSED}{DW} & \val{sitlGTWind}{PROPOSED}{cpu} \\
\addlinespace
Hover, EKF2 & LQR & \val{sitlHover}{LQR}{RMSE5}$\pm$\val{sitlHover}{LQR}{RMSE5sd} & \val{sitlHover}{LQR}{RMSEest}$\pm$\val{sitlHover}{LQR}{RMSEestsd} & \val{sitlHover}{LQR}{DWfive}$\pm$\val{sitlHover}{LQR}{DWfivesd}$^\ast$ & \val{sitlHover}{LQR}{cpu} \\
 & {\bfseries CL-Both G} & {\bfseries \val{sitlHover}{PROPOSED}{RMSE5}$\pm$\val{sitlHover}{PROPOSED}{RMSE5sd}} & {\bfseries \val{sitlHover}{PROPOSED}{RMSEest}$\pm$\val{sitlHover}{PROPOSED}{RMSEestsd}} & \val{sitlHover}{PROPOSED}{DWfive}$\pm$\val{sitlHover}{PROPOSED}{DWfivesd}$^\ast$ & \val{sitlHover}{PROPOSED}{cpu} \\
\bottomrule
\multicolumn{6}{@{}l}{\footnotesize $^\ast$activity for $t>5$\,s of a 25\,s hover at 2\,m; step: mean wall-clock time of the controller step.}
\end{tabular}}\par\smallskip\parbox{\columnwidth}{\footnotesize SIH, v1.15.4. RMSE$_5$ against ground truth (mm), RMSE$_5$ of the estimate against the reference (mm), command activity $\Delta\Omega$ (rad/s) and step time ($\mu$s). Mean$\pm$SD over five sensor-noise seeds; GT: ground-truth state feedback, one run.}
\end{table}

\emph{Tracking.} With EKF2 feedback, the estimated position differs from the true position by \val{sitlSii}{LQR}{esterr} and \val{sitlSii}{PROPOSED}{esterr}\,mm RMS in S2 for LQR and CL-Both, comparable in size to the error against ground truth. The ground-truth error equals the estimate-tracking error minus the estimation error, so the two do not add in quadrature, and their cross term was not computed. CL-Both has the lower ground-truth RMSE$_5$ on \val{sitlSii}{all}{wins} of 5 seeds in S2 and on \val{sitlWind}{all}{wins} of 5 with wind, with paired reductions of \val{sitlSii}{all}{redmin}--\val{sitlSii}{all}{redmax}\,\% and \val{sitlWind}{all}{redmin}--\val{sitlWind}{all}{redmax}\,\%. Measured against the estimate, which removes the estimation error, CL-Both reduces the error from \val{sitlSii}{LQR}{RMSEest} to \val{sitlSii}{PROPOSED}{RMSEest}\,mm in S2 and from \val{sitlWind}{LQR}{RMSEest} to \val{sitlWind}{PROPOSED}{RMSEest}\,mm with wind. These estimate-tracking errors exceed the ground-truth-feedback errors because the controllers act on estimates held between EKF2 updates and on the GPS and barometer corrections of the estimate.

\emph{Activity.} The activity limit does not transfer unchanged. After the first 5\,s of hover, CL-Both commands a mean $\Delta\Omega=$\val{sitlHover}{PROPOSED}{DWfive}\,rad/s against \val{sitlHover}{LQR}{DWfive}\,rad/s for LQR, a ratio of means of \val{sitlHover}{all}{ratio}, but the paired ratio ranges from \val{rR}{sitl}{hovlo} to \val{rR}{sitl}{hovhi} over the five seeds; over the S2 path the ratio of mean $\Delta\Omega$ is \val{sitlSii}{all}{ratio}. The guard was calibrated with the additive Gaussian noise of S7, whereas here the noise reaches the controller through the PX4 sensor models, filters and EKF2. A re-selection in the PX4 stack, planned before the runs, flew CL-Both with $\gamma_o\in\{160,320,640\}$ in hover on seeds 7101--7105: the median paired activity ratios are \val{rV}{sitl160}{medratio}, \val{rV}{sitl320}{medratio} and \val{rV}{sitl640}{medratio}, no gain keeps the ratio within 1.05 on every seed (largest ratios \val{rV}{sitl160}{maxratio}, \val{rV}{sitl320}{maxratio} and \val{rV}{sitl640}{maxratio}), and activity does not fall with the gain, so a lower gain alone does not restore the budget.

\emph{Sensor timeouts.} These per-seed ratios are confounded by the host. In many runs the simulated barometer or magnetometer reports a wall-clock timeout during flight (\val{sitl}{all}{nto} of the \val{sitl}{all}{nruns} campaign runs and \val{rV}{sitl}{nto} of the \val{rV}{sitl}{nruns} re-selection runs, counting timeouts at any time), and runs with such a timeout tended to command more activity: in the follow-up below, the median hover $\Delta\Omega$ of LQR is \val{rR}{sitl}{lqrclean}\,rad/s without and \val{rR}{sitl}{lqrto}\,rad/s with an in-flight timeout. Only zero to two seed pairs per case of the campaign and re-selection are free of in-flight timeouts, and those give hover ratios of \val{rR}{sitl}{cleanlo} and \val{rR}{sitl}{cleanhi} (Table~\ref{tab:r2-sitl}).

The simulated barometer and magnetometer of SIH are checked against wall-clock time; on the WSL2 host used here they report timeouts in most runs, before and during flight, at a CPU load of about 1\,\%. A run is counted as timeout-free in flight if no timeout is logged after the controller starts. In the campaign and re-selection runs of Section~\mref{sec:sitl}, the pairs in which both runs are timeout-free in flight give the activity ratios (CL-Both/LQR) hover: 1.12, 0.91; S2: 1.60, 1.18; S2 with wind: none; hover re-selection $\gamma_o=160$: 1.09; 320: none; 640: none.

\emph{Filtered estimates.} A follow-up, planned before the runs, filtered the estimates seen by the adaptive part (regressors, adaptation errors and targets; the nominal law keeps the raw estimates) by a first-order filter with time constant $\tau_e$, at $\gamma_o=1280$, on ten new seeds (7301--7310). With $\tau_e=0$, 20 and 50\,ms the median paired hover ratios are \val{rR}{sitl}{med0}, \val{rR}{sitl}{med20} and \val{rR}{sitl}{med50}, and no setting keeps the ratio within 1.05 on every seed; \val{rR}{sitl}{nto} of the \val{rR}{sitl}{nhov} hover runs report an in-flight timeout. Among runs without one, the median hover $\Delta\Omega$ relative to that of LQR is \val{rR}{sitl}{rel0}, \val{rR}{sitl}{rel20} and \val{rR}{sitl}{rel50} (an unpaired comparison added after the runs). Filtering the estimates therefore reduces the excess activity, but the budget is not shown to hold on every seed. On the S2 path with $\tau_e=20$\,ms, the setting with the lower median hover ratio, CL-Both keeps the lower ground-truth error on \val{rR}{sitl}{swins} of \val{rR}{sitl}{sn} seeds (mean RMSE$_5$ \val{rR}{sitl}{scl} against \val{rR}{sitl}{slqr}\,mm for LQR; \val{rR}{sitl}{snofilt}\,mm without filtering), but its median paired S2 activity ratio rises from \val{rR}{sitl}{smed0} to \val{rR}{sitl}{smed20}: the filter does not reduce the maneuver activity of the adaptive law. With $\tau_e=0$ the module reproduces the archived rotor commands bit for bit. Table~\ref{tab:r2-sitl} gives the hover and S2 results on seeds 7301--7310 (S2: 7301--7305) at $\gamma_o=1280$; the S2 setting $\tau_e=20$\,ms was chosen by the rule fixed before the runs (lower median hover ratio).

\begin{table*}[!t]\centering\caption{PX4 SITL follow-up with filtered estimates in the adaptive part}\label{tab:r2-sitl}
\footnotesize\setlength{\tabcolsep}{4pt}\begin{tabular}{@{}llrrrrrrrr@{}}\toprule Case & $\tau_e$ & seeds & median ratio & max ratio & $\le1.05$ & clean pairs (ratios) & unpaired & timeout runs & RMSE$_5$ CL / LQR (mm) \\\midrule
S2 & 0\,ms & 5 & 1.33 & 2.21 & 0/5 & 0 (--) & -- & 5 & 488 / 619 \\
S2 & 20\,ms & 5 & 1.54 & 2.35 & 0/5 & 0 (--) & 1.20 (1) & 4 & 483 / 619 \\
hover & 0\,ms & 10 & 1.14 & 3.89 & 3/10 & 2 (1.14, 1.14) & 1.16 (5) & 5 & 468 / 644 \\
hover & 20\,ms & 10 & 1.04 & 1.97 & 5/10 & 1 (1.05) & 1.05 (3) & 7 & 565 / 644 \\
hover & 50\,ms & 10 & 1.07 & 2.71 & 5/10 & 2 (1.05, 1.03) & 1.03 (4) & 6 & 466 / 644 \\
\bottomrule\end{tabular}\par\smallskip\parbox{\textwidth}{\footnotesize $\gamma_o=1280$. Ratio: paired $\Delta\Omega$ of CL-Both over LQR on the same seed (hover: $t>5$\,s; S2: whole run). Clean: both runs timeout-free in flight. Unpaired: median $\Delta\Omega$ of timeout-free runs over the median of timeout-free LQR runs (added after the runs). Timeout runs: CL-Both runs with an in-flight timeout. Unpaired: number of timeout-free CL-Both runs in parentheses. RMSE$_5$ against ground truth, mean over seeds. Per-run values are in the supplementary package.}\end{table*}

\emph{Attempt to remove the timeouts.} A test planned before the runs flew LQR in hover on seeds 7401--7405 with four host settings, to be adopted for a new campaign only if all five runs were free of timeouts. No setting met that rule (Table~\ref{tab:r3-smoke}); the campaign was therefore not repeated. Over these 20 runs the median hover activity is 0.585\,rad/s in the 3 runs without any timeout and 0.608\,rad/s in the 17 runs with one. A native Linux host was not available.

\begin{table}[!t]\centering\caption{PX4 SITL host settings tested against the sensor timeouts}\label{tab:r3-smoke}
\footnotesize\setlength{\tabcolsep}{3pt}\begin{tabular}{@{}llrrrr@{}}\toprule Opt. & setting & runs$^a$ & count$^b$ & pre.$^c$ & med.\ $\Delta\Omega$\\\midrule
O1 & settling 60\,s & 5/5 & 19 & 4/5 & 0.608 \\
O2 & priority \texttt{nice -20} & 3/5 & 8 & 5/5 & 0.586 \\
O3 & simulation speed 0.5 & 4/5 & 7 & 5/5 & 0.596 \\
O4 & O1--O3 combined & 5/5 & 23 & 4/5 & 0.928 \\
\bottomrule\end{tabular}\par\smallskip\parbox{\columnwidth}{\footnotesize LQR, hover, seeds 7401--7405. $^a$Runs with a timeout. $^b$Timeouts over the whole run. $^c$Runs with a ``Yaw estimate error'' preflight failure (the vehicle is armed by command). Median $\Delta\Omega$ in rad/s.}
\end{table}

\emph{Timing.} The mean controller step inside the PX4 process takes \val{sitlSii}{PROPOSED}{cpu}\,$\mu$s for CL-Both and \val{sitlSii}{LQR}{cpu}\,$\mu$s for LQR in S2, with 99th percentiles of \val{sitlSii}{PROPOSED}{cpuP} and \val{sitlSii}{LQR}{cpuP}\,$\mu$s, measured with a wall-clock timer on a 24-core x86 Windows host running the simulation in WSL2. Runs with timeouts are retained in every table.

\section{Composite Laws Without RBF Neurons}\label{app:lin}

To separate the RBF neurons from the lag-aligned target, the eight RBF entries of $\bv{\phi}_o$ and of $\bv{\phi}_i$ are set to zero, which leaves the bias, the linear terms and, in the inner loop, the gyroscopic products (7 of 15 and 10 of 18 entries). Lin-CL-att, Lin-CL-Both and Lin-CL-LPF keep the targets and all hyperparameters other than the gain of guarded CL-att, CL-Both and CL-LPF and are retuned on the gain line $\gamma_o\in\{10,20,\ldots,5120\}$, $\gamma_i=10\gamma_o$. The seed-1 rule, the 20-seed guard over the seed-1 gain and the two gains below it, the validation seeds 5001--5100 and the fresh draws 9001--9100 are those of the main comparison. The plan was written before the runs and is in the supplementary package. Table~\ref{tab:lin} compares each profile with its RBF counterpart.

\begin{table*}[!t]\centering\caption{Guarded composite profiles with and without RBF neurons}\label{tab:lin}
\small\setlength{\tabcolsep}{3.5pt}\begin{tabular}{lrrrrrrrrrr}\toprule
 & \multicolumn{3}{c}{Selection} & \multicolumn{2}{c}{Deterministic (mm)} & \multicolumn{5}{c}{Fresh draws}\\\cmidrule(lr){2-4}\cmidrule(lr){5-6}\cmidrule(lr){7-11}
Profile & $\gamma_o$ & guard bound & pass & $J$ & Eval6 & $\Delta\Omega_{\rm S2}$ & S2 & S4 & $\tilde\Delta_{\rm S2}$ [95\,\% CI] & $\tilde\Delta_{\rm S4}$ [95\,\% CI]\\\midrule
CL-att G & 80 & -- & 100/100 & 49.6 & 60.5 & 16.06 & 49.2 & 57.4 & -- & -- \\
Lin-CL-att G & 160 & 12.99 & 100/100 & 47.1 & 57.5 & 16.51 & 46.2 & 54.7 & $-3.0$ [$-3.2$, $-2.8$] & $-3.0$ [$-3.3$, $-2.4$] \\
{\bfseries CL-Both G} & 1280 & -- & 100/100 & 37.3 & 46.4 & 18.81 & 31.1 & 46.6 & -- & -- \\
Lin-CL-Both G & 1280 & 12.94 & 100/100 & 37.9 & 46.8 & 18.96 & 31.4 & 47.9 & 0.6 [0.4, 0.9] & 0.5 [0.3, 0.7] \\
{\bfseries CL-LPF G} & 1280 & -- & 100/100 & {\bfseries 37.2} & {\bfseries 45.6} & 18.71 & {\bfseries 30.9} & {\bfseries 46.0} & -- & -- \\
Lin-CL-LPF G & 1280 & 12.97 & 100/100 & 37.8 & 46.1 & 18.87 & 31.3 & 47.2 & 0.6 [0.4, 0.9] & 0.6 [0.3, 0.8] \\
\bottomrule\end{tabular}\par\smallskip\parbox{\textwidth}{\footnotesize $\gamma_o$: guarded gain; guard bound: $\max(\max\Delta\Omega,\ \mathrm{mean}+3\,\mathrm{SD})$ over seeds 4001--4020 in rad/s (budget 13.03); pass: S7 checks within budget on validation seeds 5001--5100; $\Delta\Omega_{\rm S2}$: mean S2 activity (rad/s); S2, S4: median RMSE$_5$ (mm); $\tilde\Delta$: median paired difference, profile without RBF neurons minus its RBF counterpart, over the 50 fresh draws of each scenario, with percentile bootstrap intervals (10\,000 resamples). All runs complete. RBF rows from Tables~\mref{tab:tune}, \mref{tab:main} and \mref{tab:mc}.}
\end{table*}

The two profiles with lag-aligned inner-loop targets select the same gains as with the RBF entries, and their design cost $J$ rises by about 0.63\,mm, Eval6 by 0.4--0.5\,mm and the median paired fresh-draw differences are 0.54--0.62\,mm, with bootstrap intervals that exclude zero. Without the RBF entries, CL-att passes the guard at $\gamma_o=160$ instead of 80; its $J$ falls by 2.5\,mm, and its Eval6 and median paired fresh-draw differences by about 3.0\,mm. All three profiles remain well below MRAC ($J=86.5$\,mm), which uses the same linear regressor without the composite term, and NN-MRAC ($J=82.6$\,mm). In these tests the RBF neurons therefore change the composite costs by at most about 3\,mm, and the improvement over gradient MRAC comes from the composite term with the lag-aligned target. The ablation fixes every hyperparameter except the gain at the values selected for the RBF profiles, which may favor the RBF profiles slightly.

\section*{Data and Code Availability}
The supplementary package of this article contains the MATLAB/Simulink sources, the PX4 SITL module and patch, the archived result files (CSV, JSON and small MAT files, with one summary row per simulation or SITL run) and the analysis scripts that produce the reported numbers, tables and figures. Raw simulation trajectories are not stored, because the scripts regenerate them; raw PX4 logs are not included. The package is publicly available at \url{https://github.com/Quang61123/lag-aligned-cl-mrac} under the MIT License.

\section*{Acknowledgment}
Claude (Anthropic) was used as an AI-assisted tool to support code development, verification, and refinement, as well as manuscript preparation and language editing (grammar, clarity, and wording). The authors reviewed the code, numerical results, and manuscript content and take full responsibility for the scientific integrity, accuracy, and conclusions of this work.

\bibliographystyle{IEEEtran}
\bibliography{references}
\end{document}